\documentclass{report}
\usepackage{suthesis-2e}
\newcommand{\mbf}[1]{{\mathbf{#1}}}
\newcommand{\mbx}{{\mathbf{x}}}
\newcommand{\mby}{{\mathbf{y}}}

\newcommand{\ac}{\textit{a~contrario} }
\newcommand{\act}{\textit{a-contrario} }
\newcommand{\be}{\begin{equation}}
\newcommand{\ee}{\end{equation}}

\newcommand{\beh}{\begin{equation*}}
\newcommand{\eeh}{\end{equation*}}

\newcommand{\bi}{\begin{itemize}}
\newcommand{\ei}{\end{itemize}}
\newcommand{\ba}{\begin{array}}
\newcommand{\bfg}{\begin{figure}}
\newcommand{\efg}{\end{figure}}
\newcommand{\bc}{\begin{center}}
\newcommand{\ec}{\end{center}}
\newcommand{\bbm}{\begin{bmatrix}}
\newcommand{\ebm}{\end{bmatrix}}

\newcommand{\refsec}[1]{Sec.~\ref{sec:#1}}
\newcommand{\reffig}[1]{Fig.~\ref{fig:#1}}
\newcommand{\refeq}[1]{Eq.~\ref{eq:#1}}
\newcommand{\reftab}[1]{Table~\ref{tab:#1}}

\newcommand{\ie}[1][ ]{{\em i.\thinspace{}e\@.{}}#1}
\newcommand{\eg}[1][ ]{{\em e.\thinspace{}g\@.{}}#1}
\newcommand{\ea}{{\em et al }}
\newcommand{\argmin}{\operatorname*{argmin}}
\newcommand{\argmax}{\operatorname*{argmin}}

\newcommand{\pose}[2]{^{#1}\mathbf{T}_{#2}}
\newcommand{\img}[1]{\mathbf{I}_{#1}}
\newcommand{\pt}{\mathbf{p}}
\usepackage{amsmath,bm,amssymb}
\usepackage{amsfonts}
\usepackage{setspace}
\usepackage{multirow}
\usepackage{eurosym}
\usepackage{subcaption} 
\usepackage{amsmath,bm,amssymb}
\usepackage{pdfpages}
\usepackage{arabtex}
\usepackage{utf8}
\setcode{utf8}

\dept{Computer Science}
\usepackage{graphicx}

\begin{document}
\includepdf[pages=-]{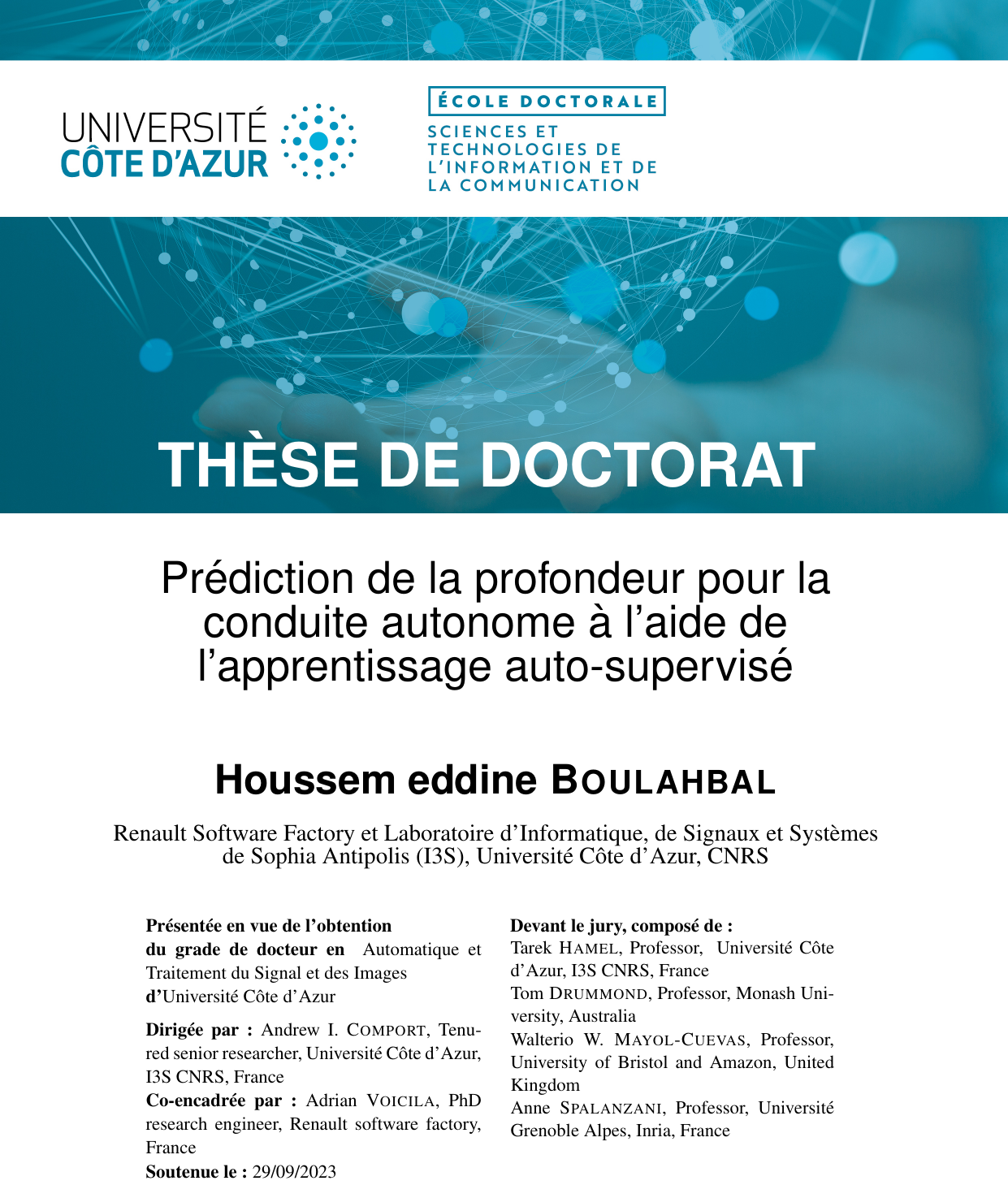}

\pagenumbering{roman}
\setcounter{page}{7}

\prefacesection{Résumé}
La perception de l'environnement est un élément essentiel de la conduite autonome. Elle permet au véhicule de comprendre son environnement et de prendre des décisions informées. La prédiction de la profondeur joue un rôle central dans ce processus, car elle aide à comprendre la géométrie et le mouvement de l'environnement. Cette thèse se concentre sur le défi de la prédiction de la profondeur en utilisant des techniques d'apprentissage auto-supervisé en utilisant des cameras monoculaire. En premier lieu, le problème est abordé d'un point de vue plus large, en explorant les réseaux adversaires génératifs conditionnels (cGAN) en tant que technique potentielle pour obtenir une meilleure généralisation. Ce faisant, une contribution fondamentale aux GAN conditionnels, le cGAN ac, a été proposée.
La deuxième contribution concerne une méthode auto-supervisée pour translater une image à une carte de profondeur, en proposant une solution pour les scènes rigides à l'aide d'une nouvelle méthode basée sur les transformeurs qui génère une pose pour chaque objet dynamique. Le troisième aspect important concerne l'introduction d'une approche de prévision du future de carte profondeur en utilisant la vidéo. Cette méthode sert d'extension aux techniques auto-supervisées pour prédire les profondeurs futures. Elle implique la création d'un nouveau modèle de transformateur capable de prédire la profondeur future d'une scène donnée. En outre, les diverses limitations des méthodes précédemment mentionnées ont été abordées et un modèle de cartes de profondeur vidéo-vidéo a été proposé. Ce modèle tire parti de la cohérence spatio-temporelle de la séquence d'entrée et de la séquence de sortie pour prédire une séquence de profondeur plus précise. Ces méthodes ont des applications significatives dans la conduite autonome et les systèmes avancés d'aide à la conduite. L'approche est auto-supervisée, ce qui élimine le besoin de labellisation manuelle des cartes de profondeur pendant la phase d'apprentissage, la rendant ainsi efficace et rentable. Dans l'ensemble, cette thèse apporte plusieurs contributions au domaine de la conduite autonome en développant une approche auto-supervisée de la prédiction de la profondeur. L'approche proposée est efficace, avec le potentiel d'améliorer la sécurité et la fiabilité des systèmes de conduite autonome. Les implications de ces résultats sont importantes pour la conception de systèmes avancés d'aide à la conduite et de véhicules autonomes, ce qui nous rapproche de l'objectif d'une conduite entièrement autonome.

\textbf{Keywords}
Profondeur, apprentissage profond, auto-supervisé, prédiction, conduite autonome

\prefacesection{Abstract}
Perception of the environment is a critical component for enabling autonomous driving. It provides the vehicle with the ability to comprehend its surroundings and make informed decisions. Depth prediction plays a pivotal role in this process, as it helps the understanding of the geometry and motion of the environment. This thesis focuses on the challenge of depth prediction using monocular self-supervised learning techniques. The problem is approached from a broader perspective first, exploring conditional generative adversarial networks (cGANs) as a potential technique to achieve better generalization was performed. In doing so, a fundamental contribution to the conditional GANs, the \ac cGAN was proposed. The second contribution entails a single image-to-depth self-supervised method, proposing a solution for the rigid-scene assumption using a novel transformer-based method that outputs a pose for each dynamic object. The third significant aspect involves the introduction of a video-to-depth map forecasting approach. This method serves as an extension of self-supervised techniques to predict future depths. This involves the creation of a novel transformer model capable of predicting the future depth of a given scene. Moreover, the various limitations of the aforementioned methods were addressed and a video-to-video depth maps model was proposed. This model leverages the spatio-temporal consistency of the input and output sequence to predict a more accurate depth sequence output. These methods have significant applications in autonomous driving (AD) and advanced driver assistance systems (ADAS). The approach is self-supervised, which eliminates the need for manual labeling of depth maps during training, making it efficient and cost-effective. Overall, this thesis makes several contributions to the field of autonomous driving by developing a self-supervised approach to depth prediction. The proposed approach is effective and efficient, with the potential to enhance the safety and reliability of autonomous driving systems. The implications of the findings are important for the design of advanced driver assistance systems and autonomous vehicles, bringing us one step closer to achieving the goal of fully autonomous driving.

\textbf{Keywords}
Depth, self-supervision, prediction, autonomous driving, deep learning
\prefacesection{Acknowledgments}
I would like to sincerely express my gratitude to everyone who has supported me throughout this incredible journey. Firstly, I am deeply thankful to God for His guidance and blessings, as I wouldn’t have reached this point without His steadfast support.
\\

I am immensely thankful to my supervisor, Andrew Comport, for granting me a unique and exceptional opportunity. His invaluable feedback and guidance have played a pivotal role in shaping my thesis and fostering my growth as a researcher. Working under his supervision in such an exceptional environment has been truly enriching. I would also like to extend my heartfelt gratitude to Adrian Voicila for his supervision and support, which were crucial in successfully completing this work. Furthermore, I want to extend my deepest appreciation to Tarek Hamel, whose support was instrumental in the launch of this thesis. I am genuinely grateful for his dedication.
\\

I am profoundly grateful to my friends and colleagues. Your camaraderie and positivity made this time truly extraordinary. I extend my sincere thanks to my family, both immediate and extended, for their unwavering support, unending encouragement, and constant presence in my life. Your support has been a wellspring of strength throughout this journey. I am endlessly grateful for each of you. Finally, to all who contributed, in ways big or small, your support has been instrumental in propelling me forward.

\afterpreface
\pagenumbering{arabic}
\setcounter{page}{1}
\chapter{Introduction}
\label{ch:ch1}
\section{Motivation}
\label{sec:motivation}
\begin{figure}
    \centering
    \includegraphics[width=\textwidth]{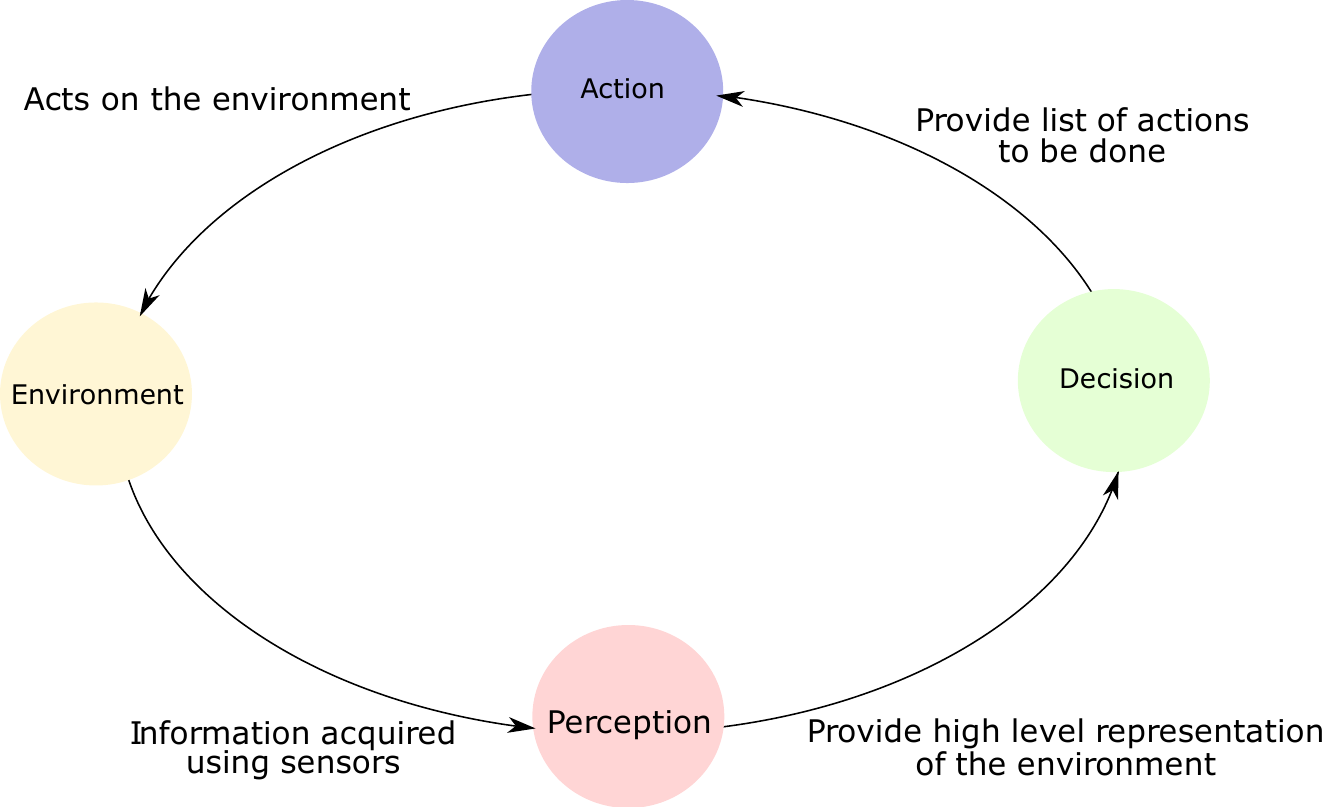}
    \caption{Illustration of steps of the cycle taken by an autonomous agent}
    \label{fig:steps_ad}
\end{figure}

Autonomous driving (AD) is one of the most complex research and engineering challenges. It refers to the ability of a vehicle to operate without human intervention. It is driving innovation for computer vision and mobile robotics. 
The potential benefits of AD are immense, including increased safety and comfort, reduced traffic congestion, and improved mobility for people with disabilities. 
The challenge of AD is multifaceted, and it requires the development of systems that can enable a vehicle to \textbf{perceive} its environment (this refers to a vehicle's ability to accurately interpret its surroundings, including identifying the geometry of the scene, vehicles, pedestrians …etc.), \textbf{make decisions}, and take appropriate \textbf{actions} as shown in~\reffig{steps_ad}. These tasks are not trivial, and they require the integration of various disciplines, such as computer vision, machine learning, mobile robotics, control theory, and others. Moreover, the development of AD systems requires addressing a wide range of issues, including legal and ethical considerations, societal impacts, and economic feasibility. Solving this challenge is a huge step towards developing general intelligence. 

Recent years have seen tremendous progress on deep learning (DL), especially in computer vision, with impressive results on several tasks such as classification~\cite{he2016deep,szegedy2015going,dosovitskiy2020image,liu2021swin}, detection~\cite{he2017mask,ren2015faster,redmon2016you,lin2017focal,law2018cornernet,dai2017deformable}, segmentation~\cite{badrinarayanan2017segnet,liu2015parsenet,chen2020spatialflow,kirillov2019panoptic} and depth prediction~\cite{safadoust2021self,lee2021attentive,Wang2021,Ranjan2019,Gordon2019,Godard2017,Zhou2017,Godard2019,Johnston2020,Rares2020}. One particular advantage of deep learning is the ability to extract features directly from data without the need for handcrafted expert systems. Deep learning methods have shown the potential to extrapolate into unseen situations that are not present in the training set for large scale datasets. This is practical, as one aspect of developing intelligent agents is the adaptability to new cases. Advances made in deep learning have motivated the research community to pursue and rely on deep learning methods as a dominant solution for addressing perception of the environment and providing these systems with intelligence.

The current endeavor to resolve the problem of general intelligence is to create systems inspired from humans. Our intelligence includes the ability to learn, memorize and adapt to new situations. When we are starting to learn driving, we first learn the possible actions we could perform on the car and how these actions change the perceived environment. However, becoming an effective driver also requires developing a “specialized driving perception”. An experienced driver has learned to identify all the pertinent agents in the scene, track and predict their behavior, asses the risk of each possible action and take the right action. The driver has \textbf{learned} a specialized driving perception that was based on his prior perception “model” and “fine-tuned” it to the new challenges of driving. During “inference” the driver applies all the aforementioned abilities intuitively. A key component of this driving perception is the ability to anticipate future states of the environment. Similarly, the development of an autonomous driving system requires the creation of an intelligent system that can anticipate future states of the environment. This will provide the vehicle with the ability to comprehend its surroundings and make well-informed decisions.

To enable an autonomous vehicle to perceive its surroundings, a range of sensors are installed, with the camera being a primary sensory modality. The visual information is central to understanding the environment. However, the RGB representation captured from cameras is not pertinent to AD applications as it does not explicitly provide the information about the geometry, the entities present in the scene and its motion.
Depth and semantics extract pertinent information from the high-dimension information present in the RGB representation, thus provide an alternative representation of the state of the scene that is suitable for making decisions. Depth information can provide an estimation of the distance between objects in the scene and enable the creation of a 3D representation of the environment. Moreover, semantic information can provide a high-level abstract representation of the entities present in the scene, such as road boundaries, traffic signals, and other vehicles. Therefore, depth information plays a pivotal role in providing a more comprehensive representation of the environment, enabling more effective decision-making for autonomous vehicles.

In the community, there has been an ongoing and in-depth discussion regarding the use of cameras versus LiDAR technology as the primary source of depth perception. While LiDAR is undoubtedly a powerful technology for depth perception, it has several advantages, including real-time and high accurate depth sensing at the sensor level. However, the cost of implementing it in automotive applications is a significant drawback. In contrast, cameras have emerged as a more cost-effective and practical solution for achieving accurate depth perception, making them a popular choice for many automotive manufacturers. In addition to their affordability and versatility, cameras also offer significant advantages in terms of power consumption and computing resources. Compared to LiDAR, cameras require less power to operate and fewer computing resources to process data. Moreover, the advances in deep learning over recent years have significantly improved the capabilities of camera-based-depth perception. 
One promising avenue of research in this area is the use of self-supervised learning approaches for depth prediction using cameras. These approaches circumvent the need for huge labeled datasets that are expensive and laborious to collect. By leveraging the vast amounts of unlabeled data available in the form of videos, it is possible to train deep learning models to predict depth without the need for explicit supervision or ground truth data.

To clarify ambiguous terminology found across different papers in the literature, the term “prediction” will be considered here to encompass both “depth inference” and “future forecasting”. The term “inference” will be reserved for the prediction at time $t$ (\ie the mapping $\mbf{D}_t= f(\img{t};\bm\theta)$ where $\mbf{D}_t$ is the depth map at time $t$, $\img{t}$ is the image at time $t$ and $\bm\theta$ are the model's parameters) and ”forecasting” will be used for future predictions $(t +1 : t +k)$ (\ie the mapping $\mbf{D}_{t+1:t+k}= f(\img{t-n:t};\bm\theta)$ such as $k$ is the future horizon and $n$ is the past context).

\section{Objective}
This thesis aims to investigate the use of depth maps as a representation of a scene, with a focus on developing self-supervised deep learning models for monocular depth prediction. The ultimate goal is to solve the prediction of the future depth as a way to bring intelligence into the systems As discussed in~\refsec{motivation}. 

Depth maps provide a rich representation of a scene, allowing the understanding of motion and geometry. 
By utilizing deep learning, It is possible to develop models that are highly accurate and can generalize to new scenarios, allowing for the development of more effective and efficient intelligent systems. However, this requires huge and annotated datasets, which is impractical. Therefore, self-supervision approaches will be leveraged to train these models, allowing to exploit the vast amounts of unlabeled data available, and providing for the development of models that are both accurate and efficient.
In order to enable a wide range of applications, the explored models in this thesis are predominantly monocular, which is the most challenging setting, since recovering 3D from 2D images is an ill-posed problem. However, this will enable a wide range of applications.

Predicting the future depth of a scene is a challenging task, but has significant implications for intelligent systems. By accurately predicting future depth, systems can better anticipate and react to changes in their environment, allowing for more efficient and effective operation. 

\section{Contribution}

This thesis focuses on the challenge of depth prediction using self-supervised learning techniques. several key contributions were made.  
Firstly, the problem is approached from a broader perspective, exploring conditional generative adversarial networks (cGANs) as a potential technique to achieve better generalization was performed. In doing so, a fundamental contribution to the conditional GANs, the \ac cGAN was proposed.
The second significant contribution involves the development of a self-supervised method for single image-to-depth inference. This method proposes a solution to overcome the rigid-scene assumption of the classical SfM model by utilizing a novel transformer-based approach that outputs a pose for each dynamic object.
The third contribution revolves around the proposal of a video-to-depth forecasting approach. This includes the development of a novel transformer model capable of forecasting the future depth of a scene, thereby extending the application of self-supervised methods to predict future depths.
Finally, the various limitations of the aforementioned method were addressed and a video-to-video depth prediction model was proposed. This model leverages the spatio-temporal consistency of the input and output sequence to predict a more accurate depth sequence output. 
These methods have significant applications in autonomous driving (AD) and advanced driver assistance systems (ADAS). Our approach is self-supervised, which eliminates the need for manual labeling of depth maps during training, making it efficient and cost-effective.

\section{Outline}
The organization of this thesis is as follows: Chapter~\ref{ch:ch2} provides an overview of the background information relevant to the thesis. This includes a review of machine learning basics, and a literature review on deep learning depth methods. Additionally, a brief literature review is presented that explores the use style transfer (will be used in Chapter~\ref{ch:ch3})of semantic information (will be used in Chapter~\ref{ch:ch3} and Chapter~\ref{ch:ch4}) and the Transformer module as building block for the proposed architectures.

 Chapter~\ref{ch:ch3} The problem is approached with a wider perspective. The generalization of deep learning models was explored. This was done through domain adaptation methods, by utilizing generative adversarial networks, specifically conditional GANs for style transfer. During this exploration, a fundamental limitation of cGANs is revealed, namely their lack of complete conditionality. To address this issue, the chapter presents an innovative solution referred to as the “\ac method”. The main objective of the \ac method is to enhance conditional GANs and empower them with full conditionality.

Chapter~\ref{ch:ch4} explores single-image-to-depth inference and extends the classical self-supervised methods with dynamic objects. Before aiming to forecast the future, it is essential at first to understand the present. In this chapter, we propose a solution for the static-scene assumption of the classical SfM model, using a novel transformer-based method that outputs a pose for each dynamic object. This model is a single image-to-depth mapping. The depth model takes a single image as input and outputs the corresponding depth map.

Chapter~\ref{ch:ch5} explores video-to-depth forecasting. This was a first attempt to forecast the future depth using self-supervision. The input to this model is a sequence of present and past images, and the model output a depth map that represents the future depth at step $k$. A novel transformer-based architecture is proposed to aggregate the temporal information, this enables the network to learn a rich spatio-temporal representation.

Chapter~\ref{ch:ch6} explores a video-to-video depth model. This model takes a sequence of images of past and present images and outputs a sequence of the present and future depth maps. This method addresses the limitations of the previous methods and extends the forecasting into a sequence of future depth. A self-supervised model that simultaneously predicts a sequence of future frames from video input with a novel spatial-temporal attention (ST) network is presented.

Finally, in Chapter~\ref{ch:ch7}, summarizes the main contributions and presents perspectives for future work.

\chapter*{List of publication}
The work presented in this thesis led to the following publications: 

\textbf{International publications :}
\begin{itemize}
    \item \textbf{Journal paper:} Boulahbal Houssem Eddine, Adrian Voicila, and Andrew I. Comport. "Instance-aware multi-object self-supervision for monocular depth prediction." IEEE Robotics and Automation Letters 7.4 (2022): 10962-10968.

    \item \textbf{Conference paper:} Boulahbal Houssem Eddine, Adrian Voicila, and Andrew I. Comport. "Are conditional GANs explicitly conditional?." British Machine Vision Conference. 2021.
    
    \item \textbf{Conference paper:} Boulahbal Houssem Eddine, Adrian Voicila, and Andrew I. Comport. "Instance-aware multi-object self-supervision for monocular depth prediction." 2022 35th International Conference on Intelligent Robots and Systems (IROS). IEEE, 2022.
        
    \item \textbf{Conference paper:} Boulahbal Houssem Eddine, Adrian Voicila, and Andrew I. Comport. "Forecasting of depth and ego-motion with transformers and self-supervision." 2022 26th International Conference on Pattern Recognition (ICPR). IEEE, 2022.

    \item \textbf{Conference paper:} Boulahbal Houssem Eddine, Adrian Voicila, and Andrew Comport. "STDepthFormer: Predicting Spatio-temporal Depth from Video with a Self-supervised Transformer Model." arXiv preprint arXiv:2303.01196 (2023). \textbf{To be submitted}

\end{itemize}

\textbf{National publication :} 
\begin{itemize}

\item \textbf{Conference video poster:} Boulahbal, Houssem Eddine, Adrian Voicila, and Andrew I. Comport. "Un apprentissage de bout-en-bout d'adaptateur de domaine avec des réseaux antagonistes génératifs de cycles consistants." Journée des Jeunes Chercheurs en Robotique. 2020.

\end{itemize}

\chapter*{}

\chapter{Background}
\label{ch:ch2}
\section{Machine learning basics}
Learning is the process of acquiring new knowledge or skills through experience or study. Machine learning attempts to create algorithms that can gain knowledge from and make decisions based on data. Primarily, machine learning involves designing models that are capable of extracting knowledge or insights from data employing the learning procedure.

According to Mitchell~\cite{mitchell1997machine} “A computer program is said to learn from experience $E$ with respect to some class of tasks $T$ and performance measure $P$, if its performance at tasks in $T$, as measured by $P$, improves with experience $E$.” In the context of deep learning, the computer program is a model $\hat{\mby}=f(\mbx;\bm{\mathbf{\bm{\theta}}})$ defined by its parameters $\mathbf{\bm{\theta}}$. $\mbf{x}$ is the input variable and $\hat{\mby}$ is the output of the model. 
The experience, or dataset, is a collection of examples $\mbf{D} = {(\mbx_1,\mby_1), (\mbx_2,\mby_2), ..., (\mbx_N,\mby_N)}$ where each example is a pair of input $\mbx$ and output $\mby$. The task varies depending on the user's need, \ie classification, object detection …etc. The performance measure P assesses how the model performs on the given task.  

For deep learning, the back-propagation algorithm is commonly used to learn these parameters through optimization with gradient descent. More often, the performance metric P is not differentiable and cannot be used with the back-propagation framework. In this case, a surrogate loss function is used instead. It acts as a proxy for the performance metric. By minimizing the loss function, the performance measure is improved. For example, the cross-entropy loss is the proxy loss for classification precision, and minimizing it leads to improved classification precision. Learning can be defined more formally as: 
\begin{align}
    \label{eq:risk}
    \mbf{R}(\bm{\theta}) = \min_{\bm{\theta}} \mathbb{E}_{(\mbx,\mby) \sim p_{data}(\mbx,\mby)} L(f(\mbf{x};\bm{\theta}),\mbf{y})
\end{align}
Learning involves finding the set of parameters $\bm{\theta}$ that minimizes the expected value of the loss function L across the data generating distribution $p_{data}(\mbx,\mby)$. The quantity $\mbf{R}(\bm{\theta})$ is known as the risk. In practice, however, this quantity cannot be optimized as $p_{data}(\mbx,\mby)$ is not known. Instead, the empirical distribution represented by the training set $\mbf{D}_{train}$ is used and \refeq{risk} becomes: 
\begin{align}
\label{eq:empiricalrisk}
    \mbf{r}(\bm{\theta}) = \min_{\bm{\theta}} \sum_{D_{train}} L(f(\mbf{x};\mathbf{\bm{\theta}}),\mbf{y})
\end{align}
This quantity is known as the \textbf{empirical risk} it is also known as the \textbf{generalization error}.

\subsection{Generalization}

Generalization is achieved when the model is capable of learning representative and abstract features that capture complex relationships and patterns in the data. In order to assess the model for generalization,~\refeq{empiricalrisk} is also calculated for the validation dataset $\mbf{D}_{val}$ : 
\begin{align}
\label{eq:empiricalrisk_val}
    \mbf{r}(\bm{\theta}) = \min_{\bm{\theta}} \sum_{D_{val}} L(f(\mbf{x};\mathbf{\bm{\theta}}),\mbf{y})
\end{align}
A model reaches good generalization when the training error~\refeq{risk} is very low and the gap between the validation~\refeq{empiricalrisk_val} and training errors~\refeq{risk} is very low. Poor generalization can be divided into two categories:
\begin{itemize}
    \item \textbf{Over-fitting :} If the gap between the training and validation generalization error is big, this indicates that the model has memorized the training set, and it is not able to generalize for unseen data. This happens when the model is too complex and has learned patterns and relationships that are specific to the training data
    \item \textbf{Under-fitting: } This is characterized when the training error and validation error are poor, the validation error may even be lower than the training error. This happens when the model is too simple, and it is not capable of learning a meaningful and useful feature.
\end{itemize}
Generalization could be improved using several techniques including regularization, increasing the dataset size, and using ensemble methods. 

\subsection{Representation learning}
 The complexity of the data can sometimes make the learning process challenging. For example, an RGB image with a resolution of $1024\times 512$ and 8-bit representation per pixel contains a total of $(256 \times 256 \times 256)^{1024\times 512}$ possible images. Modeling the distribution of this high-dimensional space is intractable.

\begin{figure}[h]
\centering
\begin{subfigure}{.5\textwidth}
  \centering
  \includegraphics[width=.8\linewidth]{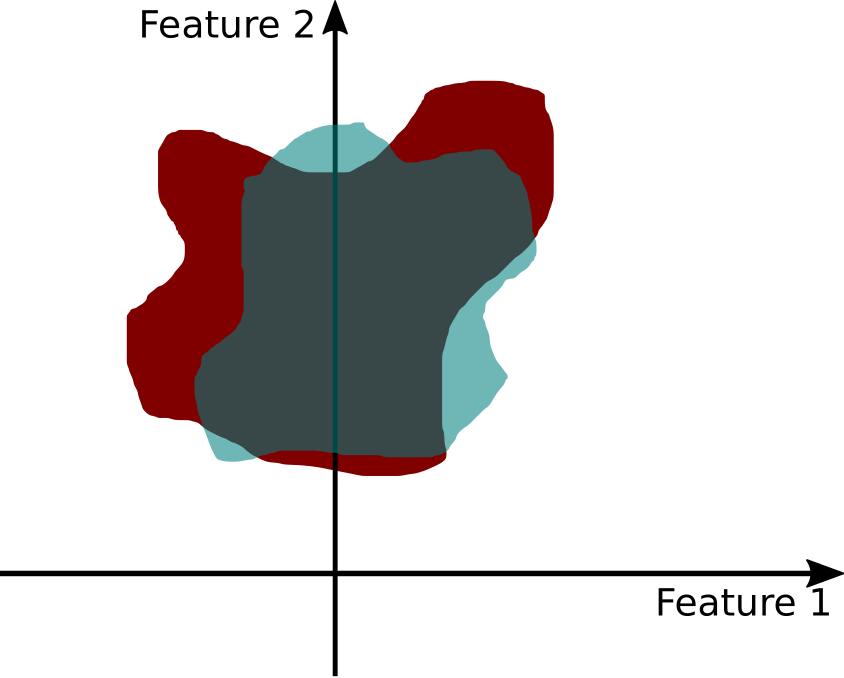}
  \caption{This representation is not suitable for \\classification applications}
  \label{fig:manifoldA}
\end{subfigure}%
\begin{subfigure}{.5\textwidth}
  \centering
  \includegraphics[width=.8\linewidth]{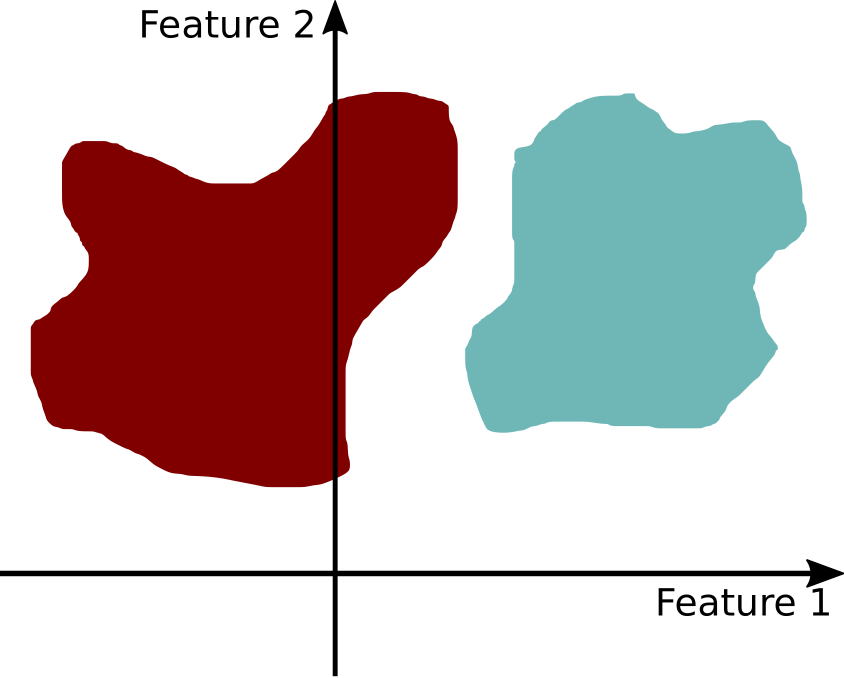}
  \caption{This representation is better suitable for classification applications as it shows clearly the decision boundary}
  \label{fig:manifoldB}
\end{subfigure}
    \caption{An example of a possible mapping of a dataset that contains two classes using two features}
    \label{fig:manifold}
\end{figure}

Despite the fact that images are typically represented as high-dimensional data, it has been observed that natural images actually reside on a low-dimensional manifold. In other words, the set of all natural images exists in a lower-dimensional subspace of the high-dimensional space. One of the key insights behind the manifold hypothesis is that natural images exhibit a high degree of structure and regularity, which can be captured by low-dimensional representations. 
This assumption, which is central to machine learning, enables the identification of a meaningful representation of the data. The goal of learning is to find a good representation of this manifold. Good generalization occurs when the model is able to learn a good representation that not only accurately represents the training examples \ie interpolation, but can also accurately predict the behavior of examples that were not seen during training \ie extrapolation.~\reffig{manifold} illustrates how the data can be mapped onto two different manifolds. A good representation allows us to leverage it for the chosen tasks. For example, the representation \reffig{manifoldB} is better than \reffig{manifoldA} for classification, as it is easier to define the decision boundary that separates the two domains.  

Learning algorithms can be classified into different categories based on the type of supervision they require. Supposing $\hat{\mbf{y}}=f(\mbf{x};\bm{\theta})$ we can make the following definitions :

\begin{itemize}
    \item \textbf{Supervised learning:} In this setting the ground-truth $\mby$ is known, and it is leveraged when calculating the loss function. The model is trained on a labeled ground-truth dataset, and it makes predictions based on this input/output mapping. Some examples of supervised learning tasks include classification: Predicting which category a new example belongs to (e.g., spam or not spam); and regression: Predicting a continuous value (e.g., the depth of an object in the scene).
    
    \item \textbf{Unsupervised learning} The model does not have access to the ground-truth labels. the model is trained to discover the underlying pattern of the data.  Some examples of unsupervised learning tasks include image generation, clustering, and domain adaptation. 

    \item \textbf{Self-supervised learning} is a special class of unsupervised learning where the model is trained against a proxy task, and when improving the proxy the downstream task is also improved. An example of self-supervision is self-supervised depth prediction. the proxy task is the image reconstruction and the depth is the downstream task. Training this model for image reconstruction provides a sufficient gradient signal to supervise the depth.  

    \item \textbf{Semi-supervised} These models leverage both supervised and unsupervised techniques. When working with partially labeled datasets, it can be beneficial to incorporate both supervised and unsupervised techniques in order to make the most of the available data, especially when the labeling cost is high.

\end{itemize}
Based on the applications, two types of models could be distinguish : generative models and predictive models. For a dataset $p_{data}(\mbx,\mby)$, the generative models try to model the distribution of the data $p(\mbx)$. Predictive models on the other hand model the distribution of the output y as $p(\mby|\mbx)$ usually y have small dimensions such as a class label.


\subsection{Generative models}
Generative models aim to model the distribution of $p(\mbf{x})$ or a conditional distribution $p(\mbf{x}|\mbf{y})$ such as $\mbf{y}$ is the label of $\mbf{x}$. By learning the distribution of the dataset, it is possible to generate new samples $\mbf{x}_{sythetic} \sim p(\mbf{x})$. Application of these models includes image generation, text generation, text-to-images, image-to-image translation, domain adaptation. 
The most used generative models are: Variational auto-encoders (VAEs), Generative adversarial networks GANs, and Diffusion models.

\subsubsection{Variational auto-encoders}
An autonecoder is a neural network that is used to reconstruct its input. It consists of an encoder that models $p(\mbf{h}|\mbx)$ and a decoder that models $p(\mbx_{rec}|\mbf{h})$. AE can be used as a generative model by leveraging the decoder. an auto-encoder defines a generative model of the form:  
\begin{align}
    \label{eq:autoencoder}
    p_{model}(\mbf{x}) = \int p(\mbf{x}|\mbf{h})  \: p(\mbf{h}) dh
\end{align}
Where $h$ is the latent variable. If the autoencoder is trained as a generative model, it is optimized to maximize the likelihood of $p_{model}(\mbf{x})$ with respect to $p_{data}(\mbf{x})$. However, the exact inference of \refeq{autoencoder} is intractable. It is possible to approximate this quantity using the evidence lower bound defined as follows: 
\begin{align}
    p_{model}(\mbf{x}) &=   \int p(\mbf{x}|\mbf{h})   p(\mbf{h}) dh \\ 
    log [p_{model}(\mbf{x})]  &= log \big[\int p(\mbf{x}|\mbf{h})   p(\mbf{h}) \frac{ q(\mbf{h}|\mbf{x})}{ q(\mbf{h}|\mbf{x})} dh \big]   \\ 
    log [p_{model}(\mbf{x})]  &= log \big[E_{q(\mbf{h}|\mbf{x})} \big[ \frac{ p(\mbf{x}|\mbf{h})   p(\mbf{h})}{q(\mbf{h}|\mbf{x})}   \big]    
\end{align}
Using Jansen inequality : $E[log(z)] \geq log E[z]$ :  
\begin{align}
    log [p_{model}(\mbf{x})] \geq E_{q(\mbf{h}|\mbf{x})} \Big[ log \big[\frac{ p(\mbf{x}|\mbf{h})   p(\mbf{h})}{q(\mbf{h}|\mbf{x})} \big] \Big]
     =  E_{q(\mbf{h}|\mbf{x})} p(\mbf{x}|\mbf{h}) +  E_{q(\mbf{h}|\mbf{x})} \frac{ p(\mbf{h})}{q(\mbf{h}|\mbf{x})} 
\end{align}
Further simplifying with the KL-divergence : 
\begin{equation}
    \label{eq:ELBO}
    log [p_{model}(\mbf{x})] \geq E_{q(\mbf{h}|\mbf{x})} p(\mbf{x}|\mbf{h}) - D_{KL}(q(h|x) \parallel q(h)) 
\end{equation}
maximizing~\refeq{ELBO} is equivalent to maximizing the likelihood of the decoder's output and minimizing the distance of the distribution $q(\mbf{h}|\mbx)$ and $p(\mbf{h})$. q is chosen to be a Gaussian distribution. This second term makes the approximate posterior distribution $q(\mbf{h} | \mbf{x})$ and the model prior p(h) approach each other.

The VAE model is both elegant, theoretically pleasing, and simple to implement. It also achieves good results and is among the leading approaches in generative modeling. However, when used in image generation, the output of the model tends to be blurry. One possibility is that the blurring is an intrinsic effect of maximum likelihood. The denoising VAE provides the base for other models, such as Diffusion models.

\subsubsection{Generative adversarial networks (GANs)}
Generative Adversarial Networks (GANs)~\cite{goodfellow2020generative} have introduced an alternative framework for training generative models that have led to a multitude of publications with high impact over a very large number of applications. The training of these models involves two networks that compete against each other. A generator that models the distribution $p(\mbf{x})$. It tries to fool the discriminator by generating samples that are as close as possible to real samples. The discriminator models $p(real|\mbf{x})$. The discriminator tries to classify the real and synthetic samples. This is formulated as a zero-sum game between two networks G and D competing to reach a Nash equilibrium. This game is commonly formulated through a min-max optimization problem as follows : 
    \begin{align}
        \min_{G\in\mathbb{G}}\max_{D\in\mathbb{D}} V(G,D) \label{eq:cost}
    \end{align}
The function $V$ determines the payoff of the discriminator. The discriminator receives $-V(G, D)$ as its own payoff, and the generator receives $V(G,D)$ as its own payoff. In other words, this formulation could be interpreted as a learned loss function as the discriminator provide the supervision signal to update the generator. A common way to supervise the discriminator is based on the cross entropy : 

\begin{align}
 \mathcal{L}_{adv}= \min_{G} \max_{D} \: \mathbb{E}_{(\mbf{x})\sim p_{data}(\mbf{x})}log [D(\mbf{x})] +  \mathbb{E}_{\mbf{z}\sim p_(\mbf{z})}log[1 - D(G(\mbf{z}))] 
 \label{eq:gan_classic_loss}  
\end{align}
At convergence, the generator’s samples are indistinguishable from real data, and the discriminator outputs $\frac{1}{2}$ everywhere. 
 Conditional GANs~\cite{mirza2014conditional}, introduced shortly after, have extended GANs to incorporate conditional information as input and have demonstrated resounding success for many computer vision tasks such as image synthesis~\cite{Isola2017ImagetoImageTW,Park2019SemanticIS,Wang2018HighResolutionIS,chen2017photographic,Sushko2020YouON,Tang2020LocalCA,Liu2019LearningTP}, video synthesis \cite{Wang2018VideotoVideoS,Chan2019EverybodyDN,Liu2019NeuralRA}, image correction\cite{Kupyn2018DeblurGANBM,Zhang2020ImageDU,Qu2019EnhancedPD}, text-to-image\cite{Reed2016GenerativeAT,zhang2018stackgan++,Xu2018AttnGANFT,Li2019ObjectDrivenTS}. In all these works, the underlying GAN model as proposed in~\cite{goodfellow2014generative} and~\cite{mirza2014conditional} have formed the basis for more advanced architectures and their properties have been analysed in detail and established in terms of convergence\cite{Kodali2018OnCA,Nie2018JRGANJR}, mode collapse\cite{Srivastava2017VEEGANRM}, Nash equilibrium\cite{Unterthiner2018CoulombGP,Farnia2020GANsMH}, vanishing gradients\cite{Arjovsky2017TowardsPM} \\\\
\textbf{Conditional GANs}

 A GAN is considered conditional~\cite{mirza2014conditional} when the generator's output is conditioned by an extra input variable $\mathbf{y}$ such that $G(\mathbf{y})\approx=p(\mathbf{x}|\mathbf{y})$ and discriminator's output is conditioned such that $D(\mathbf{x},\mathbf{y}) \approx p(\mathbf{z}|\mathbf{x,y})$ where $\mathbf{z}$ is the probability of the input being true or generated given $(\mathbf{x},\mathbf{y})$. 
The condition variable can be any kind of information such as a segmentation mask, depth map, image, or data from other modalities. In the literature there are various methods that have been proposed for incorporating conditional information into the generator including the introduction of new modules: Conditional Batch Normalization (CBN)~\cite{Vries2017ModulatingEV}, Conditional Instance Normalization (CIN)~\cite{Dumoulin2017ALR}, Class Modulated Convolution (CMConv)~\cite{Zhou2020SearchingTC}, Adaptive Instance Normalization~(AdaIN)~\cite{Huang2017ArbitraryST}, Spatial Adaptive Normalization (SPADE)~\cite{Park2019SemanticIS}. Recently,~\cite{Tang2020LocalCA} introduced a classification-based feature learning module to learn more discriminating and class-specific features.
Additional generator losses have also been proposed including feature matching~\cite{Salimans2016ImprovedTF}, perceptual loss~\cite{Johnson2016PerceptualLF}, and cycle-consistency loss~\cite{zhu2017unpaired}. All these methods propose approaches that improve the conditionality of the generator, however, they do not act on making the discriminator conditional.  

Alternatively, several methods have been proposed which investigate how to incorporate conditional information into the discriminator of adversarial networks. \cite{mirza2014conditional} proposed an early fusion approach by concatenating the condition vector to the input of the discriminator.  \cite{Miyato2018cGANsWP,Kang2020ContraGANCL,Ntavelis2020SESAMESE,Liu2019LearningTP} proposed a late fusion by encoding the conditional information and introducing it into the final layers of the discriminator. \cite{Sushko2020YouON} replaces the discriminator with a pixel-wise semantic segmentation network. Several papers improve results by adding various loss terms to the discriminator~\cite{Li2017TripleGA,Odena2017ConditionalIS,Dong2019MarginGANAT,chen2019self}, however, they don't explicitly focus on testing and constraining the conditionality of the discriminator. Similar to\cite{Miyato2018cGANsWP}, \cite{kavalerov2021multi} proposes an auxiliary classifier to the discriminator and use of Crammer-Singer multi-hinge loss to enforce conditionality. However, this method is task specific to only generation conditioned on class labels. The conditionality will be analysed further in Chapter~\ref{ch:ch3}

\subsection{Predictive models}
\label{sec:predictive-model}
Predictive models aim to model the distribution of $p(\mbf{y}|\mbf{x})$ such that $\mbf{y}$ is a low-dimensional label for the input $\mbf{x}$. Prediction, estimation, inference and forecasting, these terms are used interchangeably in the literature and depend on the context of the task. However, a more formal definition to these terms is provided here: 

\begin{itemize}
    \item \textbf{Prediction}: as mentioned earlier, predictive models aim to model the distribution of $p(\mbf{y}|\mbf{x};\bm{\theta})$. For a new measurement, $\mbf{x}_{new}$, these models “predicts” a new $\mbf{y}_{new}$. Prediction could be related to time such as predicting the $y_{t+n}$ based on $x_t$ that is modeling, $p(\mbf{y_{t+n}}|\mbf{x_t};\bm{\theta})$ or not related to time such as predicting the bounding box of an object that is modeling $p(\mbf{b}|\mbf{x};\bm{\theta})$ and $\mbf{b}$ is the bounding box of an object present in the image $\mbf{x}$.
    
    \item \textbf{Inference:} This term is often used in the literature to determine the latent variables that generate the observed data $p(\mbf{h}|\mbf{x})$. In the deep learning community, the term inference and prediction are interchangeable~\cite{murphy2022probabilistic}, Both aim to model a distribution of some variable $\mby$ given some other variable $\mbx$ \ie $p(\mbf{y}|\mbf{x};\bm{\theta})$ but they focus on different aspects of this process. Inference involves determining the distribution of latent variables based on the observed data. One common method for doing this is Bayesian inference or using approximate inference. While prediction is used to determine the actual output $\mbf{y}$ of the predictive model. 

    \item \textbf{Forecasting: }it is a specialized formulation of prediction that is concerned with predicting the future value 
    $\mbf{y_{t+n}}$ based on the present and the past measurements $\mbf{x_{t-k:t}}$. It models the distribution $p(\mbf{y_{t+n}}|\mbf{x_{t-k:t}})$

    \item \textbf{Estimation}: Kent \ea~\cite{kent2001estimation} defines the process of estimation as “using the value of a statistic derived from a sample to estimate the value of a corresponding population parameter”. In the context of deep learning, it involves finding the model parameters that minimize the error between the model's predictions and the true values of the output variables.
\end{itemize}

\section{Depth prediction}
\label{sec:background-depth}
The ability to perceive the 3D structure of the world is a fundamental aspect of human vision. It allows us to perceive the distance and relative position of objects in the world, as well as to judge the size, shape, and orientation of objects. One common use is in the field of robotics is autonomous driving (AD) and advanced driver-assistance system (ADAS), depth perception is used for path planing, automatic emergency braking (AEB), automatic cruise control (ACC) and many more. 

The human brain is able to recognize the depth of the objects based on several clues that are present in the visual scene such as disparity, motion parallax, perspective, memorization, and shading. For example, the lateral separation of the eyes enables identifying an object from two angles, The brain uses this difference to compute the relative disparity. Motion parallax is the relative motion of objects at different distances as the viewer moves or the camera pans. The brain uses this motion to infer the depth of objects. 


\subsection{Camera versus LiDAR for autonomous driving}
The use of cameras and LiDAR in autonomous driving is an ongoing topic of discussion among researchers and engineers. Although both technologies have their own advantages and disadvantages, the cameras are generally considered to be more affordable and versatile, while LiDAR is known for its superior accuracy and range.

LiDAR (Light Detection and Ranging) is an active 3D sensor that emits light to measure the distance to objects and provide a high-resolution 3D representation of the environment. LiDAR has become an increasingly popular technology in a wide range of applications such as autonomous vehicles due to its several advantages.

\begin{itemize}
    \item One of the main advantages of LiDAR is its high accuracy. Lidar can acquire high-precision data with a resolution of a few centimeters, making it suitable for applications that require precise measurements. In addition, LiDAR has a long range and can detect objects several kilometers away. This makes it useful in applications such as self-driving cars, localizing and mapping. Another advantage of LiDAR is its 3D imaging capabilities. LiDAR can produce real-time 3D point clouds of the environment, allowing for the creation of detailed 3D maps and models. Furthermore, LiDAR as active sensor can operate in any poor lighting conditions, making it useful for both day and nighttime applications. However, despite its advantages, LiDAR has several disadvantages, the most notable being the cost, which can range from \euro$ 4,000$ to \euro$80,000$, making it less cost-effective than cameras. Additionally, LiDAR systems are weather sensitive and can be affected by conditions such as fog or rain, which can affect their accuracy and effectiveness. In addition, LiDAR systems consume a significant amount of power and are typically larger and heavier than cameras.

\end{itemize}

On the other hand, cameras are passive sensor that detect and measure the intensity of electromagnetic radiation, such as light. it consists of a lens to direct the incoming light and a light sensor to measure the intensity. Cameras have several advantages compared to LiDAR : 
\begin{itemize}
    \item The main advantages of cameras is their cost-effectiveness. They are widely available at a lower cost compared to LiDAR systems. Given the small margins in the automotive industry, the use of LiDAR is not practical and cameras are a more cost-effective sensor option. Cameras also have the added benefit of being able to perform multiple tasks, such as object detection, lane segmentation, and traffic sign recognition, making them a more attractive option for self-driving vehicles. Despite its advantages, cameras have some other limitations, including lighting conditions. Cameras can be affected by poor lighting conditions, which can make it harder to obtain accurate images. They also have a limited range, which means they can only detect objects within a certain distance. Using cameras for mapping can raise privacy concerns, particularly when they are used in public spaces.
\end{itemize}
Fusing LiDAR and camera modalities has proven to be an extremely effective approach for autonomous vehicles and other applications. The LiDAR sensor provides precise 3D spatial information, while the cameras provide high-resolution visual data. By using data from both modalities, it is possible to improve the accuracy and robustness of object detection, location, and tracking. In particular, LiDAR can provide accurate depth information and help detect objects in low light conditions or obscured by other objects. On the other hand, cameras can provide detailed visual information that can be used to detect objects and improve their identification. Furthermore, integrating both modalities reduces the disadvantages of each individual sensor, such as using LiDAR to detect objects in poor weather conditions and cameras to detect objects in optimal weather conditions. In summary, the fusion of LiDAR and camera data provides a more complete and accurate view of the environment, enabling the development of safer and more reliable autonomous systems.

\subsection{Deep learning for depth prediction}
In recent years, deep learning has emerged as a powerful tool in the field of depth prediction. This is due in large part to the success of convolutional neural networks (CNNs), which have surpassed other classical methods that rely on hand-crafted features. One of the key advantages of deep learning is its ability to learn features directly from data. This allows the model to automatically adapt to the task at hand and model complex situations, making it more powerful tool than traditional methods that relied on hand-crafted features.
Furthermore, CNNs have a hierarchical structure, which allows them to learn features at different levels of abstraction. This hierarchical structure is especially useful for depth prediction, as it allows the model to learn both low-level features, such as edges and textures, as well as high-level features, such as object shape and context.

\subsubsection{Depth prediction taxonomy}
\begin{figure}
    \centering
    \includegraphics[width=\textwidth]{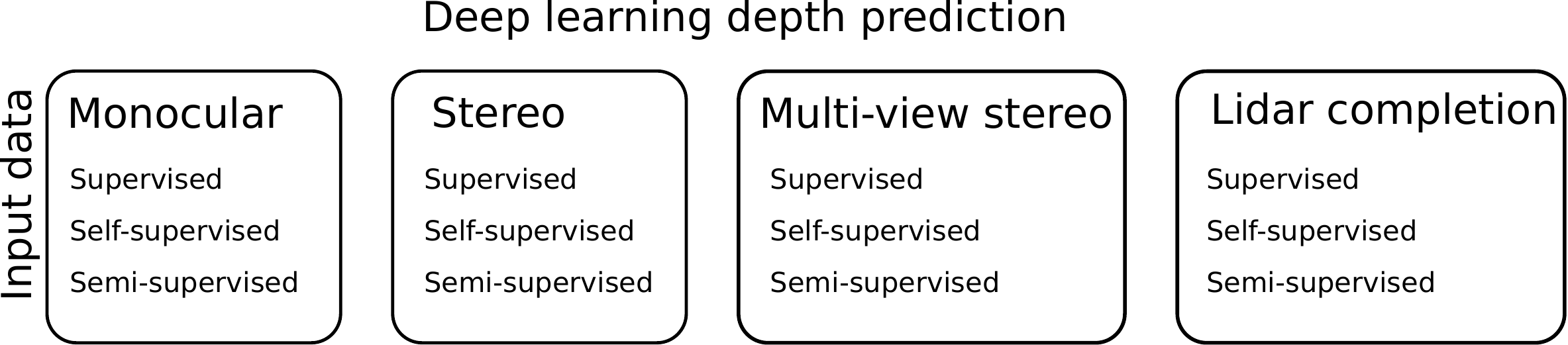}
    \caption{Taxonomy of deep learning methods for depth prediction}
    \label{fig:taxonomy-depth}
\end{figure}

Depth prediction is a challenging task that has been widely studied in the field of computer vision. One way to organize the various methods and models for depth prediction is through a taxonomy based on the input data or the supervision used \reffig{taxonomy-depth}.
\subsubsection{Classification based on input data}
One way to classify the methods for depth prediction is based on the input data used. This includes:

\begin{itemize}
    \item \textbf{Monocular depth prediction:} uses a single camera and predict the depth information from it.
    \item \textbf{Stereo depth prediction:} uses two cameras to capture the scene from different viewpoints and predict the depth information.
    
    \item \textbf{Multi-view depth prediction:} use multiple images captured from different viewpoints as input and estimate depth information by combining information from all the images. These methods have been shown to be effective in improving the performance of depth prediction on large-scale datasets. Multi-view depth prediction methods benefit from the strong epipolar geometry prior. This constraint provides additional information that can be used to improve the accuracy of depth prediction.

    \item \textbf{LiDAR's completion:} LiDAR completion is a method that uses a partial depth map acquired by a LiDAR sensor as input and infers the missing depth information. LiDAR's sensors are known for their high accuracy and resolution, which makes them suitable for depth prediction. LiDAR's completion aim to generate a dense depth map from the sparse LiDAR map using camera information. 

\end{itemize}

Deep learning models perform best when provided with more information. Among the different methods for depth prediction, LiDAR completion has the lowest entropy (\ie provide the lowest uncertainty), as it already provides a sparse and accurate depth map that only needs to be completed. Multi-view systems come in second, as this method leverages more information and have a strong epipolar geometry constraint. On the other hand, monocular depth prediction has the highest entropy. This is because extracting 3D information from a single 2D image is an ill-posed problem. The scale of the objects in the scene is not known. Monocular depth prediction methods have to rely on other cues, such as texture, color, and motion, to estimate depth. These methods have the highest uncertainty and are prone to the scale ambiguity problem. However, the performance gap of these methods is closing and the prevalence of monocular cameras is making their applications more accessible and interesting. Another aspect to consider is the application: a monocular camera is found on all devices nowadays. Performing depth on these devices is more interesting as it enables a wide range of applications.

\subsubsection{Classification based on learning}
Another way to classify depth prediction methods is based on the type of supervision used during training. This includes:

\begin{itemize}
    \item \textbf{Supervised depth prediction:} uses ground truth depth maps as the supervision during training. Supervised learning is the best-performing method for depth prediction in the benchmark, as it uses acquired depth maps obtained from depth sensors as supervision for the network. Assuming good accuracy of the ground-truth provided by depth sensors, these models have the best performance as the information provided for supervision is accurate and reliable. However, the use of ground-truth depth maps for supervision is not always feasible, as they can be expensive to acquire or unavailable. Furthermore, as the amount of the available labeled data is very limited, the generalization of these models are not guaranteed when deployed in complex environment.

    \item \textbf{Self-supervised depth prediction:} these methods have been proposed as an alternative to supervised methods. These methods use a differentiable warping to reconstruct a set of a source images (monocular video, stereo video …etc.) into a target images. This is done using a depth network and pose a network when the pose is not known (\eg the case of monocular video). These models are very practical as they can be easily trained, as they only require a video as input. The data is cheap to collect and widely available, as billions of videos are already available on platforms such as YouTube.

    \item \textbf{Semi-supervised depth prediction:}  proposed as a solution that combines the advantages of both supervised and self-supervised methods. These models use a combination of ground-truth depth maps and other forms of supervision, such as depth estimates from self-supervised methods or depth priors. These methods have been shown to be effective in improving the performance of depth prediction on large-scale datasets, while also being more practical as they do not require ground-truth depth maps for all the data.
\end{itemize}

\subsection{Self-supervised monocular depth}
The focus of this thesis is on the self-supervised learning (SSL) of monocular image or video to depth of the corresponding scene. In this section, a brief introduction is presented on the pipeline of the SSL depth. Further development of the warping, the objective, and training will be done in Chapter~\ref{ch:ch4}.

\begin{figure}
    \centering
    \includegraphics[width=\textwidth]{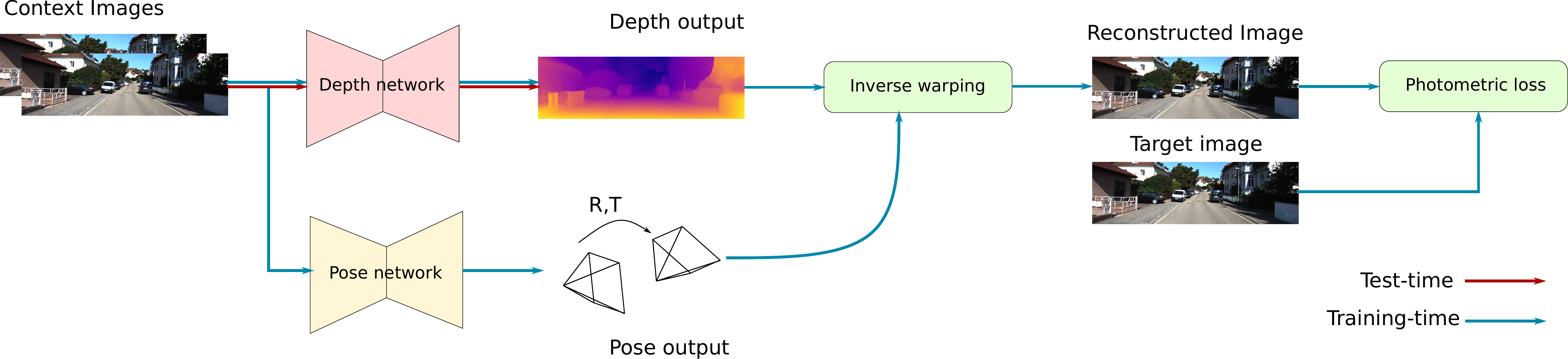}
    \caption{Training-time versus test-time for self-supervised depth prediction. The red arrow shows the test-time pipeline, while the blue arrows show the training-time pipeline. The warping function and the pose network are used only during training.}
    \label{fig:training-vs-test}
\end{figure}

\reffig{training-vs-test} Shows the pipeline for self-supervised monocular depth prediction. It composes two networks, a depth network and a pose network. During training, the forward pass is propagated through the depth network and the pose network. The output of these networks will then be used to reverse warp a source image to reconstruct the target image. The photometric loss is calculated and the backward pass optimize both depth and pose network parameter simultaneously. At test-time, \textbf{only the depth network }is used and the pose network, unless used to reconstruct the image, it is discarded.

The SSL depth pipeline comprises several crucial elements, including the depth network, the pose network, and the photometric loss. The depth network leverages convolutional neural networks (CNNs) or other deep learning architectures to learn the intricate relationships between pixel intensities and depth values. The pose network, on the other hand, is responsible for estimating the camera's motion or pose between different frames in a video sequence. By accurately determining the camera's movement, the pose network aids in aligning the source and target images during the reconstruction process.

\subsection{Limitations}

Self-supervised learning (SSL) methods for monocular depth prediction have shown promise, but it is important to acknowledge their limitations. In this section, we discuss some of these limitations:

\begin{itemize}
    \item \textbf{Reliance on the Assumption of Photometric Consistency}: SSL methods heavily rely on the assumption that pixel intensities remain consistent across different views of the same scene, which may not hold true in challenging real-world scenarios. Variations in lighting conditions, occlusions, this assumption and lead to inaccurate depth predictions. Several methods were proposed to account for this problem.
    
    \item \textbf{Requirement for Large Amounts of Training Data}: SSL methods often demand a substantial amount of unlabeled data for effective training. While unlabeled data is typically abundant and easy to obtain, it can be challenging to ensure its quality and diversity. Inadequate or biased training data may result in suboptimal depth estimation performance and generalization to real-world scenarios.
        
    \item \textbf{Limitation of Monocular Input}: The reliance on monocular input limits the accuracy of depth estimation compared to methods that utilize stereo or multi-view setups. Monocular depth estimation inherently suffers from scale ambiguity, as a single 2D image cannot provide sufficient information to uniquely determine 3D values.
    
    \item \textbf{Rigid Scene Assumption}: The vast majority of self-supervised learning methods for monocular depth estimation assume a rigid scene, meaning they do not handle dynamic objects. Moving objects within the scene can disrupt the consistency assumption and introduce errors in depth estimation. This limitation restricts the applicability of SSL methods in scenarios with significant object motion or dynamic environments.
\end{itemize}

\subsection{Related work}
Depth prediction has been successful with self-supervised learning from videos. The seminal work of Zhou \ea~\cite{Zhou2017} introduced the core idea to jointly optimize the pose and depth network using image reconstruction and a photometric loss.  Due to its simplicity and generality, this approach has attracted significant attention from researchers, leading to a series of related works, including~\cite{Vijayanarasimhan,xu2021moving,luo2019every,chen2019self,Bian2019,Chen2019,Rares2020,Wang2021, mccraith2020monocular, Watson2021}. 

Recognizing the inherent challenges associated with the ill-posed nature of depth prediction, researchers have endeavored to tackle various aspects. Addressing the rigid scene assumption, Vijayanarasimhan et al.~\cite{Vijayanarasimhan}, Xu et al.~\cite{xu2021moving}, and Luo et al.~\cite{luo2019every} employed optical flow and motion clustering techniques to overcome this limitation. Chen et al.~\cite{chen2019self}, on the other hand, focused on enhancing generalization by incorporating camera parameter learning into the framework. Meanwhile, Bian et al.~\cite{Bian2019}, Chen et al.~\cite{Chen2019}, Rares et al.~\cite{Rares2020}, and Wang et al.~\cite{Wang2021} directed their efforts towards mitigating the scale ambiguity problem, proposing innovative approaches that enforce depth scale and structure consistency. Furthermore, to enhance the performance of depth prediction models during inference, McCraith et al.~\cite{mccraith2020monocular} and Watson et al.~\cite{Watson2021} introduced test-time refinement strategies. These techniques allow for dynamic variation of model parameters using a photometric loss, thereby refining the depth estimation results.

In summary, this thesis explores the advancements in the field of depth prediction through self-supervised learning from videos. It addresses the problem of rigid scene assumption with a novel method~\cite{boulahbal2022instance} presented in Chapter~\ref{ch:ch4}. Furthermore, the framework is extended to enable future forecasting~\cite{boulahbal2022forecasting}, as discussed in Chapter~\ref{ch:ch5}. Finally, a novel method is introduced in Chapter~\ref{ch:ch6}, which outputs a sequence of future depth predictions~\cite{boulahbal2023stdepthformer}. 

\subsection{Depth evaluation}
\label{sec:depth-eval}

The evaluation of depth estimation methods often relies on the KITTI benchmark~\cite{Geiger2012CVPR}, which is widely used in the field. The KITTI dataset provides a set of sequence captured from a car-mounted camera, accompanied by accurate depth maps obtained using LiDAR sensors. The Eigen \ea~\cite{Eigen} train/validation split is widly used to train and evaluate the models. To evaluate the quality of depth estimation results, several metrics are commonly employed. These metrics are defined in ~\reftab{depth_metrics}

\begin{table}[h]
\centering
\begin{tabular}{lccc}
\hline
\textbf{Metric} & \textbf{Formula} & \textbf{Range} & \textbf{Description} \\ \hline
RMSE            & $\sqrt{\frac{1}{N}\sum_{i=1}^{N}(d_i - \hat{d}_i)^2}$ & $[0, \infty)$ & Root Mean Squared Error \\
Abs rel         & $\frac{1}{N}\sum_{i=1}^{N}\frac{|d_i - \hat{d}_i|}{d_i}$ & $[0, \infty)$ & Absolute Relative Difference \\
Sq rel          & $\frac{1}{N}\sum_{i=1}^{N}\frac{\sqrt{(d_i - \hat{d}_i)^2}}{d_i}$ & $[0, \infty)$ & Squared Relative Difference \\
Log RMSE        & $\sqrt{\frac{1}{N}\sum_{i=1}^{N}(\log(d_i) - \log(\hat{d}_i))^2}$ & $[0, \infty)$ & Logarithmic Root Mean Squared Error \\
$\delta<1.25$   & $\frac{1}{N}\sum_{i=1}^{N}\mathbb{I}(\max(\frac{d_i}{\hat{d}_i}, \frac{\hat{d}_i}{d_i}) < 1.25)$ & $[0,1]$ & $\%$ of pixels with $\delta<1.25$ \\
$\delta<1.25^2$ & $\frac{1}{N}\sum_{i=1}^{N}\mathbb{I}(\max(\frac{d_i}{\hat{d}_i}, \frac{\hat{d}_i}{d_i}) < 1.25^2)$ & $[0,1]$ & $\%$ of pixels with $\delta<1.25^2$ \\
$\delta<1.25^3$ & $\frac{1}{N}\sum_{i=1}^{N}\mathbb{I}(\max(\frac{d_i}{\hat{d}_i}, \frac{\hat{d}_i}{d_i}) < 1.25^3)$ & $[0,1]$ & $\%$ of pixels with $\delta<1.25^3$ \\ \hline
\end{tabular}
\caption{Depth Benchmark Evaluation Metrics}
\label{tab:depth_metrics}
\end{table}


\section{Style transfer using conditional GANs}

Style transfer is a fascinating technique that allows the transformation of the visual style of an image while preserving its underlying content. Conditional generative adversarial networks (cGANs) have emerged as a powerful approach for achieving style transfer in an automated and data-driven manner. By training GANs on large datasets of paired images with different styles, these models can learn to capture the essence of each style and apply it to new images. The following section delves into two notable examples of conditional GANs for style transfer: Pix2Pix and CycleGAN. These two networks will be used extensively in Chapter~\ref{ch:ch3}.

\subsection{Pix2Pix for paired datasets}
This section provides a brief introduction to this architecture. Additional information can be found in~\cite{Isola2017ImagetoImageTW}. 
Pix2Pix, introduced by Isola~\ea~\cite{Isola2017ImagetoImageTW} is a pioneering architecture that showcases the potential of conditional GANs in image-to-image translation tasks. It provides a framework for translating images from a source domain to a target domain while maintaining their content. Pix2Pix's strength lies in its ability to bridge the gap between domains by learning the mapping between paired images.

Generative Adversarial Networks (GANs) are models designed to generate realistic images by learning a mapping from a random noise vector $(\mbf{z})$ to an output image $(\mbf{y})$, denoted as $\mbf{y}=G(\mbf{z})$. However, conditional GANs take it a step further by learning a mapping from both an observed image $(\mbf{x})$ and a random noise vector $(\mbf{z})$ to the output image $(\mbf{y})$, expressed as $\mbf{y}=G(\mbf{x}, \mbf{z})$. The main objective is for the generator (G) to produce outputs that are indistinguishable from authentic images, thereby fooling a discriminator (D) trained to identify the generator's fakes. 

\begin{figure}
    \centering
    \includegraphics[width=\textwidth]{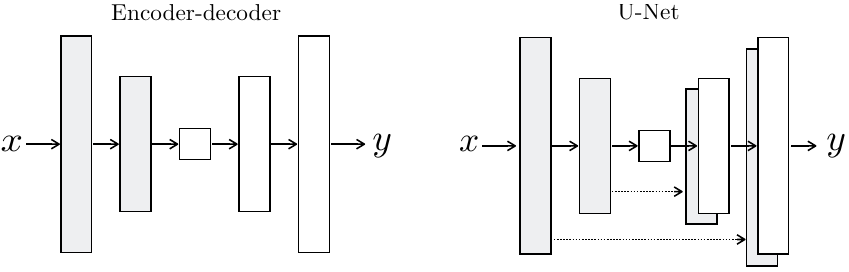}
    \caption{The Pix2Pix architecture with encoder-decoder structure and skip connections, inspired by the U-Net architecture. The encoder captures essential features from the source image, while the decoder generates the corresponding output image in the target domain. The inclusion of skip connections helps preserve details, leading to high-quality outputs. Figure from~\cite{Isola2017ImagetoImageTW}}.
    \label{fig:pix2pix_gen}
\end{figure}

As shown in~\reffig{pix2pix_gen}, The Pix2Pix architecture employs an encoder-decoder structure with skip connections, inspired by the U-Net architecture. The encoder captures essential features from the source image, while the decoder generates the corresponding output image in the target domain. The skip connections aid in preserving details during the translation process, resulting in high-quality outputs.

\begin{figure}
    \centering
    \includegraphics[width=\textwidth]{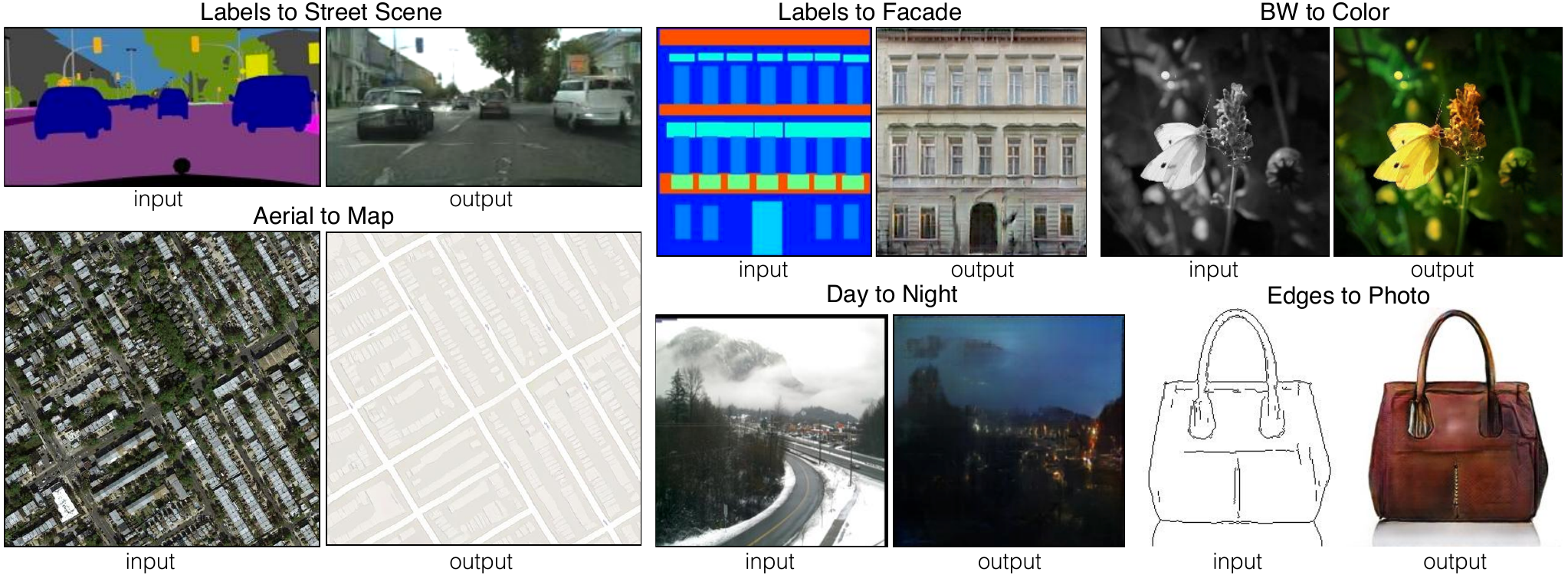}
    \caption{The figure displays a curated set of images demonstrating the effectiveness of Pix2Pix in transforming images from the source domain to the target domain while preserving essential details and structural integrity. Figure from~\cite{Isola2017ImagetoImageTW}}
    \label{fig:pix2pix-res}
\end{figure}

\subsubsection{Objective}

The objective of a conditional GAN can be expressed as
\begin{align}
    \mathcal{L}_{cGAN}(G,D) = &\mathbb{E}_{x,y}[\log D(x,y)] + \nonumber \\
                 &\mathbb{E}_{x,z}[\log (1-D(x,G(x,z))],
    \label{cGAN_equation}
\end{align}
where $G$ is the generator that tries to minimize this objective against a discriminator $D$ that tries to maximize it, i.e. $G^*  = \arg\min_G \max_D \mathcal{L}_{cGAN}(G,D)$.

Previous approaches have found it beneficial to mix the GAN objective with a more traditional loss, such as L2 distance \cite{pathak2016context}. The discriminator's job remains unchanged, but the generator is tasked to not only fool the discriminator, but also to be near the ground truth output in an L1 sense. This will be explored further in Chapter~\ref{ch:ch3} when the conditionality of cGAN is analyzed
\begin{align}
    \label{L1_equation}
    \mathcal{L}_{L1}(G) = \mathbb{E}_{x,y,z}|y-G(x,z)|.
\end{align}

The final objective is
\begin{align}
    \label{full_objective}
    G^*  = \arg\min_G\max_D \mathcal{L}_{cGAN}(G,D) + \lambda \mathcal{L}_{L1}(G).
\end{align}

Pix2Pix requires a paired dataset for training. In other words, it relies on having images from the source domain and their corresponding images in the target domain. These paired images serve as the training data, enabling Pix2Pix to learn the mapping between the two domains. \reffig{pix2pix-res} shows the qualitative results of the style transfer of Pix2Pix and showcases a selection of images that have undergone style transfer. It illustrates how the model successfully transforms images from the source domain into the target domain while preserving important details and structure.

This basic architecture will be used to study the conditionality and based on that analysis a novel method will be proposed to improve the conditionality of cGANs.

\subsection{CycleGAN for unpaired datasets}
This section provides a brief overview of CycleGAN~\cite{zhu2017unpaired}, a framework for unpaired image-to-image translation. While traditional methods like Pix2Pix require paired images for training, CycleGAN addresses this limitation by enabling style transfer between domains without the need for explicit pairs of corresponding images.

\begin{figure}
    \centering
    \includegraphics[width=0.4\textwidth]{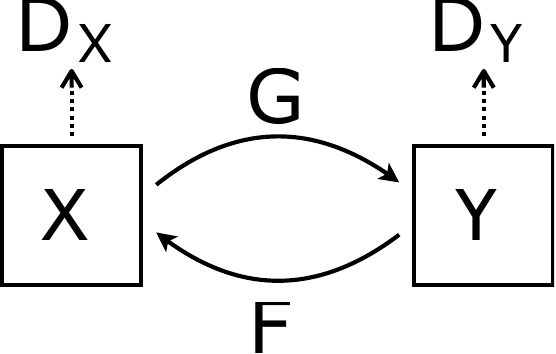}
    \caption{CycleGAN pipeline. The figure illustrates two generator networks, $G: X \rightarrow Y$ and $F: Y \rightarrow X$, along with two discriminator networks, $D_x$ and $D_y$. The generators learn to translate images between domains, while the discriminators distinguish between translated and real images. Figure adopted from~\cite{zhu2017unpaired}}
    \label{fig:cyclegan_arch}
\end{figure}

As shown in~\reffig{cyclegan_arch}, CycleGAN consists of two generator networks, namely the generator $G: X \rightarrow Y$ and the reverse generator $F: Y \rightarrow X$, along with two discriminator networks, $D_x$ and $D_y$. The generators learn to translate images from one domain to another and back, while the discriminators aim to distinguish between the translated images and real images from each domain. To ensure content preservation during the translation process, CycleGAN incorporates a cycle consistency loss, which enforces that the reconstructed image should resemble the original image.

The objective of CycleGAN involves adversarial losses~\cite{goodfellow2014generative} for both mapping functions. For the mapping function $G: X \rightarrow Y$ and its discriminator $D_Y$, the objective is expressed as:

\begin{align}
\label{eq:GAN_}
\mathcal{L}_{\text{GAN}}(G,D_Y,X,Y) = & \ \mathbb{E}_{y \sim p_{\text{data}}(y)}[\log D_Y(y)] \nonumber \
+\mathbb{E}_{x \sim p_{\text{data}}(x)}[\log (1-D_Y(G(x)))],
\end{align}

where $G$ generates images $G(x)$ that resemble images from domain $Y$, and $D_Y$ distinguishes between translated samples $G(x)$ and real samples $y$. A similar adversarial loss is introduced for the mapping function $F: Y \rightarrow X$ and its discriminator $D_X$ as well.

However, adversarial losses alone cannot guarantee that the learned function can accurately map an individual input to a desired output. To address this, CycleGAN introduces cycle consistency, which ensures that the learned mapping functions are cycle-consistent. This means that for each image $x$ from domain $X$, the image translation cycle should be able to bring $x$ back to the original image ($x \rightarrow G(x) \rightarrow F(G(x)) \approx x$), and vice versa for images from domain $Y$. This behavior is incentivized using a cycle consistency loss:

\begin{align}
\mathcal{L}_{\text{cyc}}(G, F) =  \mathbb{E}_{x\sim p_{\text{data}}(x)}|F(G(x))-x| 
+ \mathbb{E}_{y\sim p_{\text{data}}(y)} |G(F(y))-y|.
\end{align}

The full objective of CycleGAN combines the adversarial losses and the cycle consistency loss:

\begin{align}
\mathcal{L}_(G,F,D_X,D_Y) = \mathcal{L}_{\text{GAN}}(G,D_Y,X,Y) +
 \mathcal{L}_{\text{GAN}}(F,D_X,Y,X) +
+ \lambda \mathcal{L}_{\text{cyc}}(G, F),
\label{eq:full_objective_gan}
\end{align}

where $\lambda$ controls the relative importance of the two objectives. The aim is to solve the optimization problem:

\begin{equation}
G,F = \arg\min_{G,F}\max_{D_x,D_Y} \mathcal{L}(G, F, D_X, D_Y).
\label{eq:cycle_gan}
\end{equation}

\begin{figure}
    \centering
    \includegraphics[width=\textwidth]{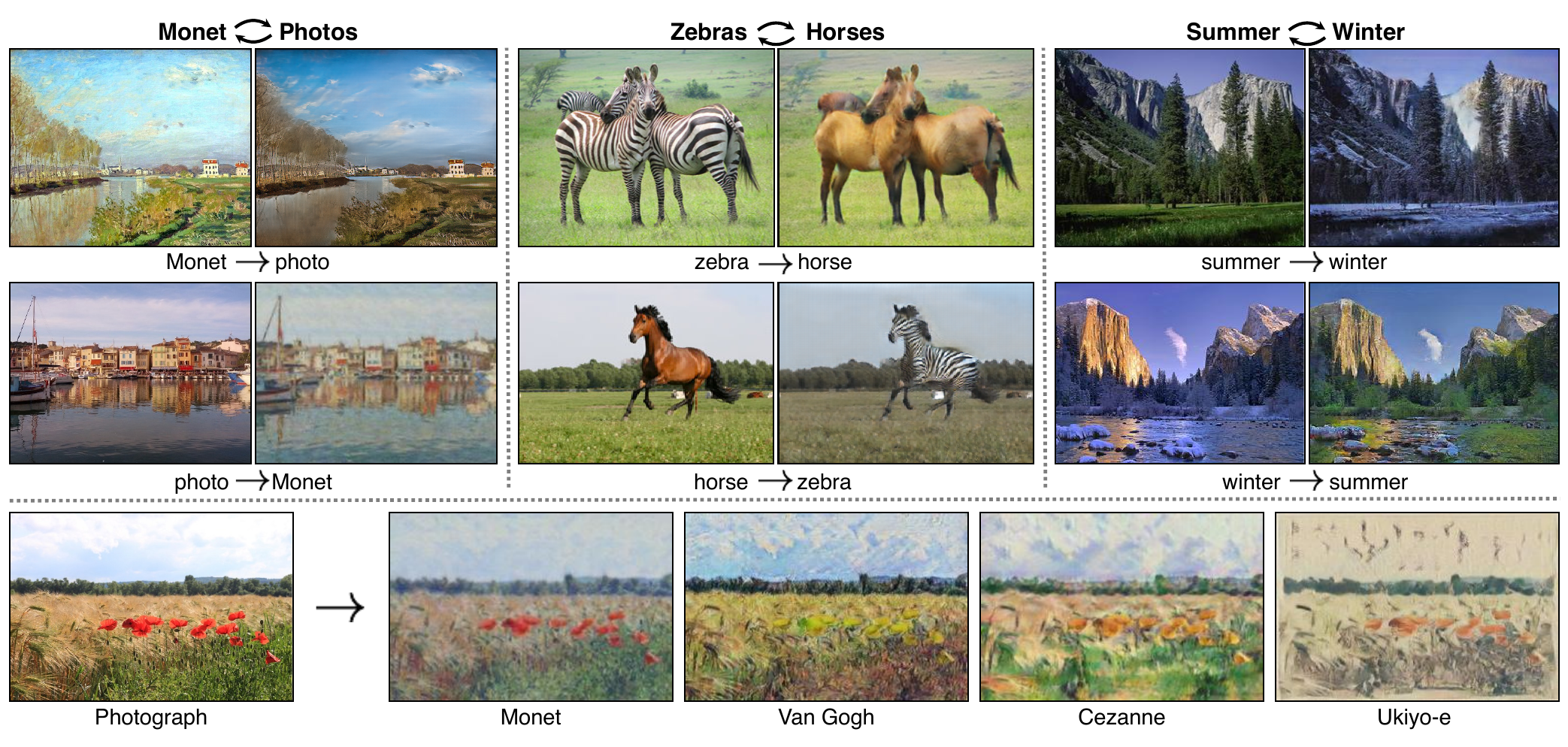}
    \caption{ Qualitative results achieved by CycleGAN in various translation settings. The images demonstrate the effective conversion from one domain to another while preserving essential visual characteristics. CycleGAN exhibits versatility in handling diverse types of translations, maintaining the original input's structure and content while incorporating distinct attributes of the target domain.}
    \label{fig:qualitative_cyclegan}
\end{figure}

\reffig{qualitative_cyclegan} illustrates the qualitative results achieved by CycleGAN in various translation settings, demonstrating its ability to effectively convert images from one domain to another while maintaining crucial visual characteristics. These results showcase CycleGAN's versatility in handling different types of translations. The translated images successfully retain the overall structure and content of the original input, while incorporating the unique attributes of the target domain.

In this thesis, the exploration of Generative Adversarial Networks (GANs) began in the early stages of 2020. During that time, GANs gained popularity for their remarkable ability to perform style transfer and generate visually appealing images. However, since then, more recent methods have emerged that exhibit even more impressive results, surpassing the capabilities of GANs. One such method is the employment of diffusion models~\cite{rombach2022high,ramesh2021zero}. These models have demonstrated superior performance in various image generation tasks. Nonetheless, the focus of this thesis primarily revolves around depth prediction rather than image generation. Therefore, while these advanced techniques like diffusion models exist, they fall beyond the scope of this particular study.

\section{Image segmentation}
\label{sec:eval_metric_seg}
The utilization of segmentation representation in Chapter~\ref{ch:ch3} and Chapter~\ref{ch:ch4} of this thesis indicates its significance, thereby suggesting the need to expand upon the background information.
Intelligent systems require the ability to reason about their environment in order to make accurate decisions. However, the raw representation of image intensity (i.e., a pixel value matrix) is not adapted to this goal. A more abstract representation is required, where each pixel encodes a more abstract information rather than an intensity value. This representation is known as segmentation. Image segmentation involves dividing an image into distinct meaningful domains, transforming it into a simplified, high-level representation. This type of representation has received a lot of attention in recent years due to its wide range of applications, including autonomous driving, medical imaging, etc. There are three main types of segmentation: semantic segmentation, instance segmentation and panoptic segmentation. 

\subsection{Semantic segmentation}
Semantic segmentation is a form of image segmentation that assigns class labels to each pixel within an image, where these class labels correspond to objects or regions present within the image. For instance, in a semantic segmentation of a city street, pixels associated with the road, buildings, trees, sky, and other objects would be assigned different class labels. This approach to segmentation is valuable for tasks such as scene understanding, object detection, and scene labeling. 

In recent years, there has been substantial progress in the field of semantic segmentation. Classical methods were based on hand-crafted features. These models are limited by the representation as it does not fully capture high-level and low-level relations, thus limiting their performance. With the recent advance of deep learning methods, researchers have extended such methods to semantic segmentation. (FCNs)~\cite{long2015fully}, revolutionized the field of semantic segmentation. It is an encoder-decoder architecture where the encoder is based on the VGG-16~\cite{simonyan2014very} architecture, and the decoder consists of convolution and transposed convolution layers. The subsequent SegNet~\cite{badrinarayanan2017segnet} architecture introduced novel layers for upsampling in place of transposed convolutions, while ParseNet~\cite{liu2015parsenet} modeled global context directly. The PSPNet~\cite{zhao2017pyramid} architecture focused on multi-scale features, proposing pyramid pooling to learn feature representations at different scales. DeepLabV3+~\cite{chen2018encoder} proposed the Atrous Spatial Pyramid Pooling (ASPP) module to improve the receptive field of the backbone.  

With the advent of transformers and visual transformer (ViT)~\cite{Vaswani}, several methods were proposed.~\cite{carion2020end} proposed a hybrid architecture with a CNN based backbone and a transformer encoder-decoder to perform as semantic segmentation decoder. (ViT)~\cite{dosovitskiy2020} has proposed the first end-to-end backbone transformer based model for segmentation. Swin transformers~\cite{liu2021swin} made it possible to process high resolution images efficiently. With the success of multi-modalites pre-training on video audio and text, semantic segmentation have benefitted from this advancement. ~\cite{su2022towards} combined all the pre-training paradigm, including supervised pre-training, weakly-supervised and self-supervised resulting in an state-of-art results on ADE20K benchmark~\cite{zhou2017scene}.~\cite{chen2022vision} pretrained a model on image, text and used an adapter to introduce the image-related inductive biases into the model. This representation is used to validate domain adaptation in chapter~\ref{ch:ch3}.

\subsection{Instance segmentation}
Instance segmentation aims to recognizing individual instances of objects within the image. This is particularly useful for tasks like object tracking and counting, or for identifying specific instances of objects for further analysis. For instance, in an instance of segmentation of a street scene, individual cars and bicycles would be distinguished and separated from one another.

\begin{figure}
    \centering
    \includegraphics[width=0.75\textwidth]{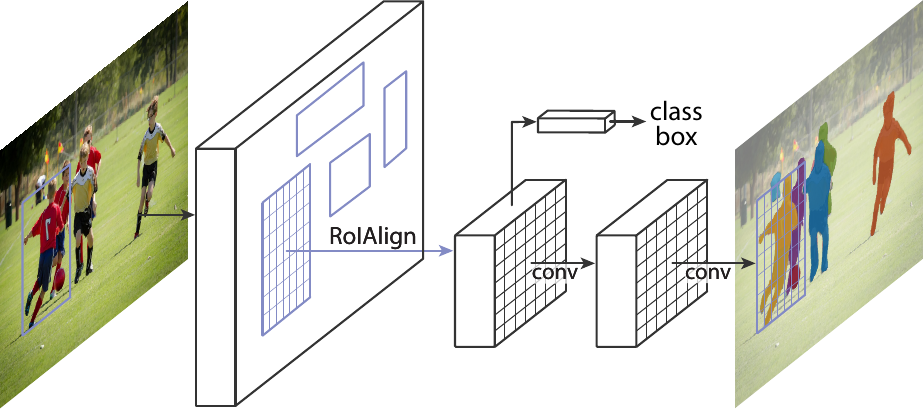}
    \caption{Overview of Mask-RCNN architecture. Similar to ~\cite{ren2015faster}, after extracting the features of an object proposal from image, the RoIAlign module is used to pool the features of tha object. 3 Parallel heads are used after, the bounding box regression head, the class head, and the mask head. Figure adopted from~\cite{he2017mask}}
    \label{fig:maskrcnn}
\end{figure}

Mask R-CNN~\cite{he2017mask} is one of the most popular models for.~\reffig{maskrcnn} Shows the architecture of the mask-RCNN model. Similar to~\cite{ren2015faster}, this model consists of a backbone that extracts the features from the image. Typically, a ResNet\cite{chen2018encoder}. In the first stage, A $RPN$ network uses these features to propose $N$ possible object in the scene. For the second stage, This method proposed a~\textit{RoIAlign} module that pools the features with better alignment to the input, and incorporates an additional object segmentation branch, in parallel to the bounding box regression and classification branch of ~\cite{ren2015faster}. Further advancements have been made with proposal based networks~\cite{liu2018path,li2017fully,chen2020blendmask}, single stage network proposal free-networks~\cite{bolya2019yolact,fu2019retinamask,lee2020centermask}, and transformers based networks~\cite{chen2021simple,chen2021simple,guo2021sotr}. This network is extended in Chapter~\ref{ch:ch4} with the object pose.

\subsection{Panoptic segmentation}
Panoptic segmentation combines semantic segmentation with instance segmentation. This task is represented with a foreground/background classes. The fourground ‘thing’ class represents countable categories in the real world, such as people, cars. Each of these objects is assigned a unique identifier along the object mask. The background ‘stuff’ class represents categories that cannot be counted, such as road and wall. It was shown that combining these two tasks improves the performance for both tasks. 
The introduction of panoptic segmentation was first presented in the seminal work of Kirillov \ea~\cite{kirillov2019panoptic}. The authors of this work formulated the task, established evaluation parameters, and presented a basic baseline. Since then, the task has received considerable attention in the research community, resulting in numerous techniques and approaches~\cite{mohan2021efficientps,kirillov2019panoptic,chen2020banet,chen2020spatialflow,wang2020axial,chen2020panonet}.

EfficientPS~\cite{mohan2021efficientps} is a panoptic segmentation networj that improves upon EfficientNet~\cite{tan2019efficientnet} by proposing 2-way FPN that both encodes and aggregates semantically rich multiscale features in a top-down and bottom-up way. This model consists of a shared backbone, EfficientNet~\cite{tan2019efficientnet}, the complexity of this model could be controlled. The two-way FPN that aggregates the features in a top-down and bottom-up way, resulting in a semantically rich multiscale features. The semantic segmentation head uses the output of the FPN to perform segmentation using DPC and modules~\cite{mohan2021efficientps}. The instance information is extracted using a Mask-RCNN~\cite{he2017mask} head. And finally, the panoptic fusion head fuses the semantic and instance segmentations and outputs the panotic segmentation. This model has several advantageous including, the ability to control the backbone's complexity, a state-of-art results on Cityscapes~\cite{cordts2016cityscapes} benchmark. The code is easy to integrate, and it is open-sourced. This model was extended with depth, ego-pose and object pose information and will be presented in Chapter~\ref{ch:ch4}.

When evaluating the performance of segmentation algorithms, several metrics are commonly used. In the context of semantic segmentation, the goal is to assign a label to each pixel in an image. Instance segmentation extends this by not only labeling each pixel, but also separating individual instances of objects. Panoptic segmentation combines both semantic and instance segmentation by providing a unified labeling scheme for all pixels, including both things and stuff classes.
\begin{itemize}

    \item \textbf{Mean Intersection over Union (mIoU)}: Also known as Jaccard Index, it measures the overlap between the predicted segmentation and the ground truth. It is computed as: $ mIoU = \frac{1}{N} \sum_{i=1}^{N} \frac{TP_i}{TP_i + FP_i + FN_i}, $ where N is the number of classes and TP, FP, and FN represent true positive, false positive, and false negative pixels, respectively, for each class.

    \item \textbf{Panoptic Quality (PQ)}: PQ measures the overall quality of panoptic segmentation, considering both semantic segmentation and instance segmentation. It combines the accuracy of segmentation masks and the alignment between predicted segments and ground truth objects. PQ is computed as follows: $PQ = \frac{ \sum_{(p,g) \in \text{TP}} \text{IoU(p,g)}  }{{\text{{TP}} + \frac{1}{2}\text{{FP}} + \frac{1}{2}\text{{FN}}}}$. The numerator sums up the Intersection over Union (IoU) ratios for all true positive (TP) instances. The denominator is combines precision and recall, dividing all the True Positives and half the False Positives and False Negatives.

\end{itemize}

\section{The transformer modules}

In the seminal work, Vaswani~\ea~\cite{Vaswani} have introduced the transformer. A neural network module that relies on attention to perform computation on sequences. Initially this module was intended for natural language processing (NLP), however, its success has shone in all data domains including, audio, image, video …etc. This module has revolutionized the field, with great success in nearly all modalities and cross-modality tasks. In this section, we will first introduce the original transformers~\cite{Vaswani} and the image variation derived from it~\cite{liu2021swin,dosovitskiy2020}. In this thesis, these modules are used extensively to construct the architecture of the proposed models. Therefore, the background of these building blocks is developed in this section.

\subsection{Attention is all you need: the transformer}

Classical sequence modeling approaches used LSTMs and RNNs which processed data sequentially, where the current state $S_t$ was based on the previous state $S_{t-1}$ and an implicit memory that represents the past. This method had limitations as the current state did not have explicit access to all previous states and only relied on the implicit memory. Transformers were introduced to overcome this problem by using an attention mechanism. This allowed each state $S_i$ to pay attention to all other states based on their importance, determined by a weighting in the attention map. For example, in the sentence "A man walked into the park with his dog. He stopped to tie his shoe." the state "He" and "his" would pay more attention to "A man" as it is more relevant to these state.

\begin{figure}
    \centering
    \includegraphics[width=0.5\textwidth]{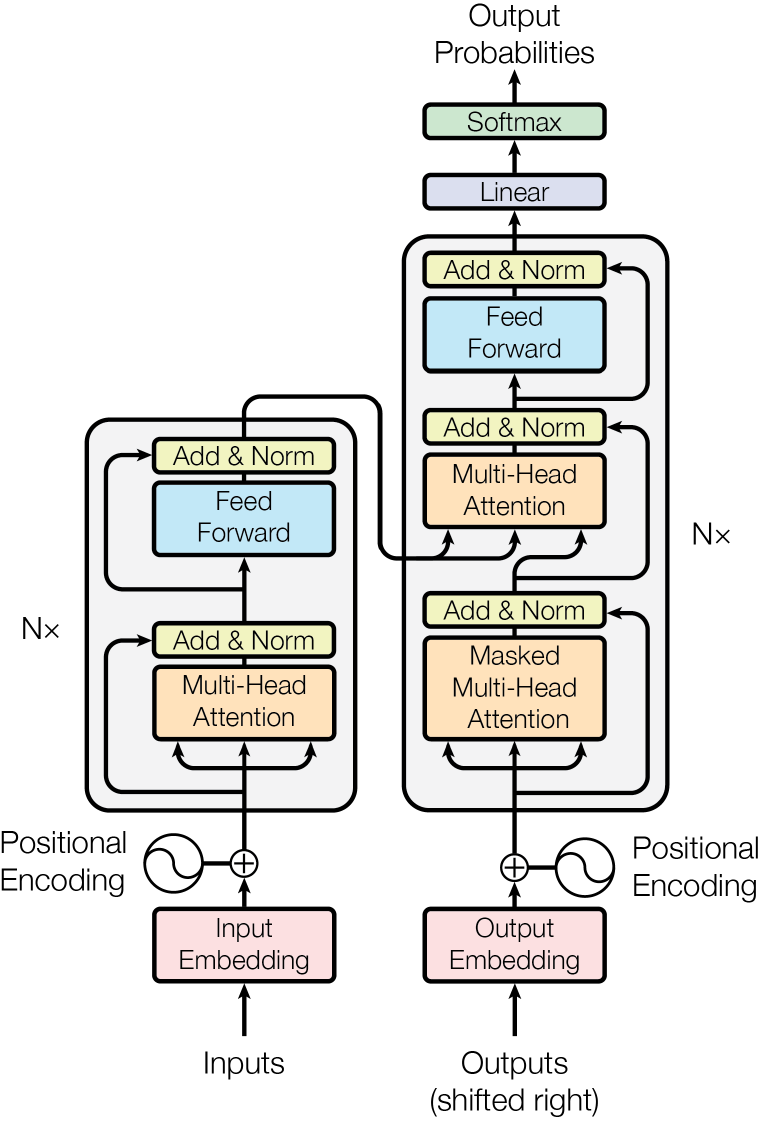}
    \caption{Overview of the transformer architecture. Figure adopted from~\cite{Vaswani}}
    \label{fig:vaswani}
\end{figure}

\reffig{vaswani} shows the architecture of the transformer. First, the text input is embedded into a continuous higher dimension. As the transformer is permutation invariant, explicit information of the embedding position should be added. A sinusoidal positional encoding is added to represent the position of each word. The first block of the transformer architecture is the \textit{encoder}. It processes the embedded and positional encoded input through a series of multi-head attention mechanisms, where each attention mechanism allows the model to attend to different parts of the input sequence and to weigh the importance of each part. The multi head attention is defined as:

First we constitute the Key, Query, Value as follows :
\begin{align}
    \text{Key: }   \mbf{K}  &= \mbf{X}\mbf{W_k} \\
    \text{Query: } \mbf{Q}  &= \mbf{X}\mbf{W_q} \\
    \text{Value: } \mbf{V}  &= \mbf{X}\mbf{W_v}
\end{align}
The multi head attention is defined as : 
\begin{align}
    \textbf{Attention} = \text{Softmax}(\frac{\mbf{Q}\mbf{K}^T}{\sqrt{d_{dim}}})\mbf{V}
\end{align}
The $\text{Softmax}(\frac{\mbf{Q}\mbf{K}^T}{\sqrt{d_{dim}}})$ produces an attention map normalized with the softmax that indicates the relative importance of the values with respect to each other. Thus, enabling the model to attend to the most relevant parts of the input. This hidden state is passed through a feedforward neural network to increase the complexity of the learned function. The skip connection allows better gradient flow. Finally, the output is normalized.

The second block of the transformer architecture is the \textit{decoder}, which generates a target sequence. Similar to the encoder, the decoder also uses multi-head attention mechanisms and feedforward neural networks. The difference is that the decoder uses the output of the encoder for the key and the value enabling cross attention. The final output of the decoder is then compared to the target sequence and the model is trained to minimize the difference between the two.

This architecture has several advantages as the attention is fully parallelizable and each state have access to all other states. However, the complexity of the attention grows quadratically with the sequence length $n$.
\subsection{The Visual transformer (ViT)}
Applying directly, self-attention to images requires that each pixel attends to every other. This does not scale for realistic images, as the cost of the attention is quadratic in the number of pixels. To this end,~\cite{dosovitskiy2020image} marked the first transformer only architecture. The idea of this paper is instead of using each pixel as input to the standard transformer, the image is divided into small patches of $16\times16$ and the image will then be represented as a sequence of patches. 

The authors applied a standard
Transformer enocder~\cite{Vaswani} directly to images, with the fewest possible modifications. To do so, they split an image
into patches of $16\times16$. Image patches are treated the same way as tokens (words) in an NLP application. They provide this sequence of linear embeddings of these patches as an input to a Transformer. 

This architecture has known great success in the computer vision community. This backbone has been applied to nearly all tasks, replacing its CNN counterpart. The flexibility and the global receptive field of transformer enable the context-aware rich feature extraction. However, as the transformer is permutation  invariant and lacks the indicative bias of convolution networks (\ie the spatial information is hard-coded in the convolution filters. a $3\times3$ convolution is applied locally on a square of $3$), it is not able to obtain great results when trained on smaller datasets from scratch.

\subsection{SwinTranformer (SwinT)}
\label{sec:swin_transfomer}
In standard ViTs, the number of tokens and token feature dimension are kept fixed throughout different blocks of the network. This is limiting, since the
model is unable to capture fine spatial details at different scales. 
The Swin Transformer~\cite{liu2021swin} proposes a multi-stage hierarchical feature extraction that computes attention within a local window, by partitioning the window into multiple sub-patches. To capture interactions between different windows (image locations), window partitioning is gradually shifted, along the depth of the network, to capture overlapping regions

\begin{figure}
    \centering
    \includegraphics[width=\textwidth]{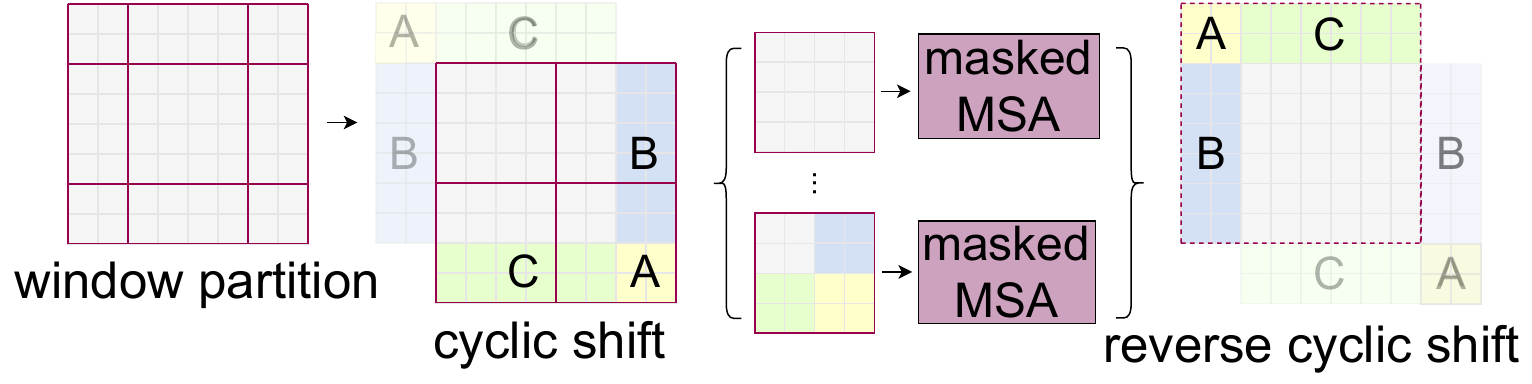}
    \caption{Overview of Swin tranformer block. Adpoted from~\cite{liu2021swin}}
    \label{fig:swin}
\end{figure}

\reffig{swin} shows the architecture. Swin Transformer incorporates a shifted window approach for computing self-attention, as demonstrated in the accompanying illustration. In Layer l (on the left), a standard window partitioning strategy is used and self-attention is computed within each window. In the subsequent Layer l+1 (on the right), the window partitioning is shifted, leading to the creation of new windows. The self-attention computation in these new windows crosses the boundaries of the previous windows in Layer l, thereby establishing connections between the windows.

\begin{figure}
    \centering
    \includegraphics[width=\textwidth]{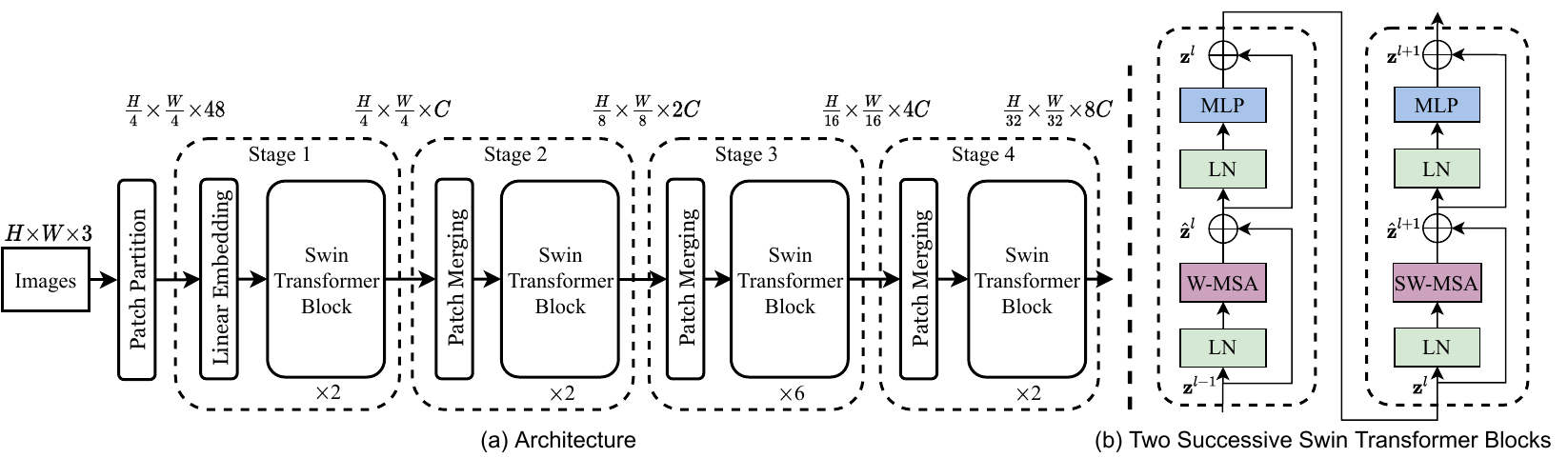}
    \caption{Overview of Swin tranformer architecture. Adpoted from~\cite{liu2021swin}}
    \label{fig:swin-overview}
\end{figure}

\chapter*{}

\chapter{Image-to-X with conditional GANs}
\label{ch:ch3}
In the initial stages of the thesis, the problem of using images for autonomous driving was approached from a broader perspective, exploring not only the depth modality as output but also other crucial modalities, such as semantic segmentation and image generation. These modalities play a vital role in enabling driving systems to comprehend the geometry of the surrounding environment and identify different participating agents. 

During the first year of the thesis, the primary objective was to model complex environmental situations that autonomous driving systems may encounter. Factors like weather conditions and night-time scenarios present significant obstacles for these systems. It is crucial to develop the capability to handle such variations and ensure consistent performance in diverse circumstances. 

This chapter therefore explores domain adaptation techniques using generative adversarial networks (GANs) to bridge the gap between the source and target domains. It focuses using on conditional GANs for style transfer. Specifically, translating annotated overcast-daytime images into night-time or other weather condition offers a cost-effective means of constructing annotated datasets. Additionally, this study reveals a limitation in traditional conditional GANs, which lack full conditional capabilities, which is also known as “hallucination”. To address this limitation, a novel approach to train cGANs called “\ac cGAN” is proposed. This method allows for more consistent output generation, making conditional GANs fully conditional. Furthermore, a noteworthy feature developed in this research is the generative model night-to-day, designed specifically for domain adaptation. This model is selected as a potential feature to be added to the Renault ADAS stack.

This chapter is based on two publications : 
\begin{itemize}
    \item Boulahbal Houssem Eddine, Adrian Voicila, and Andrew I. Comport. "Are conditional GANs explicitly conditional?." British Machine Vision Conference. 2021.

    \item Boulahbal, Houssem Eddine, Adrian Voicila, and Andrew I. Comport. "Un apprentissage de bout-en-bout d'adaptateur de domaine avec des réseaux antagonistes génératifs de cycles consistants." Journée des Jeunes Chercheurs en Robotique. 2020

\end{itemize}

\section{Introduction}
In the rapidly evolving field of artificial intelligence, one of the key challenges is the ability to generalize, especially in the domain of Autonomous Driving (AD). Autonomous vehicles are operating in various and dynamic environments where they are faced with a variety of situations, such as weather variations. To ensure safe and reliable function of autonomous vehicles, it is crucial to develop AI models capable of adapting and generalizing to these complex environmental situations.

One major obstacle in training deep learning models for autonomous driving is the availability of annotated datasets. Annotated datasets are essential for supervised learning, where the model is trained on labeled examples to recognize and interpret different objects and events in the environment. However, the manual annotation of a large-scale dataset for autonomous driving is an exceedingly laborious and time-consuming task, which makes it practically impossible to create a fully annotated dataset covering all possible environmental variations.

\begin{figure}[]
    \centering
    \includegraphics[width=\textwidth]{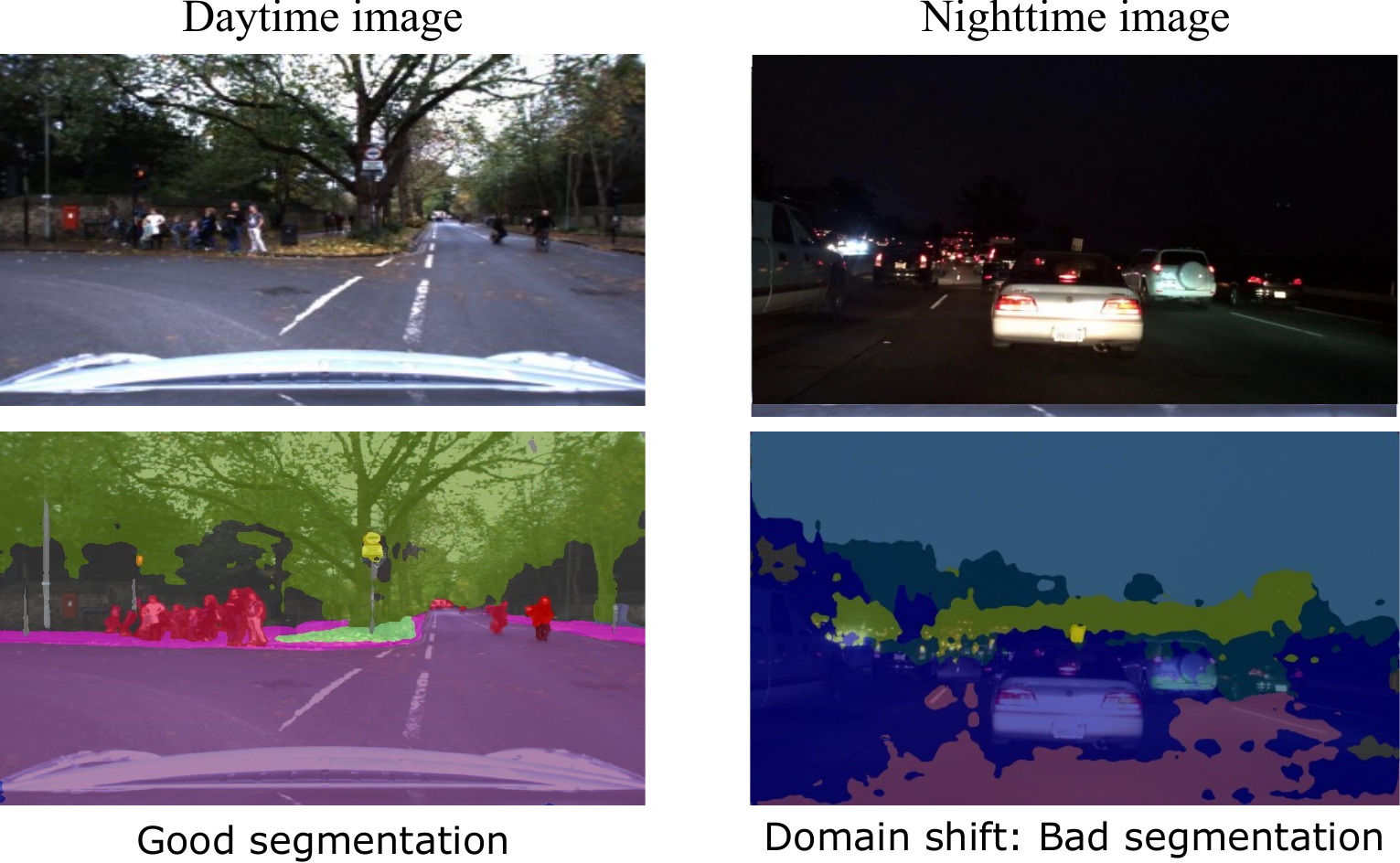}
    \caption{Comparison of DeepLabV3+~\cite{chen2018encoder} model's performance on daytime and nighttime scenarios. The model trained solely on daytime images shows significant limitations in accurately predicting outcomes in nighttime scenarios due to the dramatic differences in lighting conditions, object appearance, and overall scene dynamics.}
    \label{fig:DomainAdaptation}
\end{figure}

As shown in~\reffig{DomainAdaptation}, the model trained solely on daytime images (DeepLabV3+~\cite{chen2018encoder} performs semantic segmentation) has significant limitations when it comes to accurately predicting outcomes in nighttime scenarios. The dramatic difference between the visual characteristics of daytime and nighttime environments poses a considerable challenge for AI models, as lighting conditions, object appearance and overall scene dynamics undergo significant changes. To circumvent this annotation bottleneck, researchers have explored various techniques, and one promising approach is the use of domain adaptation techniques~\cite{Vu2019,Yang2020,Sankaranarayanan2018,Kim2019,porav2019don}. Domain adaptation aims to transfer knowledge from a source domain, where labeled data is available, to a target domain, where only unlabeled or sparsely labeled data exists. By leveraging the knowledge from the source domain, domain adaptation techniques enable the model to generalize well to the target domain without requiring extensive annotation efforts.

In this chapter, we propose the utilization of domain adaptation techniques using conditional adversarial generative networks (cGANs), specifically the CycleGAN~\cite{zhu2017unpaired} (Cycle-Consistent Generative Adversarial Network), to address the challenge of translating the source annotated domain into the target domain. The CycleGAN framework is a powerful tool that allows for the generation of synthetic data in the target domain by learning the mapping between the two domains in an unsupervised manner. Furthermore, while developing this method, the conditionality of the cGANs is further explored and novel method is proposed to address the problem of conditionality of conditional GANs.

\section{Improving generalization using cGANs}

\begin{figure}[]
    \centering
    \includegraphics[width=\textwidth]{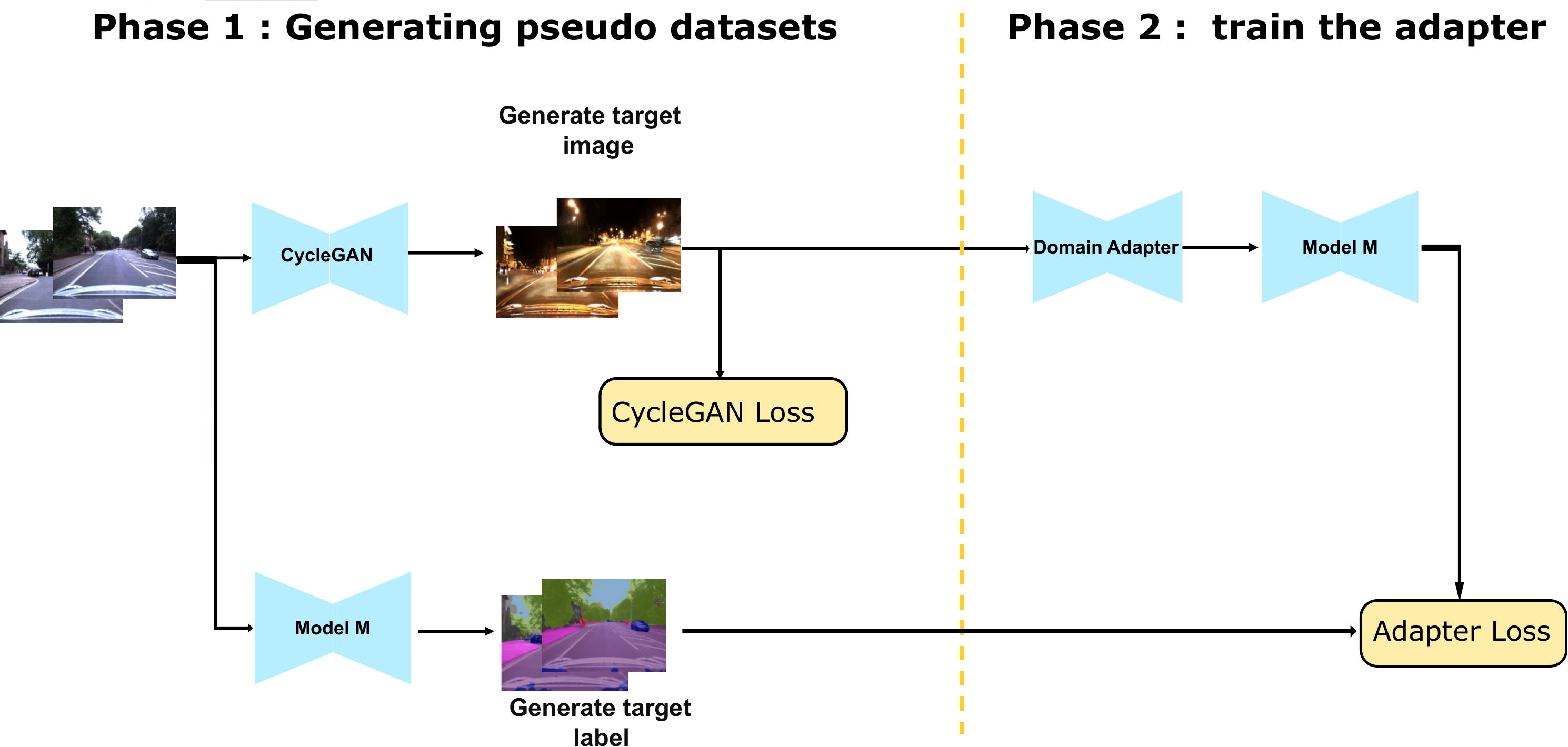}
    \caption{This figure shows a two-phase pipeline used to address the domain shift problem. In the first phase, a pseudo dataset is generated using a CycleGAN model that is trained for day-to-night image translation. This translation enables the creation of a synthetic annotated dataset of night images using the model M on the day-time images. The second phase involves training an adapter to handle the day-night variation, maximizing the performance of the model M.}
    \label{fig:PipelineAdaption}
\end{figure}

The first objective was to reproduce the results of the "Don't Worry About the Weather" paper~\cite{porav2019don}. The pipeline shown in ~\reffig{PipelineAdaption}. To achieve this, a two-phase pipeline was performed. In this context, a CycleGAN model for image translation is firstly trained. CycleGAN is a popular framework used to learn mappings between two different domains. In our case, we want to perform day-to-night translation. Using this translation, it is possible to construct an annotated dataset of synthetic night images. The annotation of these synthetic night images is done using the model M on the day-time images. The second phase involves training an adapter to account for the shift between the day-night variation, and it is trained to maximize the performance of the model M. 

\begin{figure}[]
    \centering
    \includegraphics[width=\textwidth]{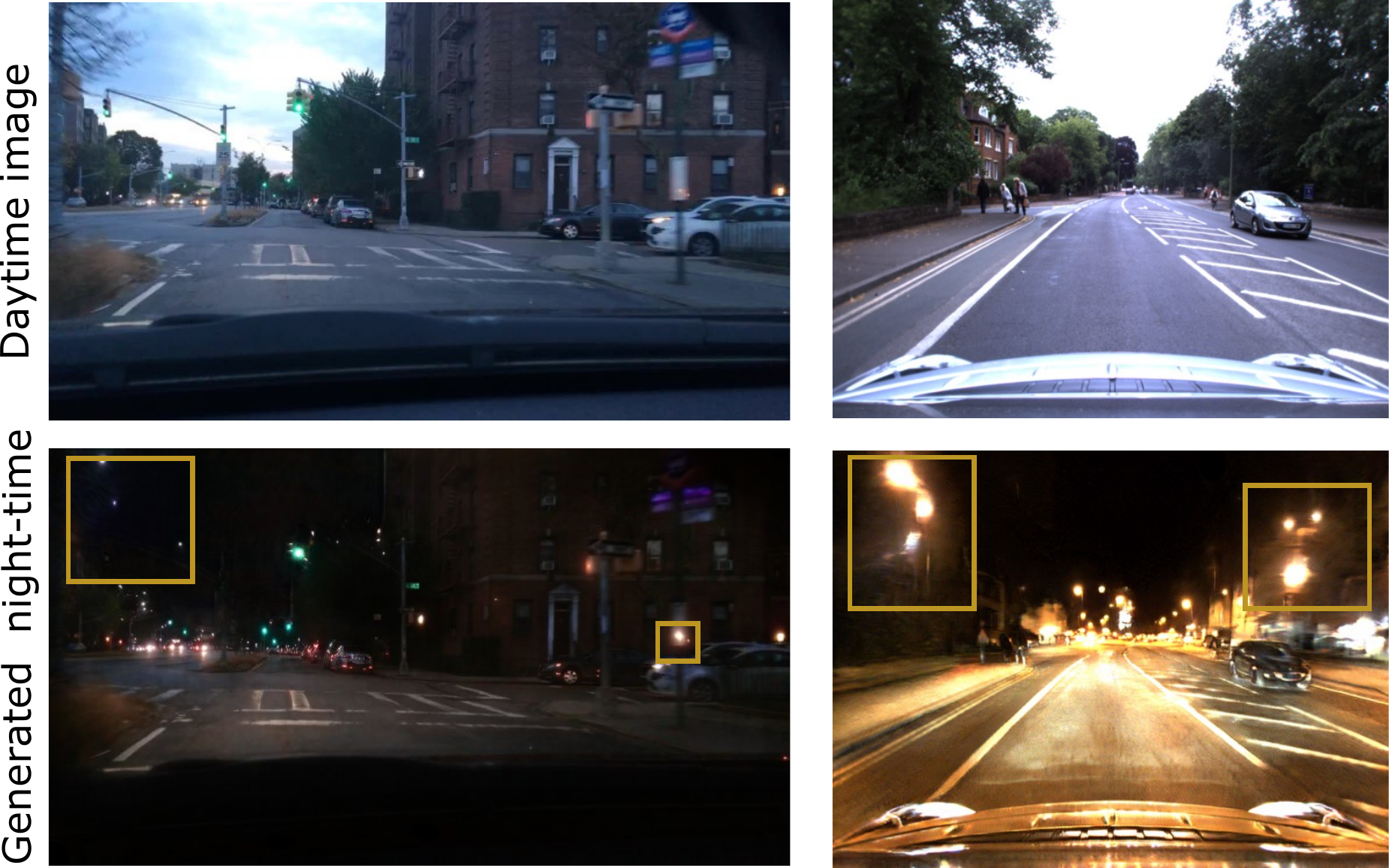}
    \caption{Day-to-night translation results. The model successfully handles variations in lighting conditions and object appearance, transforming daytime images into realistic nighttime counterparts. However, occasional hallucinations of poles and objects not present in the scene can be observed, as depicted in the examples shown in the figure.}
    \label{fig:Day-to-night}
\end{figure}

The Robotcar dataset~\cite{RobotCarDatasetIJRR} is used to train and evaluate the system. The dataset comprises a diverse collection of sensor data captured by an autonomous vehicle as it navigates Oxford city. The dataset covers a wide range of environmental conditions, such as varying weather conditions (e.g., sunny, cloudy, rainy), different times of day (daytime, nighttime), and diverse urban scenarios (e.g., streets, intersections, landmarks). Overall, the RobotCar dataset provides a comprehensive and representative collection of data for training and evaluating the system. Its diverse environmental conditions make it an ideal choice for domain adaptation experiments. \reffig{Day-to-night} shows the results of translating the day-to-night translation. As observed, the model is able to handle well the variations in lighting conditions and object appearance, successfully transforming daytime images into realistic nighttime counterparts.

To enhance the performance of the system, a domain adapter, A, is trained. This adapter is designed to improve a pretrained and frozen model M, which is the model that will be used in production. The adapter plays a crucial role in accounting for the changes that occur during the translation process. The adapter is supervised using the following loss function: 

\begin{align}
    \label{eq:adpt_eq}
    \mathcal{L}_{adpt} = \sum_{i=0}^{m} | M(A(\mbf{I}_{night}^i)) - \mbf{S}_{mask}^i | 
\end{align}
Where m is the number of images in the dataset, M is the semantic segmentation model (DeepLabV3+~\cite{chen2018encoder}), A is the adapter, $\mbf{I}_{night}^i$ is the synthetic image obtained from translating $\mbf{I}_{day}^i$ with the GAN, and $\mbf{S}_{mask}^i$ is the semantic mask obtained using $\mbf{S}_{mask}^i = M(\mbf{I}_{day}^i)$. The adapter is trained to make the prediction of the semantic segmentation consistent for day and night.

To effectively handle the domain shift caused by various weather conditions, an adapter is trained to minimize the domain shift between the source domain and the target domain. For instance, during inference, if the source image belongs to the overcast-daytime category, we simply use the model M without any adaptations. However, if the source image comes from a different condition, we apply the corresponding adapter to ensure accurate and context-specific output. This approach could also be applied for other weather variation such as snow, rain …etc.

The advantage of this approach is its ability to effectively manage various weather conditions and enhance the system's overall performance. Instead of retraining the entire model, we can train specialized adapters for each condition. Additionally, this method is versatile and can be applied to different tasks. For instance, the model M can be replaced with other modalities like depth or object detection, requiring only modifications to the objective function. Nevertheless, it is important to consider the computational cost associated with this method, particularly in the context of embedded systems such as autonomous driving.

\begin{figure}[]
    \centering
    \includegraphics[width=\textwidth]{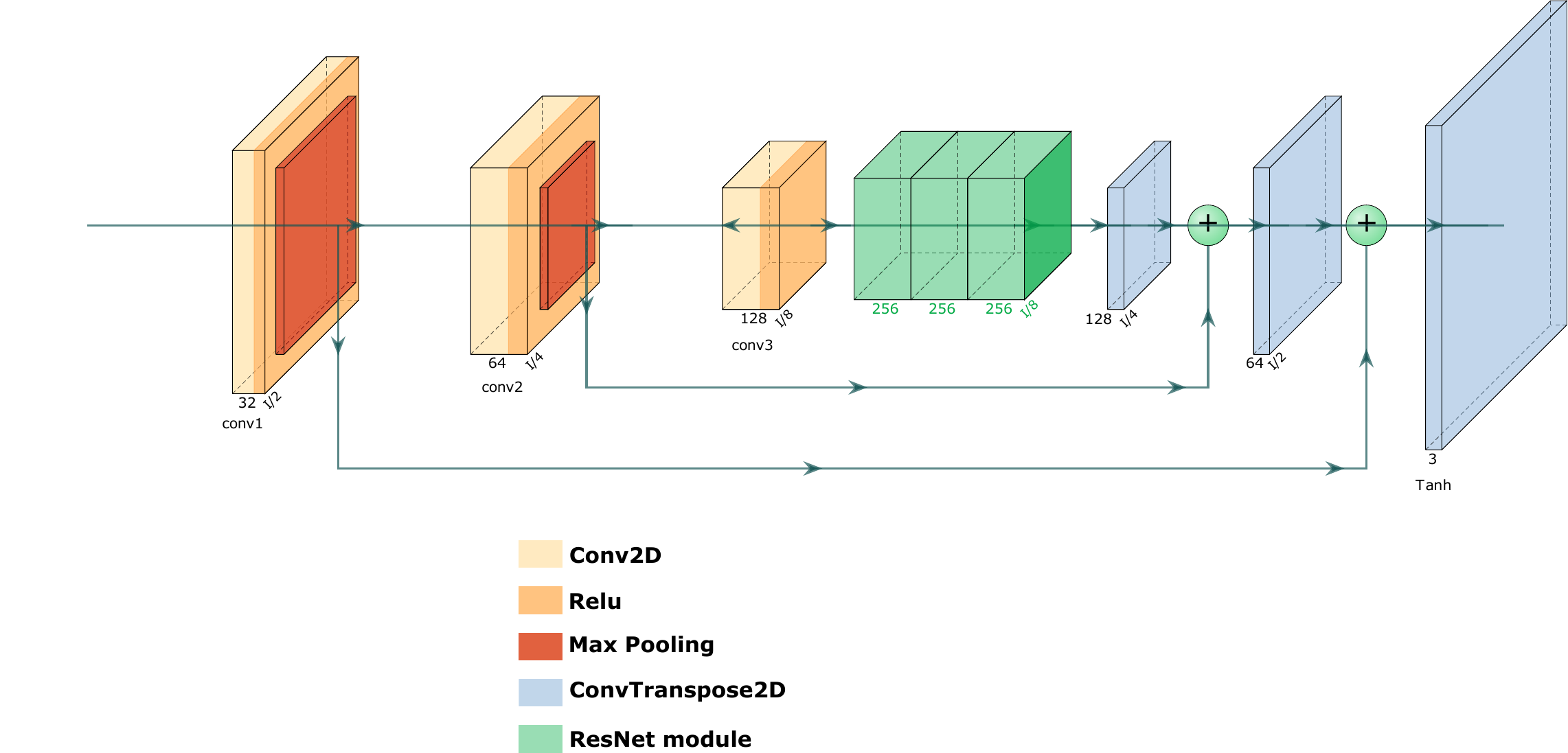}
    \caption{Adapter architecture. Leveraging encoder-decoder design. Relu is used as The activation function. Max pooling is used to downsample the feature maps. and the ConvTranspose2D are used to upsample the feature maps. The ResNet blocks are used in the lower resolution feature maps to remap the representation of night images into a representation that is optimal for segmentation.}
    \label{fig:ch3_architecture}
\end{figure}

\subsection{Architecture}
\reffig{ch3_architecture} shows the architecture of the adapter. The architecture of the adapters is based on a simple yet effective encoder-decoder design. With the aim of preserving the overall structure of the scene even when conditions change, skip-connections and a ResNet bottleneck are employed. This approach enables seamless feature transfer from the input side to the output side of the network. By incorporating skip-connections, we ensure that important features are preserved and propagated throughout the network. This combination of techniques not only facilitates direct feature transfer, but also enhances the adaptability and robustness of the model.

\subsection{End-to-end training}
After reproducing the results, one improvement that could be proposed to improve the system is to train the pipeline in end-to-end manner. That is training the CycleGAN and the adapter at the same time.

Training the model in a two-phase approach is suboptimal due to the presence of artifacts, particularly hallucinations, introduced when the CycleGAN is trained separately. These artifacts undermine the consistency of the pair (day, synthetic-night), consequently impacting the quality of the adaptation results. In~\cite{boulahbal:hal-03365983}, we proposed a novel solution to address this challenge. We advocate for training the CycleGAN model alongside the adapter in an end-to-end manner. By incorporating the adapter within the training process, the CycleGAN model can benefit from additional supervision based on semantic information. This integration enables the entire system, composed of the CycleGAN and the adapter, to collectively enhance the performance of domain adaptation.

The end-to-end training approach holds several advantages. Firstly, it allows for tighter coordination between the CycleGAN and the adapter, fostering a more seamless integration of the two components. This integration ensures that the semantic information captured by the adapter is effectively utilized during the transformation process carried out by the CycleGAN. Secondly, the additional supervision provided by the adapter's semantic information can help guide the training of the CycleGAN model. This guidance plays a crucial role in reducing the occurrence of artifacts, such as hallucinations, which often arise when the CycleGAN is trained independently. By leveraging the semantic cues, the end-to-end training framework promotes more coherent and consistent translations between the day and synthetic-night domains.

The objective function becomes : 

\begin{align}
    \mathcal{L} = \mathcal{L}_{adpt} + \mathcal{L}_{CycleGAN}
\end{align}

Where $\mathcal{L}_{adpt}$ is defined in~\refeq{adpt_eq} and $\mathcal{L}_{CycleGAN}$ is defined in~\refeq{cycle_gan}.

\subsection{Results}

As it is defined in~\cite{porav2019don}, to evaluate the performance of the proposed pipeline, an experiment was conducted using the RobotCar Dataset~\cite{RobotCarDatasetIJRR}. The experiment involved applying style-transfer techniques with cycle-consistency GAN generators to generate testing sequences specifically for night conditions. The groundtruth is obtained using the model M on day-images. the selected model M is DeepLabV3+~\cite{chen2018encoder} The mIoU is used as metric.

\begin{table}[]
    \centering
    \resizebox{\textwidth}{!}{
    \begin{tabular}{|c||c|c|c|}
            \hline
            & DeepLabV3+~\cite{chen2018encoder} without adaptation & DeepLabV3+~\cite{porav2019don} with adapter & Adapter with end-to-end training \\  
            \hline
         mIoU on RobotcarDataset~\cite{RobotCarDatasetIJRR}& 0.1850 & 0.5198 &  \textbf{0.5721}\\
          \hline
    \end{tabular}
    }
    \caption{The table presents the results of adaptation and performance enhancement in semantic segmentation, specifically for nighttime images. The proposed adaptation technique demonstrates significant improvement in the model's performance on nighttime images. Additionally, the end-to-end training further enhances the results achieved.}
    \label{tab:results_adaptation}
\end{table}

\begin{figure}[]
    \centering
    \includegraphics[width=\textwidth]{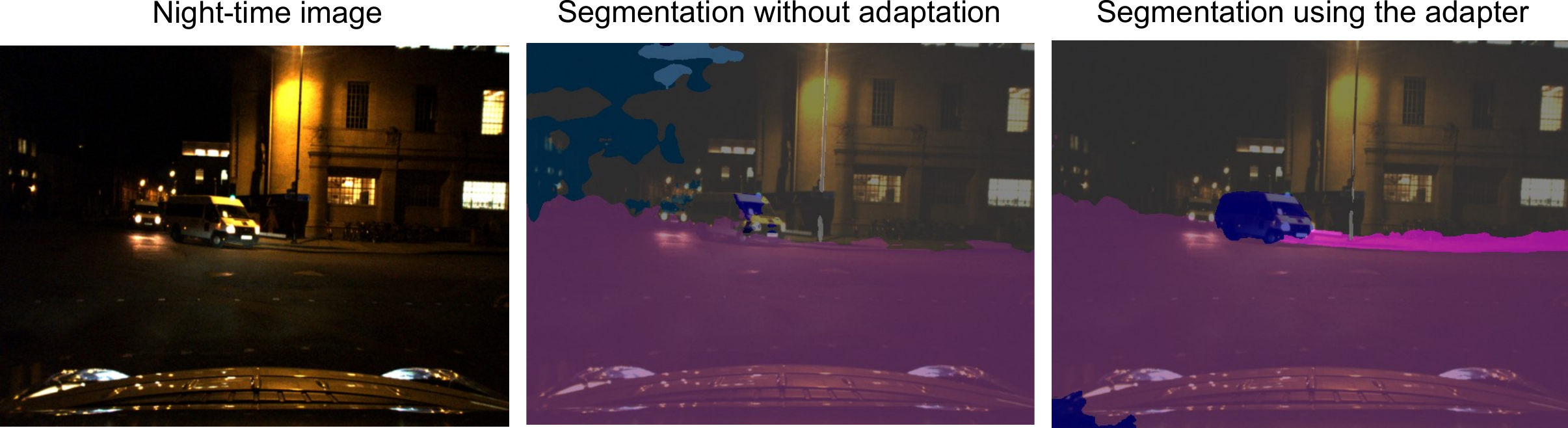}
    \caption{Comparison of the qualitative results of the segmentation model with/without the adapter.}
    \label{fig:qualitative_performance}
\end{figure}

\reftab{results_adaptation} represents the results of the adaptation, as observed. The domain adaptation that was proposed in~\cite{porav2019don} improves significantly the performance of the semantic segmentation on the night condition. The end-to-end approach improves further, and this results in a notable enhancement in overall accuracy. This suggests that the proposed domain adaptation technique effectively bridges the gap between the source and target domains, enabling the model to better handle the challenges posed by nighttime imagery. Moreover, \reffig{qualitative_performance} demonstrates the qualitative performance of the end-to-end trained model, showcasing the model's ability to accurately segment objects in low-light conditions. The segmentation outputs exhibit improved object consistency, sharper boundaries, and reduced noise, highlighting the effectiveness of the proposed adaptation method in enhancing semantic segmentation results.

\section{Conditionality of Conditional GANs}

In Figure \ref{fig:Day-to-night}, it can be observed that the translation from day to night is not fully consistent, as the model generates poles that are not present in the actual scene. The conditional GAN model, which was trained to generate night images, successfully captures the overall distribution of night scenes. However, it is not fully capable of representing night images conditioned on day images without hallucinating objects, resulting in inconsistencies in certain regions of the generated image. This issue is not unique to conditional GANs and is observed in other generative models across different domains, including language, even in lower-dimensional spaces \cite{openai2023gpt4,ouyang2022training}.

The conditionality of cGANs is at the crux of their theoretical contribution, and its impact therefore merits in-depth analysis. From the existing literature it is not clear, however, if this widely used architecture fully models conditionality. Empirically, the impressive results obtained with cGANs show that the generator automatically seeks to incorporate conditional variables into its generated output. Fundamentally, the generator is, however, free to generate whichever output as long as it satisfies the discriminator. Therefore, the conditionality of cGAN also depends on the conditionality of the discriminator. This begs the question as to whether or not the baseline architecture of cGANs explicitly models conditionality and if not, how can the core adversarial architecture be redefined to explicitly model conditionality? This is therefore the object of the following sections.

Problems with cGANs conditionality have been observed independently for different tasks in the literature. Label-to-image tasks observe that using only adversarial supervision yields bad quality results~\cite{Sushko2020YouON,Park2019SemanticIS, Wang2018HighResolutionIS,Liu2019LearningTP}. "Single Label"-to-image tasks~\cite{Brock2019LargeSG} observe class leakage. It is also well known that cGANs are prone to mode collapse~\cite{richardson2018gans}. In the following sections, it is suggested that all these problems are related to the lack of a conditional discriminator.

\begin{figure*}
    \centering
    \includegraphics[width=0.9\textwidth]{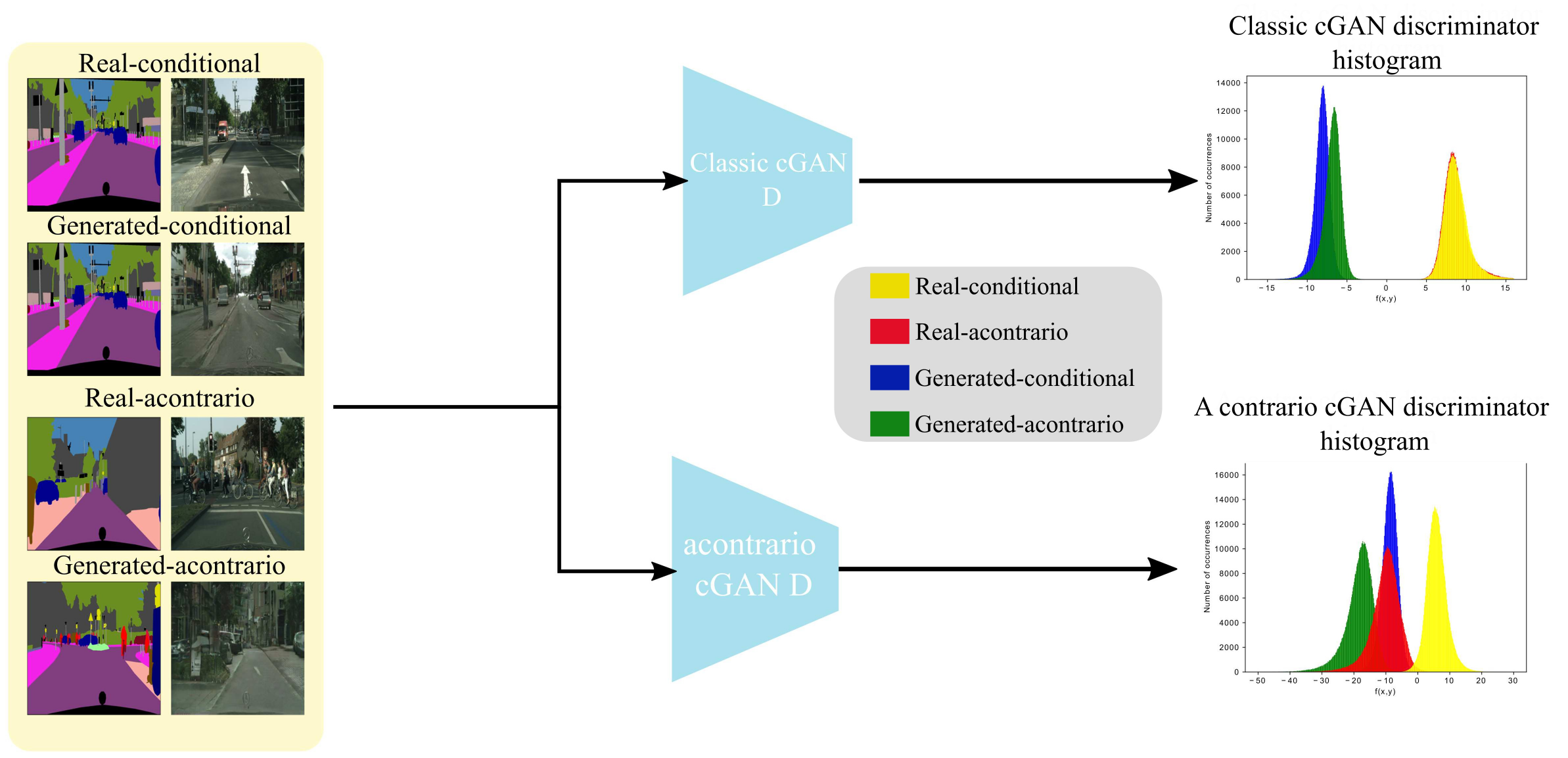}
    \caption{	 The classic cGAN and the proposed \ac cGAN discriminators are tested with 500 validation images of the Cityscapes dataset on both conditional and unconditional label-to-image input set.  Unconditional inputs (Real \ac and Generated \ac) set are obtained by randomly shuffling the original conditional sets of data. The classic cGAN discriminator fails to classify unconditional input set as false, as seen by the histogram distributions on the right (real \ac in red is classified as true). The proposed method trains the discriminator with a general \ac loss to classify an unconditional input set as fake (note that no extra training samples are required). The proposed \ac cGAN correctly classifies all four modalities (blue, green, red, yellow) correctly.}
    \label{fig:pipeline}
\end{figure*}

Consider a simple test of conditionality on a learned discriminator for the task of label-to-image translation, shown in Figure~\ref{fig:pipeline}. The conditional label input is purposely swapped with a non-corresponding input drawn randomly from the input set (\eg labels). From this test, it is revealed that the discriminator does not succeed to detect the entire set of \ac examples (defined in Section~\ref{sec:consistency}) as false input pairs. This suggests that the generator is not constrained by the discriminator to produce conditional output, but rather to produce any output from the target domain (street images in this case). Furthermore, in practice, the large majority of methods that exploit cGANs for label-to-image translation, if not all, add additional loss terms to the generator to improve conditionality. These loss terms are, however, not adversarial in their construction. For example, high resolution image synthesis approaches such as~\cite{Wang2018HighResolutionIS} suffer from poor image quality when trained with only adversarial supervision~\cite{Sushko2020YouON}. Considering the well known pix-to-pix architecture~\cite{Isola2017ImagetoImageTW}, a L1 loss applied to the generator was introduced to improve performance. This additional term seeks to enforce conditionality on the generator, but does not act explicitly on the discriminator. Subsequently, one could question if the conditionality obtained by such methods is obtained via this loss term, which is not part of the adversarial network architecture. Moreover, adding an extra loss term to the generator has now become the defacto method for improving cGANs results. For example, perceptual loss~\cite{Johnson2016PerceptualLF} and feature-matching~\cite{Salimans2016ImprovedTF} have been proposed and reused by many others\cite{Wang2018HighResolutionIS,chen2017photographic,Park2019SemanticIS,Ntavelis2020SESAMESE}. As demonstrated in the experiments, different tasks such as image-to-depth or image-to-label also exhibit these drawbacks.

In this chapter it will be argued that simply providing condition variables as input is insufficient for modelling conditionality and that it is necessary to explicitly enforce dependence between variables in the discriminator. It will be demonstrated that the vanilla cGAN approach is not explicitly conditional via probabilistic testing of the discriminator's capacity to model conditionality.
With this insight, a new method for explicitly modelling conditionality in the discriminator and subsequently the generator will be proposed. This new method not only offers a solution for conditionality, but also provides the basis for a general data augmentation method by learning from the contrary (\ac data augmentation).

\subsection{Classic cGAN}
\label{sec:ccgan}

Classical cGAN training is based on conditionally paired sets of data $\mathcal{C}(\mbf{x},\mbf{y})$ where $\mbf{x}\sim p(\mbf{x})$ is the condition variable and $\mbf{y}\sim p(\mbf{y}|\mbf{x})$ is the real training variable. The generator of a cGAN outputs a transformed set of data $\mathcal{C}_G(\mbf{x},\mbf{y}_G)$ composed of the generator output variable $\mbf{y}_G\sim p_G(\mbf{y})$ and the condition variable. These sets of data will be called "real-conditional" and "generated-conditional" respectively. 
The discriminator is defined as:
\begin{align}
    D(\mbf{x},\mbf{y}):=\mathcal{A}(f(\mbf{x},\mbf{y}))
    \label{eq:discriminator}
\end{align}
Where $f(.)$ is a neural network function of $\mbf{x}$ and $\mbf{y}$, and $\mathcal{A}$ is the activation function whose choice depends on the objective function. The cGAN objective function is defined as: 
\begin{align}
         \mathcal{L}_{adv}= \min_{G} \max_{D} \left(\: \mathbb{E}_{\mbf{x}\sim p(\mbf{x}),\mbf{y}\sim p(\mbf{y}|\mbf{x})}\big[log(D(\mbf{x},\mbf{y})]\big] +   \mathbb{E}_{\mbf{x}\sim p(\mbf{x})}\big[log[1 - D(\mbf{x},G(\mbf{x}))]\big] \right)
\label{eq:classic_loss}
\end{align}
The min-max activation function is defined as a Sigmoid $\mathcal{A}(\mbf{x})=\left({\frac{1}{1+e^{-\mbf{x}}}}\right)$. 

\subsection{Evaluating conditionality} 
\label{sec:conditioneval}
The objective of this section is to propose methods to test the conditionality of cGAN networks. State-of-the-art approaches have focused on evaluating cGAN architectures with metrics applied to the generator output. Since the generator and discriminator are coupled, these metrics essentially evaluate the full GAN architecture.

A proposal is made to test the conditionality by visualizing the probability distribution at the output of the discriminator. Due to the fact that adversarial training involves a zero-sum game between a generator and a discriminator, both the generator and discriminator should seek to reach an equilibrium (Eq~\ref{eq:cost}) at the end of training. One issue for GANs is that when the discriminator dominates, there is a vanishing gradient problem~\cite{Arjovsky2017TowardsPM}. It is therefore more difficult (but not impossible) to isolate the discriminator during training to evaluate its capacity to detect unconditional examples as false. For this reason, an optimal discriminator can be used to give insight for evaluation purposes, as in~\cite{Arjovsky2017TowardsPM,Farnia2020GANsMH}. An optimal discriminator is essentially a binary classifier which classifies between true and fake (see Eq~\eqref{eq:classic_loss}).

In order to test the optimal discriminator, consider that the generator has been fixed after a certain number of iterations and the discriminator has been allowed to converge to an optimal solution based on the following objective function:
 \begin{align}
    \max_{D\in \mathbb{D}} V(G_{fixed},D) 
    \label{eq:optimaldiscrim}
\end{align}

The evaluation subsequently involves analyzing the distributions produced by the optimal discriminator (\refeq{optimaldiscrim}) given test distributions containing unconditional or \ac sets of data-pairings. The capacity of the discriminator to correctly classify unconditional data as false is then analyzed statistically. Section~\ref{sec:consistency} provides a formal definition of these unconditional data pairings. Probability distributions are visualized and evaluated by histogram analysis on the discriminator features in the last convolution layer. 

\subsection{A contrario conditionality loss}
\label{sec:consistency}

The proposed \ac cGAN approach is based on training with unconditionally paired sets of data, obtained by randomly shuffling or re-paring the original conditional sets of data. The \ac set is defined as $\mathcal{C}_U(\mbf{\tilde x},\mbf{y})$, where $\mbf{\tilde x}\sim p(\mbf{x})$ is the \ac conditional variable ($\mbf{\tilde x} \neq \mbf{x}$) and $\mbf{y}$ is the real training variable as in Section~\ref{sec:ccgan}. In this case $\mbf{\tilde x}$ and $\mbf{y}$ are independent. The generator of the \ac cGAN outputs a transformed set of data $\mathcal{C}_{UG}(\mbf{\tilde x},\mbf{y})$ composed of the generator output variable $\mbf{y}_G\sim p_G(\mbf{y})$ and the random variable $\mbf{\tilde x}$. For the purpose of this paper these two sets of data will be called "real-\ac" and "generated-\act" respectively. The motivation to create these new sets is to train the discriminator to correctly classify unconditional data as false. Figure~\ref{fig:pipeline} shows the four possible pairings. In practice, random sampling of \ac pairs is carried out without replacement and attention is paid to not include any conditional variables into a same batch while processing.

In order to enforce conditionality between $\mbf{y}$ and $\mbf{x}$ an \ac term is proposed as:
\begin{align}
        \mathcal{L}_{ac}= \max_{D} \: & \left(\mathbb{E}_{\mbf{\tilde x} \sim p(\mbf{\tilde x}) ,\mbf{y}\sim p(\mbf{y})}\big[log(1 - D(\mbf{\tilde x},\mbf{y}))\big]  + 
        \mathbb{E}_{\mbf{\tilde x} \sim p(\mbf{\tilde x}) ,\mbf{x} \sim p(\mbf{x})}\big[log(1 - D(\mbf{\tilde x},G(\mbf{x})))\big] \right)
\label{eq:ac_loss}
\end{align}
The first term enforces the real-\act pairs to be classified as fakes. The second terms enforce the generated-\act as fake. The final loss is:
\begin{align}
    \mathcal{L}_{adv}^{'}  = \mathcal{L}_{adv}+ \mathcal{L}_{ac}
\label{eq:overall_loss}
\end{align}

\section{Experimental section}
\label{sec:experiments}
Several experiments will be presented that evaluate the conditionality of cGANs including: Monocular depth estimation on~\cite{Silberman:ECCV12}; Real image generation from semantic masks on Cityscapes dataset~\cite{cordts2016cityscapes}; "Single label"-to-image on CIFAR-10~\cite{cifar}; Semantic segmentation using pix2pix on Cityscapes dataset. 
For label-to-image generation, pix2pix\cite{Isola2017ImagetoImageTW},  pix2pixHD\cite{Wang2018HighResolutionIS}, SPADE\cite{Park2019SemanticIS} and CC-FPSE\cite{Liu2019LearningTP} were used to test the conditionality and to highlight the contribution of the \ac cGAN with respect to more recent approaches. Depth estimation and image-to-label are  structured prediction problems offer strong metrics for evaluating cGANs and various public datasets are available for training. While the scope of conditional evaluation has been limited to tasks that could provide a metric to evaluate both the conditionality and the quality of the generation, the proposed approach is general and not specific to these particular tasks. During training, the network's architecture, the additional losses, the hyper-parameters
and data augmentation schemes are kept as in the original papers ~\cite{Isola2017ImagetoImageTW, Park2019SemanticIS, Wang2018HighResolutionIS, Liu2019LearningTP}. The new additional \ac term is the only difference between the compared methods. 

\subsection{Evaluating conditionality}
\label{sec:eval}
\begin{figure*}
\centering    
\begin{subfigure}{0.24\textwidth}
  \centering
  \includegraphics[width=\linewidth]{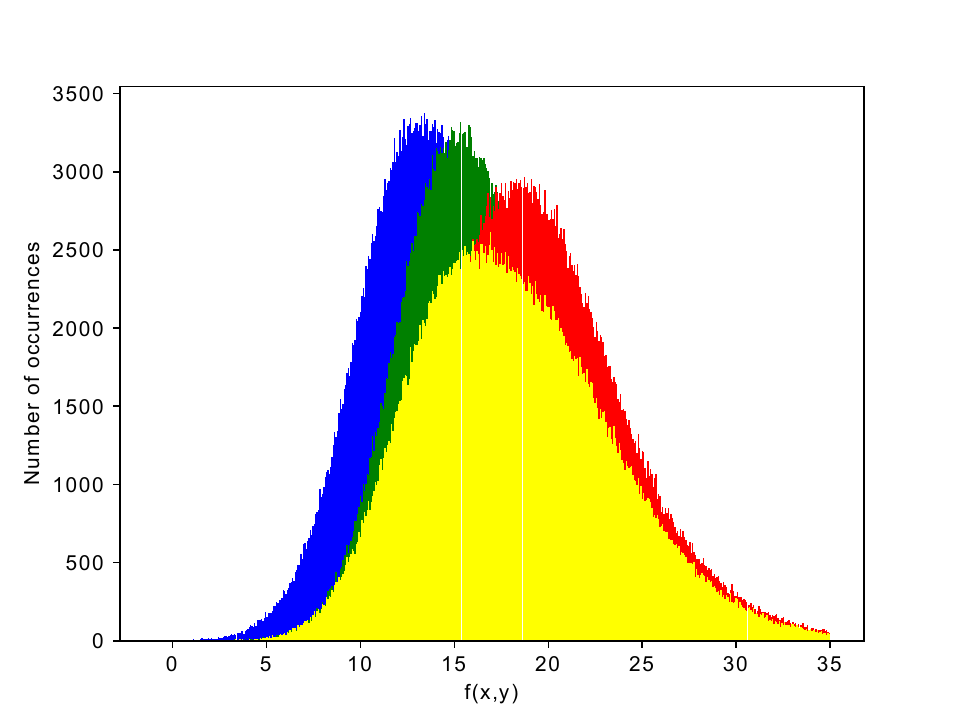}
  \caption{}
  \label{fig:baseline_D}
\end{subfigure}%
\begin{subfigure}{0.24\textwidth}
  \centering
  \includegraphics[width=\linewidth]{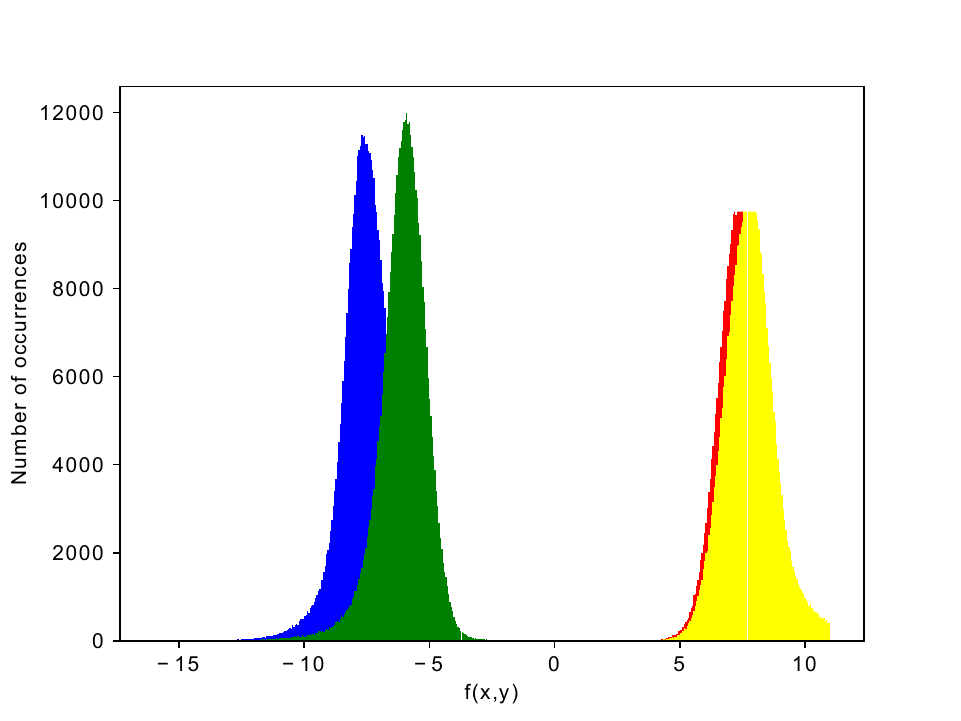}
  \caption{}
  \label{fig:baseline_200}
\end{subfigure}%
\begin{subfigure}{0.24\textwidth}
  \centering
  \includegraphics[width=\linewidth]{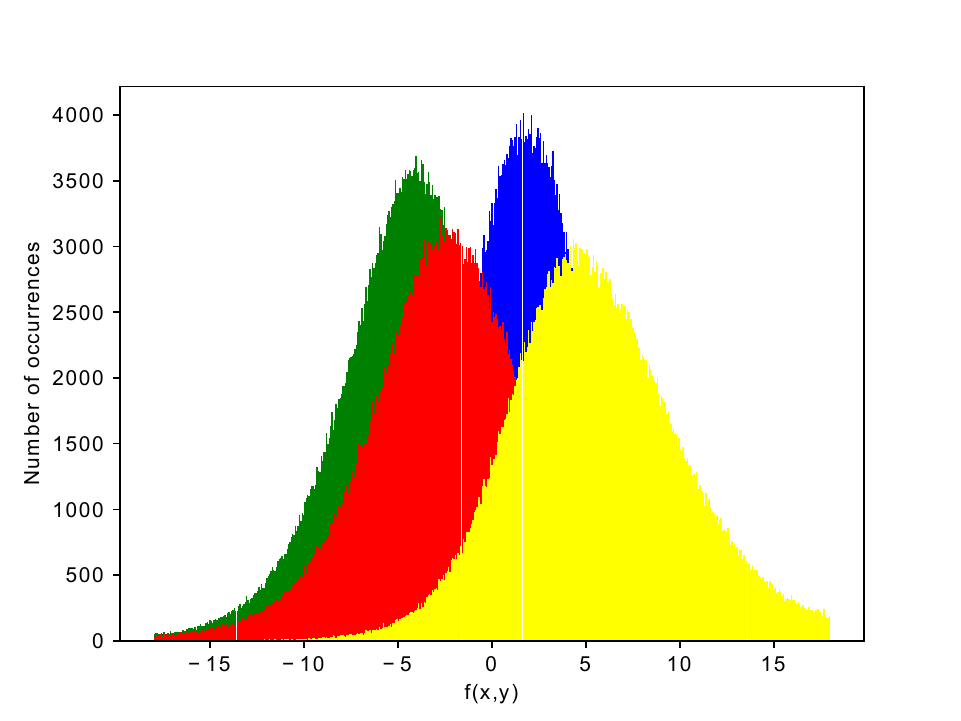}
  \caption{}
  \label{fig:explicit_D}
\end{subfigure}%
\begin{subfigure}{0.24\textwidth}
  \centering
  \includegraphics[width=\linewidth]{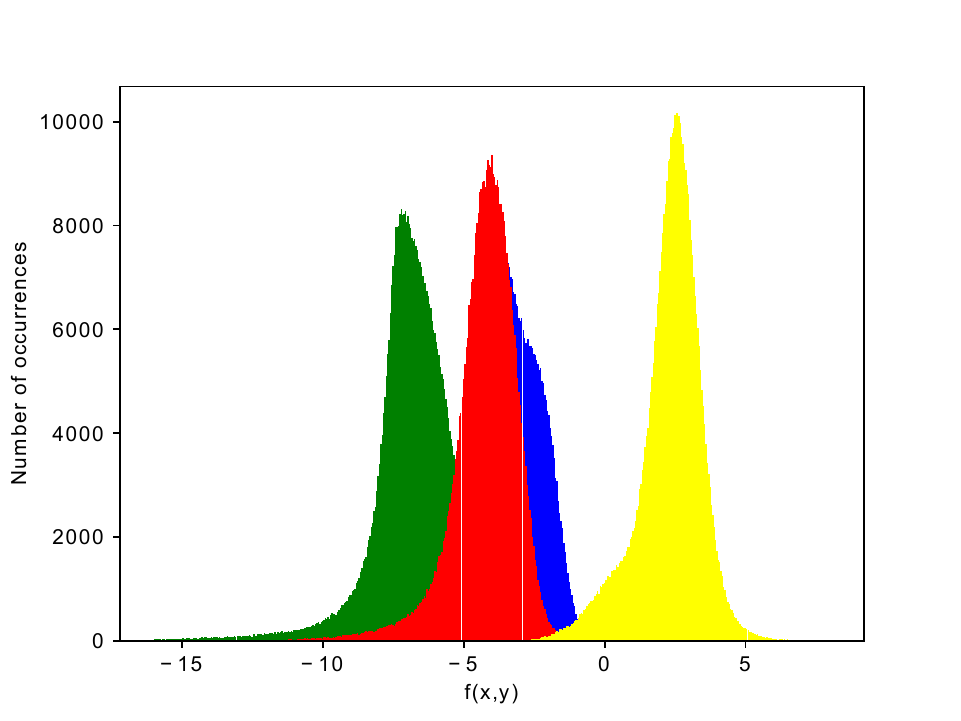}
  \caption{}
  \label{fig:explicit_200}
\end{subfigure}%
\caption{Label-to-image histogram results when validating 500 Cityscape images on a discriminator trained until epoch 200. Blue is Generated-conditional, Green is generated \ac, Red is real-\ac, Yellow is Real-conditional. (a) The trained baseline discriminator, (b) Optimal baseline discriminator, (c) \ac cGAN discriminator, (d) Optimal \ac cGAN discriminator. (a) and (c) are still learning, this indicate that there is no vanishing gradient or mode collapse~\cite{Arjovsky2017TowardsPM}. (b) doesn't detect conditionality since \ac real is classified as true (red) (d) succeeds to classify all modes correctly.}
\label{fig:dist}  
\end{figure*}

Preliminary conditionality evaluation follows the method presented in Section~\ref{sec:conditioneval} using \ac sets to evaluate an optimal discriminator. 
In a first part, experiments were carried out on the vanilla pix2pix cGAN with a discriminator PatchGAN architecture with $70\times70$ receptive field. The model was trained on the Cityscapes dataset~\cite{cordts2016cityscapes} for label-to-image translation with 2975 training images resized to $256\times256$. The pix2pix cGAN model is trained with the same hyperparameters as specified in the original paper~\cite{Isola2017ImagetoImageTW}. The evaluation histogram is calculated on the values of the last convolution layer of the discriminator ($f(x,y)$ of Eq~\eqref{eq:discriminator}) based on the $500$ validation images. Each sample from the last convolutional layer is composed of a $30\times30$ overlapping patches with one channel.The proposed approach is trained in exactly the same manner, with the only difference being the new objective function. 

Various tests were carried out to investigate the output distributions of each set of data for both the baseline architecture and the proposed method. The underlying accuracy of the implementation was first validated to ensure the accuracy reported in the original paper.  A histogram analysis was then performed for different levels of training including: training for 20, 100, 200 epochs and evaluating after each. In another experiment the discriminator was allowed to continue to converge for one epoch after fixing 20, 100 and 200 epochs of cGAN training. In particular, training is performed with the objective given in Eq~\eqref{eq:optimaldiscrim} and as proposed in~\cite{Arjovsky2017TowardsPM, Farnia2020GANsMH}. These results are plotted for each data pairing: real-conditional,  generated-conditional, real-\act and generated-\act in Figure~\ref{fig:dist}. 

Figure~\ref{fig:dist} (a) and (c) show that, since the discriminator did not converge, the generator is still learning with no vanishing gradient or mode collapse. In Figure~\ref{fig:dist} (b) the discriminator has been allowed to reach an optimal value by fixing the generator. The real \ac pairing is wrongly classified $99.9\%$ of the time, indicating that the discriminator has not learned conditionality. (d) Shows clearly four distinct distributions and shows the ability of the proposed approach to learn conditionality and correctly classify real \ac pairing $91.9\%$ of the time. Similar conditionality tests were performed for various discriminator alternative architectures, including using a separate/shared network for $\mbf{x}$ and $\mbf{y}$ and early/late/at-each-layer fusion. In all cases, conditionality was not learned.

These results strongly suggest that classic cGAN is unable to learn conditionality and that the spectacular results obtained by cGAN architectures are largely due to higher a level style constraints that are not specific to the input condition variable, since swapping condition variables produces no effect. The proposed histogram test allows to demonstrate the ability of the discriminator to classify the various underlying classes of data and shows their statistical distribution.

\subsection{Image-to-depth}
In this setting, the pix2pix model is trained on the NYU Depth V2 dataset~\cite{Silberman:ECCV12} to predict depth from monocular 2D-RGB images only. The official train/validation split of $795$ pairs is used for training and $694$ pairs are used for validation. The dataset images are resized to have a resolution of $256\times256$. The experiment is repeated 6 times, and the mean and standard deviation are reported.

\begin{figure*}[]
    \centering
    \includegraphics[width=0.9\textwidth]{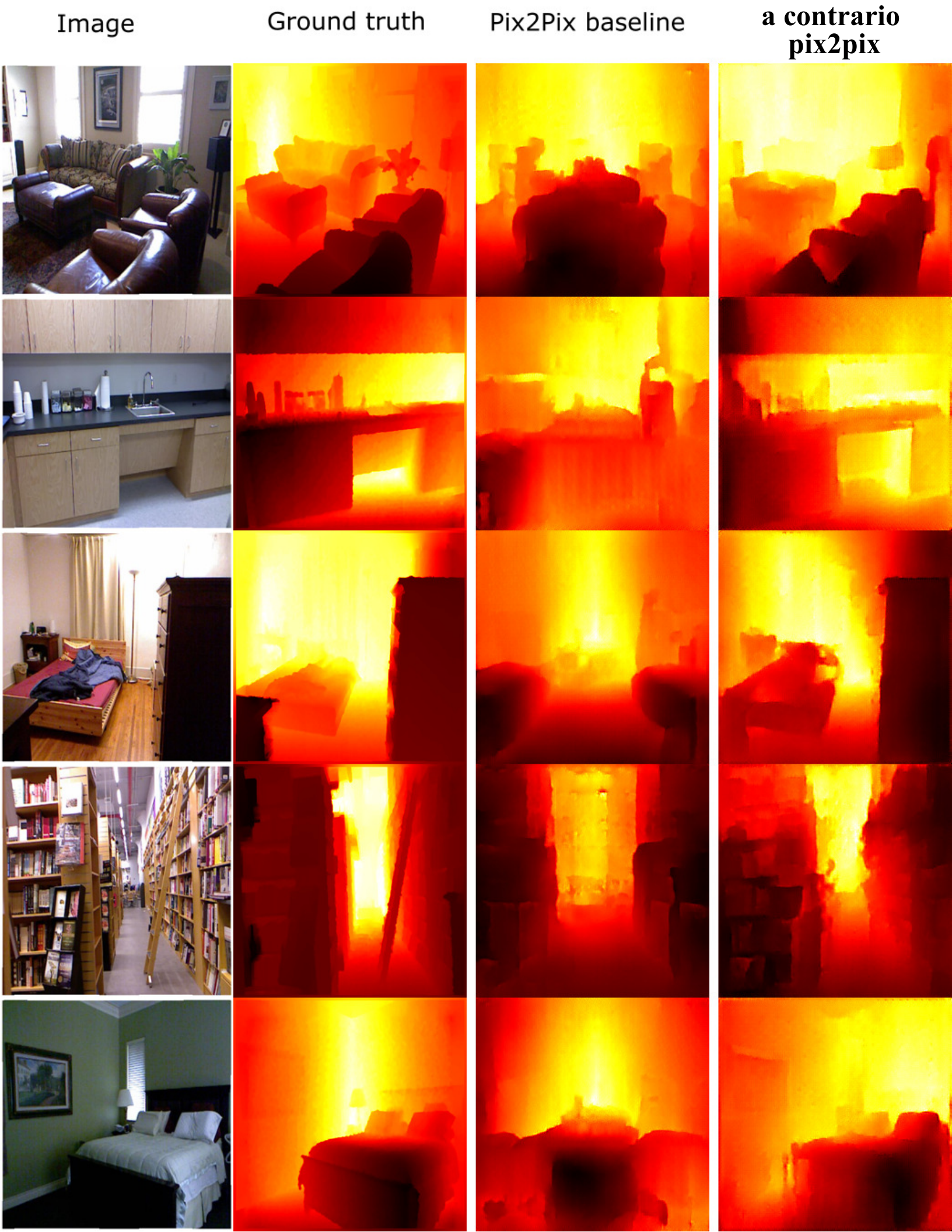}
    \caption{Qualitative results for depth prediction. The \ac cGAN shows better performance and more consistent prediction with respect to the input. The first row shows a case of mode collapse for the baseline as it ignores completely the input.}
    \label{fig:depth_qualitative}
\end{figure*}

Table~\ref{tab:Depth_stat} shows the comparison of the two models across different metrics. Clearly, the \ac cGAN reaches a better performance with log RMSE $0.3036$ versus $0.3520$ for the baseline (the mean is reported here). The qualitative results are shown in Figure~\ref{fig:depth_qualitative}.

\begin{table} 
    \centering
    \begin{tabular}{|c||c|c|c|c|}
         \hline
         Method &  RMSE $\log$ & si$\log$ & $\log_{10}$ & abs rel \\
         \hline
         baseline & 0.3520$\pm$ 0.0016 & 28.54$\pm$ 0.1932  & 0.1247 $\pm$ 0.0005 & 0.3318$\pm$ 0.0026 \\
         \hline
         \ac & \textbf{0.3036}$\pm$ 0.0055  & \textbf{23.51}$\pm$ 0.1932 & \textbf{0.1079} $\pm$ 0.0021 & \textbf{0.2868} $\pm$ 0.0093\\
         \hline
    \end{tabular}
    \vspace{0.2cm}
    \caption{	 Monocular Depth prediction experiments were repeated on the baseline and \ac cGANs 6 times with different seeds. The mean and standard deviation are reported for each metric. The results shows that the \ac cGAN outperforms the baseline on the depth metric~\cite{eigen2014depth}.
    }
    \label{tab:Depth_stat}
    
\end{table}

For the classical cGAN, the discriminator is optimized only to distinguish real and generated samples, its decision boundary is independent of the conditional variable. The baseline cGAN architecture will not penalize the generation of outputs belonging to the target domain, but that do not correspond to the input (\ie not conditional pair). Not only does this leave the generator with a larger search space (the generator is less efficient), but it can allow mode collapse, whereby the generator always produces the same output. The \ac loss explicitly avoids this by penalizing unconditional generation. As observed in the qualitative results, both methods generate smooth and depth that resemble the distribution of an indoor depth. However, for the classical cGAN baseline, the depth output is not consistent to the conditioning input. The model hallucinate a room, it has the ability to freely invent an output that has the distribution of the room depth map. The acontario method enforces the conditionality explicitly, and it is able to generate an accurate and a consistent input-output.

\subsection{Label-To-Image translation}
\label{sec:lb-to-im}

\begin{figure*}
\centering    
\begin{subfigure}{0.3\textwidth}
  \centering
  \includegraphics[width=\linewidth]{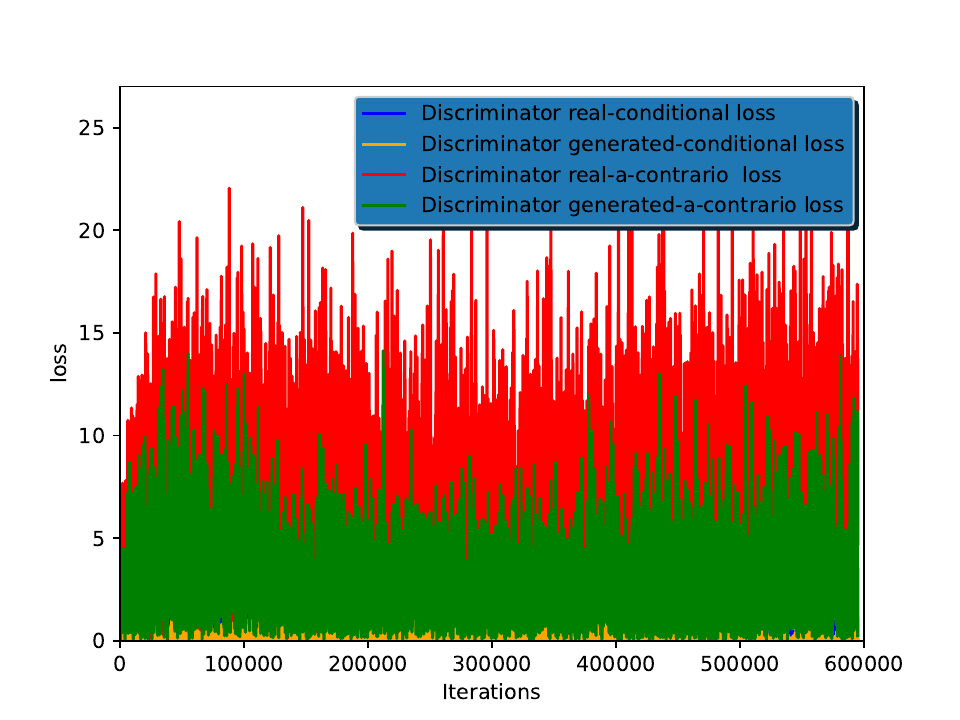}
  \caption{}
  \label{fig:cityscape_normal}
\end{subfigure}%
\begin{subfigure}{0.3\textwidth}
  \centering
  \includegraphics[width=\linewidth]{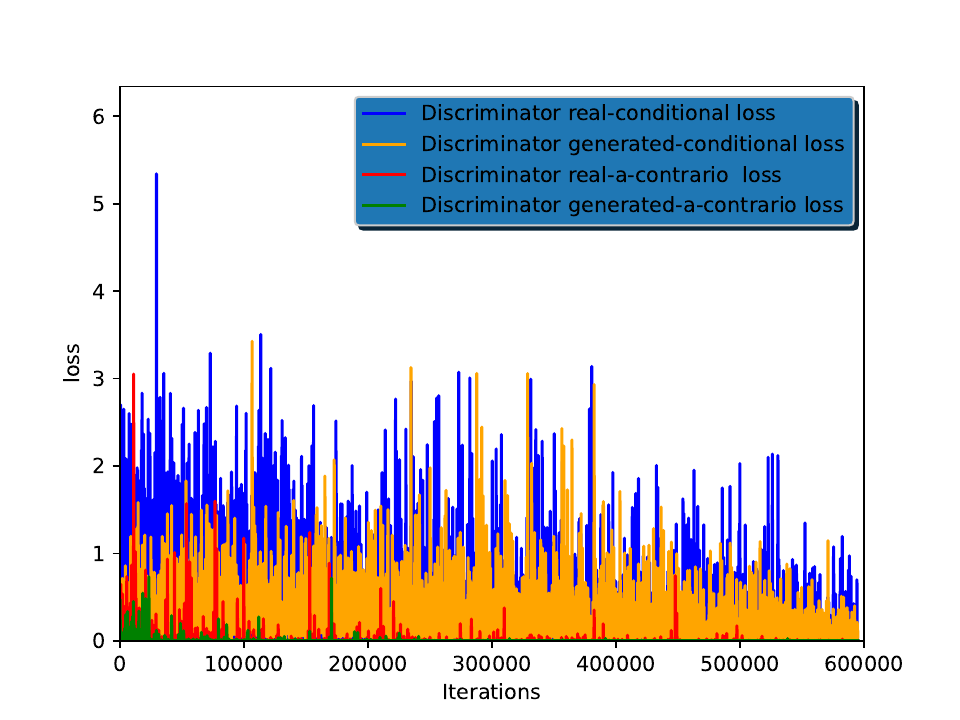}
  \caption{}
  \label{fig:cityscape_fixed}
\end{subfigure}%
\begin{subfigure}{0.3\textwidth}
  \centering
  \includegraphics[width=\linewidth]{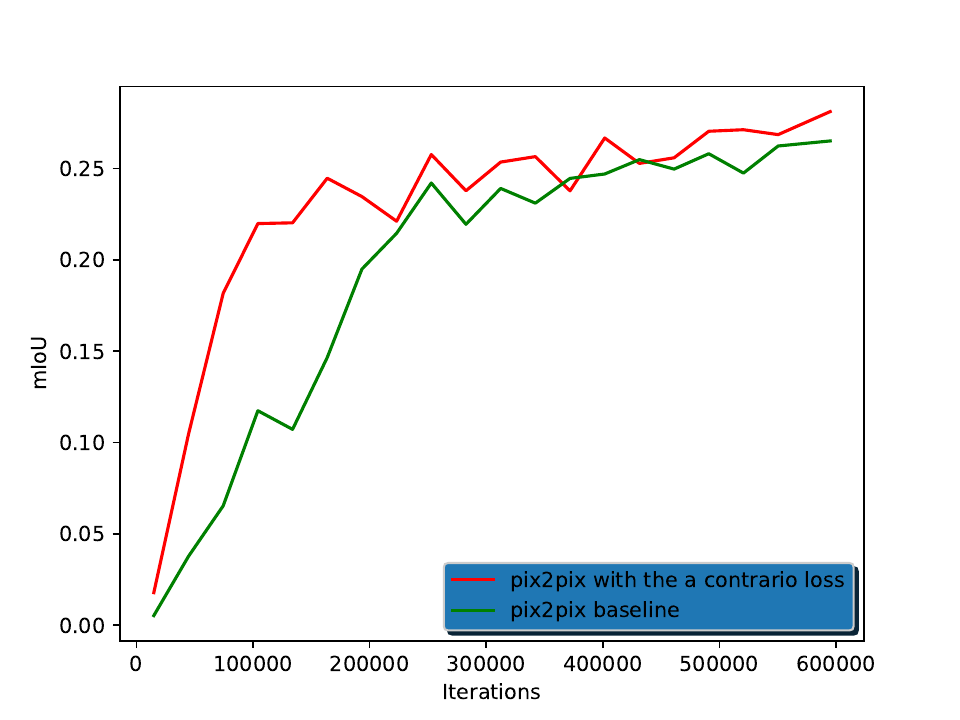}
  \caption{}
  \label{fig:miou}
\end{subfigure}%
\caption{ Comparison of the proposed approach on the Cityscape label-to-image training set. (a) The loss function for each set of data-pairing for the baseline cGAN method (\ac are for evaluation only). (b) The loss function for each set of data-pairing for the \ac cGAN method. (c) The evolution of the mIOU for both methods, performed on the validation dataset. It can be seen in (b) compared to (a) that the \ac loss converges to 0 rapidly for the proposed approach. In (c) the proposed approach is much more efficient and converges much faster and with higher accuracy.}
\label{fig:mIoU}
\end{figure*}

Generating realistic images from semantic labels is a well suited task to evaluate the effect of the \ac at a high level, since many images can be potentially generated for each semantic class label. Figure~\ref{fig:mIoU}(c) shows a comparison of the mIoU for the baseline pix2pix model and proposed pix2pix model with the additional \ac loss. It can be observed that the \ac cGAN converges faster than the baseline. The mIoU of the model with \ac at iteration $163$k is $24.46$ whereas the baseline is $14.65$. The mIoU oscillates around that value for the \ac model, indicating that the model has converged. After $595$k iterations, the mIoU for both models are very close $28.28$ and $26.41$. It is worth noting that evaluating using real images yields $29.6$. The convergence is reported for the generator where the computational cost is exactly the same. the \ac loss is specific only to the discriminator and adds a small computational cost. By restricting the search space of the generator to only conditional pairs, the generator's convergence is faster.

Table~\ref{tab:high_resolution} shows a comparison of different architectures with and without \ac augmentation. For a fair comparison, all the networks are trained from scratch and the same hyperparameter are used. The \ac loss is the only difference between the two networks. The batch size for SPADE is 32 and 16 for CC-FPSE. Through explicitly enforcing the conditionality with \ac examples, the discriminator learns to penalize unconditional generation achieving better results.  

Moreover, Figure~\ref{fig:mIoU}(a) and Figure~\ref{fig:mIoU}(b) show the comparison of the losses of the discriminator for both models on this dataset. The baseline is trained with only conditional pairs. The \ac data pairs are plotted to assess the ability of the discriminator to learn the conditionality automatically. The \ac losses remain high for the baseline and converge to $0$ for the proposed \ac cGAN. Figure~\ref{fig:dist} presented the histogram results for this experiment showed that the proposed approach better models conditionality

\begin{table}
    \centering
    \resizebox{0.8\textwidth}{!}{\begin{tabular}{|c||c|c|c c|}
         \hline
         Method & Resolution & FID & mIoU & Pixel accuracy (PA) \\
         \hline
         pix2PixHD & $256\times512$  & 66.7 & 56.9 & 92.8 \\
         \ac pix2pixHD &  $256\times512$ & \textbf{60.1}& \textbf{60.1} & \textbf{93.2}\\
         \hline
         SPADE & $256\times512$ & 65.5 & 60.2 & 93.1\\
         \ac SPADE & $256\times512$ & \textbf{59.9} & \textbf{61.5} & \textbf{93.7} \\
         \hline
         CC-FPSE & $256\times512$ & \textbf{52.4} & 61.8 & 92.8\\ 
         \ac CC-FPSE & $256\times512$ & 53.5 & \textbf{63.9} & \textbf{93.5} \\
         \hline
    \end{tabular}}
        \vspace{0.2cm}
    \caption{ 	A comparison of different architectures trained from scratch with and without \ac augmentation. The networks with \ac achieves better results with a mean improvement of $\Delta mIoU= +2.3$, $\Delta PA= +0.56 $, and $\Delta FID= -3.8$.}
    \label{tab:high_resolution}
\end{table}

\begin{figure*}
    \centering
    \includegraphics[width=0.8\textwidth]{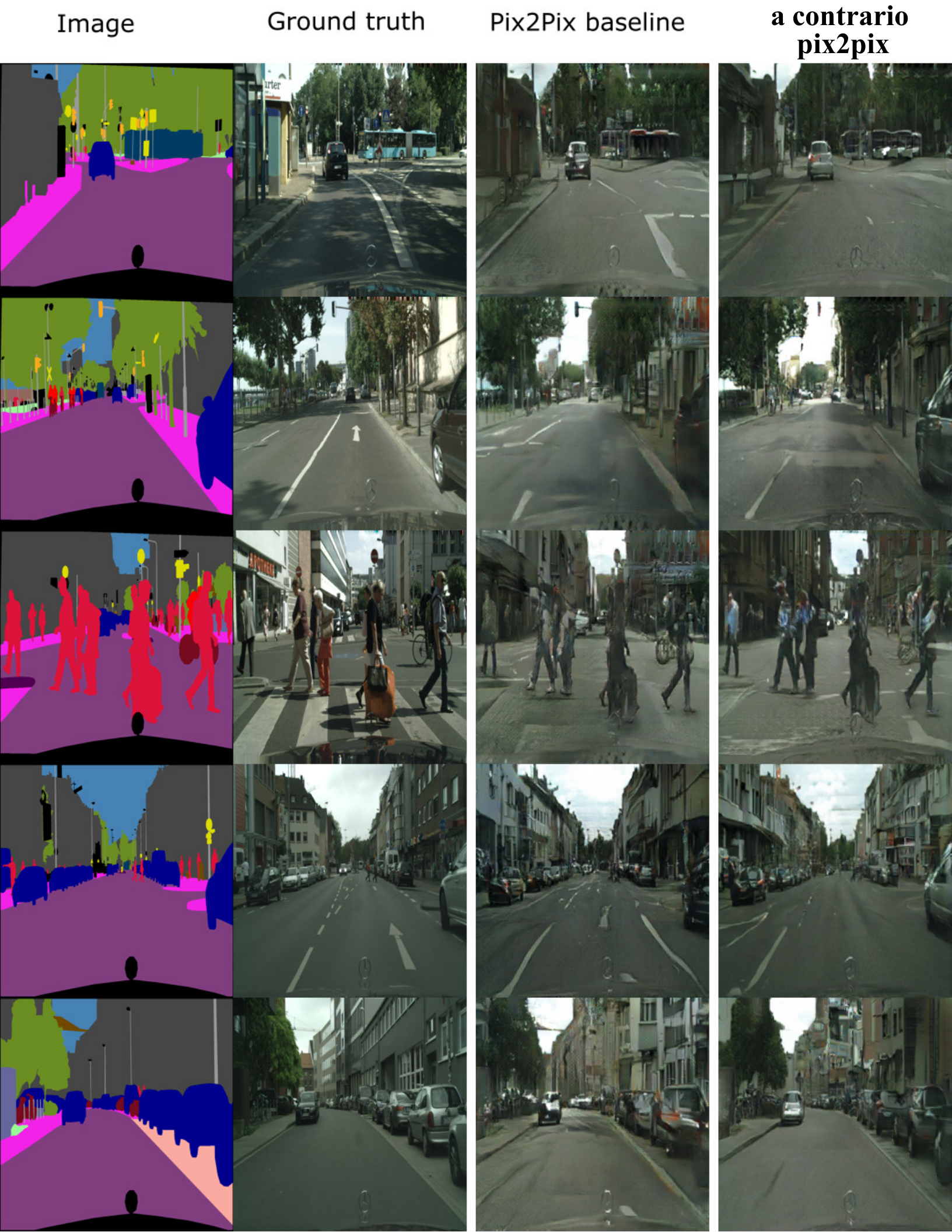}
    \caption{Qualitative results of Cityscapes label-to-image synthesis. In line with the quantitative results reported in Section \ref{sec:lb-to-im}, the qualitative results show better results for the \ac in comparison to the baseline. }
    \label{fig:cityscapes_qualitative}
\end{figure*}

\begin{figure}
    \centering
    \includegraphics[width=0.8\textwidth]{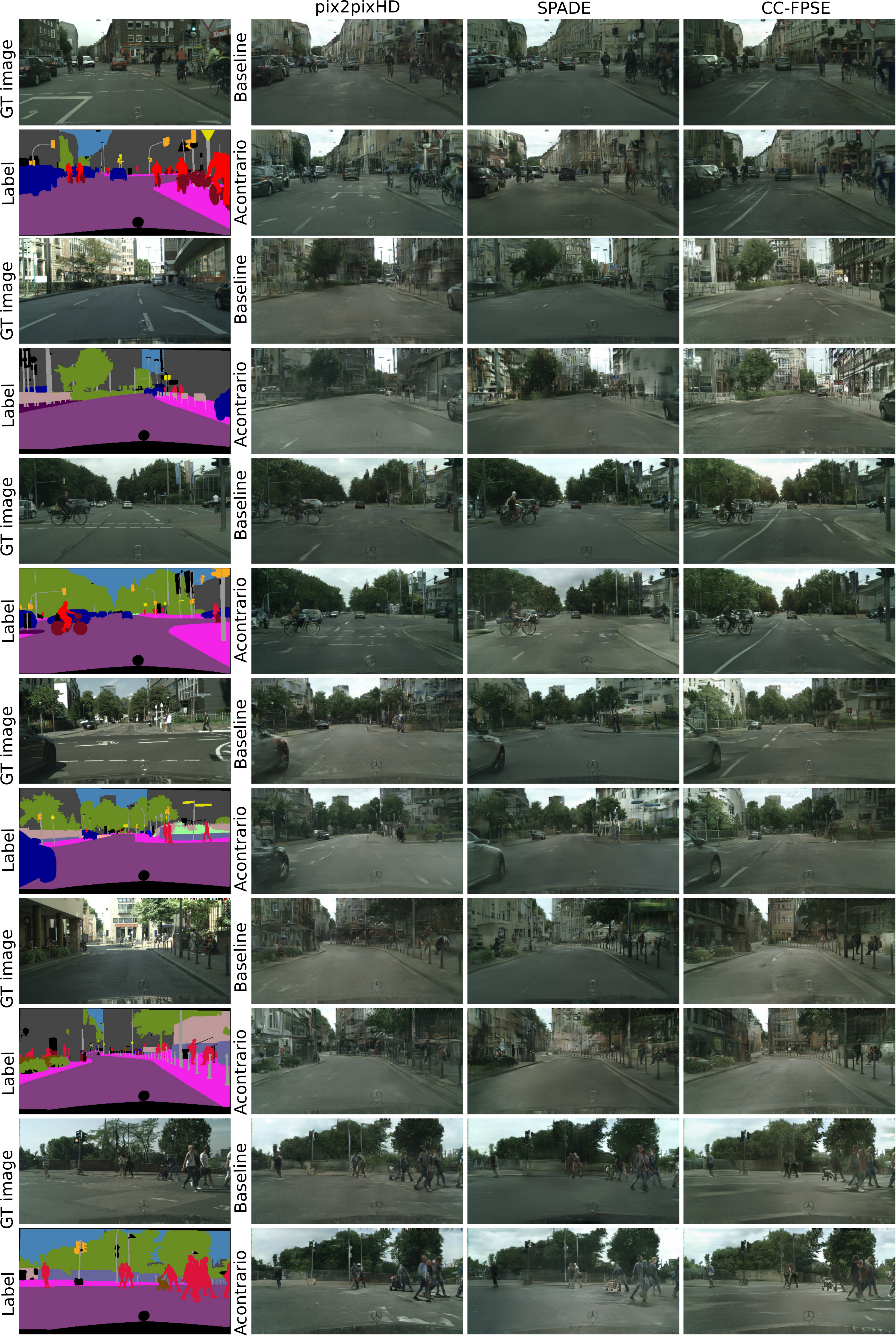}
    \caption{Qualitative comparison between different state-of-the-art methods for label-to-image trained and tested on Cityscapes\cite{cordts2016cityscapes} dataset. As observed, CC-FPSE baseline is the best baseline among classic cGAN. The \ac improves all the baseline and the best model among the 6 models is \ac CC-FPSE}
    \label{fig:state_of_art_comparison}
\end{figure}

\subsection{Single-label-to-image}
The generality of the proposed \ac cGAN can also be demonstrated by showing that it also improves architectures other than image-to-image. An example of a different task is conditioning the generated image on a single input class-label as in~\cite{Brock2019LargeSG,Miyato2018cGANsWP,Odena2017ConditionalIS,Karras2020TrainingGA,Kang2020ContraGANCL}. This different architecture is of interest because many new methods for improving cGANs are often tested on this task. Unfortunately, these methods are mainly evaluated on the FID~\cite{Heusel2017GANsTB} and IS~\cite{Salimans2016ImprovedTF} scores. As stated earlier, these metrics measure the quality/diversity and they favor models that memorise the training set~\cite{gulrajani2020towards}. They have not been designed to evaluate conditionality and therefore not sufficient for the purpose of this chapter. 
Despite that, these criteria are still important for evaluating the quality of GANs, however, an additional criterion is required for testing conditionality. 

Here a simple conditionality test is proposed specifically for "single label"-to-image generation tasks based on a pretrained Resnet-56~\cite{he2016deep} classifier trained on CIFAR-10~\cite{cifar}. BigGAN~\cite{Brock2019LargeSG} was selected as the baseline. Since BigGAN uses the Hinge-loss~\cite{lim2017geometric}, the \ac loss is adapted as follows:
\begin{align}
        \mathcal{L}_{D}=  -&\mathbb{E}_{\mbf{x} \sim p(\mbf{x}) , \mbf{y} \sim p(\mbf{y|x})}\big[min(0, -1 + D(\mbf{x},\mbf{y})]\big] -  \mathbb{E}_{\mbf{x}\sim p(\mbf{x})}\big[min(0,-1 - D(\mbf{x},G(\mbf{x}))]\big]  \notag  \\  
        -&\mathbb{E}_{\mbf{\tilde x} \sim p(\mbf{\tilde x}) , \mbf{y} \sim p(\mbf{y})}\big[min(0,-1-D(\mbf{\tilde x},\mbf{y}))\big] - \mathbb{E}_{\mbf{\tilde x}\sim p(\mbf{\tilde x}), \mbf{x} \sim p(  \mbf{x})}\big[min(0,-1 - D(\mbf{\tilde x},G(\mbf{x})))\big] \notag \\
        \mathcal{L}_{G}= - &\mathbb{E}_{\mbf{x}\sim p(\mbf{ x})} D(\mbf{x},G(\mbf{x}))
\label{eq:HingleLoss}
\end{align}
Both models are trained from scratch on CIFAR-10~\cite{cifar} dataset using the hyper-parameter specified in~\cite{Brock2019LargeSG}. The conditionality is tested by generating 10k images for each label(100k images in total) and calculating the accuracy. The results\footnote{The Pytorch IS and FID implementations were used for comparison}are shown in Table~\ref{tab:biggan_table}.
\begin{table}   
    \centering
    \begin{tabular}{|c||c|c|c|c|}
         \hline
         Method & IS score & FID score & Acc  \\
         \hline
         BigGAN\cite{Brock2019LargeSG} & 8.26 $\pm$ 0.095   & 6.84  & 86.54 \\
         \hline
         \ac BigGAN & \textbf{8.40} $\pm$ 0.067 &   \textbf{6.28} & \textbf{92.04} \\
         \hline
    \end{tabular} 
        \vspace{0.2cm}
    \caption{ A comparison of BigGAN~\cite{Brock2019LargeSG} with and without the \ac GAN. The network with \ac achieves significantly better results with an improvement of $\Delta Acc= +5.59$, $\Delta IS= +0.14 $, and $\Delta FID= -0.56$.
    }
    \label{tab:biggan_table}
\end{table}

The conditionality improved significantly over the baseline with $\Delta Acc=+5.59$ and the quality also improved with $\Delta FID= - 0.56\:$, $ \Delta{IS}=+0.14$. Similar to the observation made before \ac enforces the conditionality without compromising the quality. A failure mode of the lack of conditionality of the discriminator is class leakage : images from one class contain properties of another. While is it not easy to define a proper metric for such failure mode, it is shown that using the \ac loss the classification was improved and therefore the generation is better constrained and does not mix class properties. This result shows that \ac GAN also improves on a different SOTA task and confirms again that conditionality is an overlooked factor in current SOTA metrics.

\subsection{Image-to-label segmentation}
\begin{figure*}
    \centering
    \includegraphics[width=0.8\textwidth]{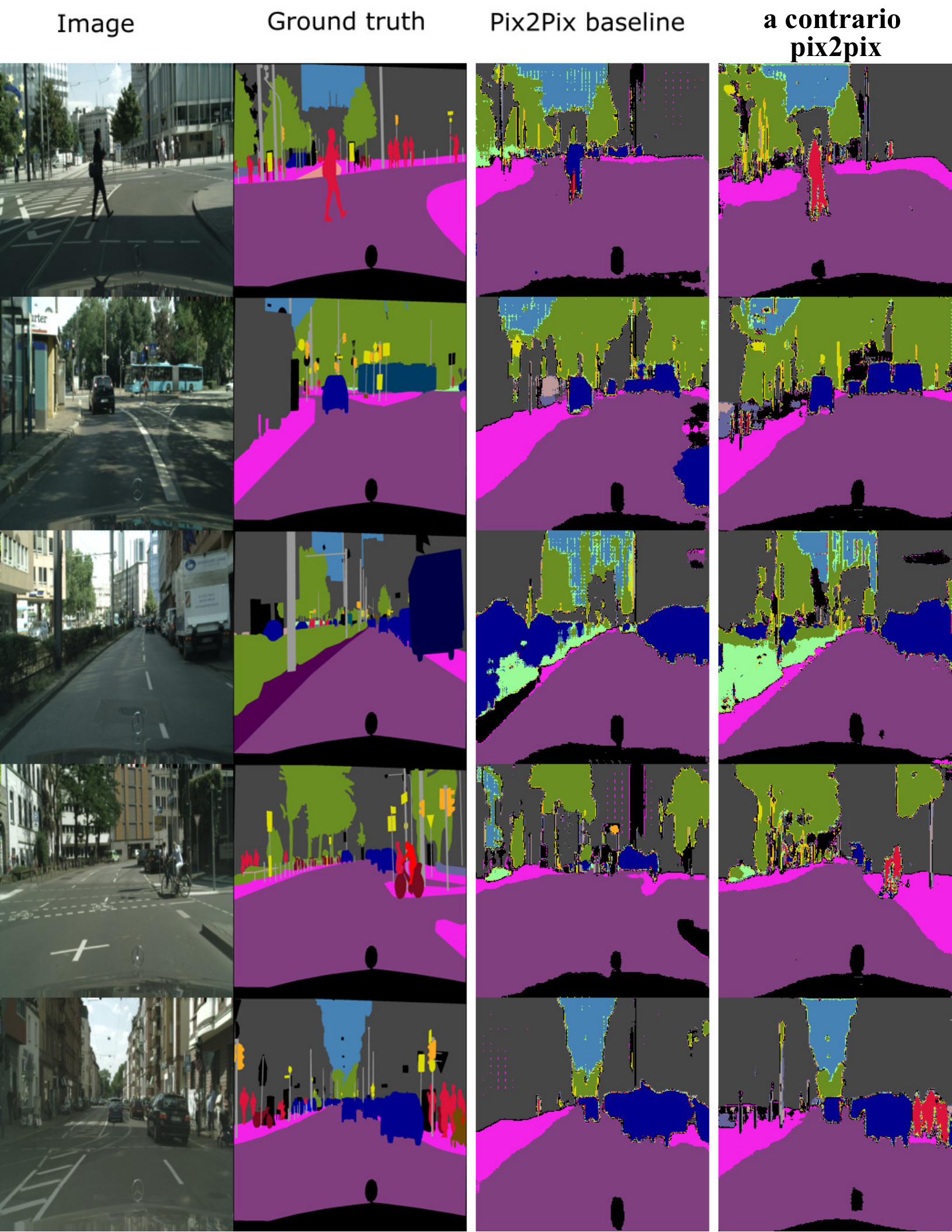}
    \caption{Qualitative results of Cityscape image-to-label task. It can be seen that the baseline model hallucinates objects. For instance, in the second row, the baseline hallucinates cars while the \ac cGAN segments the scene better. In the first row, the baseline wrongly classifies the pedestrian as a car. While training the model, the discriminator does not penalize the generator for these miss-classifications}
    \label{fig:seg_qualitative}
\end{figure*}

Image-to-label is a simpler task compared to depth prediction and label-to-image prediction as the goal of the generator is to transfer from a high-dimensional space to a lower-dimensional space. Furthermore, the evaluation is simpler since the image mask does not have multiple solutions and it is not necessary to use an external pre-trained segmentation network for comparison as in the case of label-to-image translation. It is worth mentioning that pix2pix is trained to output 19 classes as a segmentation network and is not trained as an image-to-image network as it is often done in cGAN architectures. FCN~\cite{long2015fully} trained on~\cite{Isola2017ImagetoImageTW} obtains $21.0$ mIoU. The performances are shown in Table~\ref{tab:seg_table}. The training was unstable. However, the \ac cGAN shows superior mIoU performance with $19.23$ versus $15.97$ for the baseline model. Figure~\ref{fig:seg_qualitative} shows the qualitative results of the both models. It can be observed that the model baseline has invented labels that are not specified by the input. Training with \ac helps the discriminator to model conditionality. Thus, the generator search space is restricted to only conditional space. The generator is penalized for conditionality even if the generation is realistic.   
\begin{table}  
    \centering
    \begin{tabular}{|c||c|c|c|c|}
         \hline
         Method & Pixel accuracy (PA) & Mean Acc & FreqW Acc  & mIoU  \\
         \hline
         Baseline & 66.12 & 23.31  & 53.64  & 15.97 \\
         \hline
         \ac & \textbf{72.93} &  \textbf{26.87} & \textbf{60.40} & \textbf{19.23} \\
         \hline
    \end{tabular} 
        \vspace{0.2cm}
    \caption{ Comparison on the Cityscapes dataset validation set. The proposed method consistently obtains more accurate results and finishes with a largely different score at the end of training with mIoU of $19.23$ versus for the baseline $15.97$.}
    \label{tab:seg_table}
\end{table}

\section{Discussion}

This chapter has presented a new method called \textbf{\ac~cGAN}, which explicitly models conditionality for both parts of the adversarial architecture through a novel \ac loss. This loss involves training the discriminator to learn unconditional (adverse) examples. The \ac learning approach restricts the search space of the generator to conditional outputs using adverse examples. Extensive experimentation has demonstrated significant improvements across various tasks, datasets, and architectures.

One limitation of these models is their reliance on paired datasets. The requirement of paired data may limit the applicability of the models in scenarios where such data is not readily available. This is the case in domain adaptation, where often the input is unpaired dataset of source and target domain. However, this method could potentially be adapted for domain adaptation, it is important to note that it is not within the scope of this thesis, which focuses on performing depth prediction.

It is worth mentioning that even with recent advances in transformers and diffusion models, the problem of hallucination remains crucial, particularly in the development of generative large language models where accuracy is required. The proposed method in this thesis could potentially address this issue and provide insights for improving generative language models. 

Moreover, it is important to acknowledge that inconsistency is not always a bug; it can be a desirable feature, especially in domains such as digital art where creativity is required. The ability of generative models to introduce controlled inconsistency can enhance their creative output.

\chapter*{}


\chapter{Image-to-depth inference}
\label{ch:ch4}
In the previous chapter, the objective of the thesis was approached with a broad perspective, exploring depth and other modalities and investigating generalization through domain adaptation. Building upon that, the following chapters delve into a detailed analysis of self-supervised depth prediction. The aim of this chapter is to develop a model capable of inferring depth from a single image using a self-supervised monocular approach, which poses a substantial challenge due to the inherent ambiguity in converting 2D images to 3D representations. The goal is to address a critical limitation in existing methods, which assumes a static scene. Many current approaches assume that the scene being captured remains unchanged over time with no significant object movements or variations. While this simplification facilitates depth training, it fails to account for the dynamics and temporal changes observed in real-world scenes.

To overcome the aforementioned limitation, an innovative approach is proposed in this chapter. The proposed approach relaxes the assumption of a rigid scene by inferring the pose of dynamic objects and compensating for their dynamics during model training. As a result, the performance of depth inference is enhanced. By incorporating the dynamic nature of scenes, this method represents a significant advancement in monocular self-supervised depth inference, thereby opening up possibilities for more advanced forecasting techniques.

The chapter starts by emphasizing the benefits and advantages of self-supervised learning. Utilizing unlabeled data through this approach proves advantageous due to its accessibility and cost-effectiveness compared to labeled data. The practicality of self-supervised learning makes it an attractive choice in different situations. The chapter offers a detailed introduction to self-supervised learning for depth prediction, covering problem formulation, self-supervision techniques. Lastly, the proposed method is presented, along with a discussion on the results, limitations, and future prospects.

This chapter is based on the following publication: 
 \begin{itemize}
      \item \textbf{Journal paper:} Boulahbal Houssem Eddine, Adrian Voicila, and Andrew I. Comport. "Instance-aware multi-object self-supervision for monocular depth prediction." IEEE Robotics and Automation Letters 7.4 (2022): 10962-10968.
 
    \item \textbf{Conference paper:} Boulahbal Houssem Eddine, Adrian Voicila, and Andrew I. Comport. "Instance-aware multi-object self-supervision for monocular depth prediction." 2022 35th International Conference on Intelligent Robots and Systems (IROS). IEEE, 2022.
 \end{itemize}

\section{Supervised versus self-supervised approaches}
In recent years, the field of deep learning has known a tremendous an exponential growth that has revolutionized the field of artificial intelligence. Since the unprecedented success of deep learning methods on the ImageNet~\cite{deng2009imagenet}, a plethora of expert models that can learn from massive amounts of labeled data were developed. Since then, there has been a rapid evolution of these models that have demonstrated impressive performance on a wide range of tasks.
However, the performance and generalization of these models are heavily dependent on the quality, quantity, and diversity of the training data.

\begin{figure}
    \centering
    \includegraphics[width=0.75\textwidth]{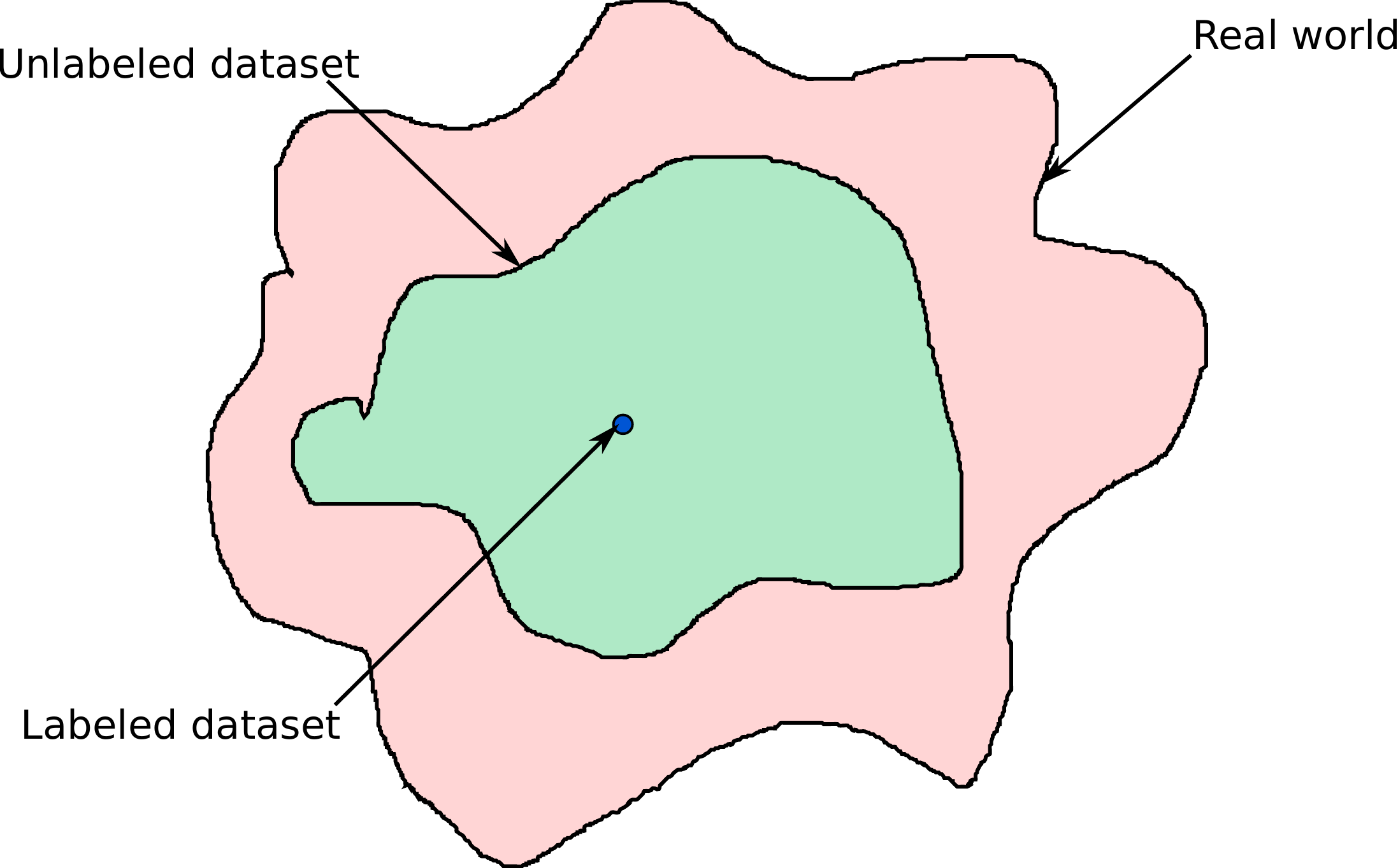}
    \caption{The size of labeled dataset represent only a tiny portion of the unlabeled dataset, which in turn represent portion of the real world}
    \label{fig:labled_unlabled}
\end{figure}

As illustrated in~\reffig{labled_unlabled}, the size of labeled dataset represent only a tiny portion of the unlabeled dataset, which in turn represent a tiny portion of the real world. Training only on the labeled dataset, yields models that are overly specialized, producing models that are susceptible to poor generalization. Furthermore, even if the model can successfully extrapolate beyond the labeled dataset, it will still only represent a small fraction of the possible scenarios that exist in the real world. Consequently, a model trained only on the labeled datasets is likely to suffer from biases, domain shifts, and poor generalization when confronted with extreme scenarios that were not part of the training dataset. 

Despite the remarkable success of supervised learning in deep learning, there are limitations to the extent that AI can progress solely based on this approach. One of the most significant challenges facing supervised learning is the difficulty of obtaining and labeling large amounts of data, especially for real-world problems that are complex and diverse, such as autonomous driving applications. This obstacle necessitates the development of alternative approaches that can learn from directly from unlabeled data. Labeling everything is just impossible.

One inspiration to learn without labels is the human intelligence. humans have the ability to learn directly through observation: Human beings possess the ability to formulate hypotheses based on experiences, conduct experiments to test these hypotheses, observe the results, and ultimately derive a conclusion. Similarly, it is also possible to make machines learn solely from the data, where the learning obtains the supervisory signals from experience,\ie data, only. This is known as self-supervision. One good example that illustrates the potential of these methods is  the GPT~\cite{openai2023gpt4} family. These models have been trained to predict the next token (word or image patch) in a sequence, leveraging a huge dataset of text crawled from the internet, arXiv papers, and books. These self-supervised models have shown impressive results. They can even outperform supervised models on several tasks with zero-shot (without being trained explicitly to perform these tasks). Furthermore, some results suggest that the quality of self-supervised representations increase logarithmically in proportion to the volume of unlabeled pretraining data used~\cite{goyal2019scaling}. This means that the performance may advance over time as advances in computational capacity and data acquisition enable ever-larger datasets to be utilized without the necessity of manually labeling new data.

As a summary, the reliance on labeled data in deep learning poses challenges for model performance and generalization. Limited labeled datasets result in specialized models with poor adaptability to real-world scenarios. Obtaining and labeling large amounts of data is difficult, hindering supervised learning progress. Self-supervised learning, inspired by human intelligence, offers a promising alternative. The following section delve into applying these methods on the depth modality.

\section{Depth prediction with self-supervised methods}
\label{sec:ssl}

Self-supervised depth prediction refers to methods that only use images for input and supervision, without the need for ground-truth labels. These methods are becoming increasingly popular due to their practicality, as they do not require manually labeled training data. 
\subsection{Problem formulation}
The aim of monocular depth prediction is to learn an accurate depth map through the mapping $\mbf{D}_t = f(\img{t-k:t}; \bm{\theta})$ where $\img{t-k:t} \in \mathbb{R}^{k \times W\times H\times3}$ is $k$ context images. $\mbf{D}_t$ is the target depth. $\bm{\theta}$ are the network parameters.  
In self-supervised learning, this model is trained via novel view synthesis by reverse warping a set of source frames $\mbf{I}_{s}$ into the target frame $\mbf{I}_{t}$ using the learned depth $\mbf{D}_{t}$ and the target to source pose $\pose{s}{t}$. The reverse warping $\mathcal{W}$ is used to reconstruct the image. It involves mapping pixels from one image to another image, where the pixel coordinates in the new image are computed based on the differentiable warping function applied to the original image the mapping is defined as:
\begin{equation}
        \label{eq:warpin_with_op}
        \widehat{\pt}_{s} \sim \pi ( \mbf{K} \pose{s}{t} H (\mbf{D}_{t}  \mbf{K}^{-1} \pt_{t}))
\end{equation}

Where $H$ is the homogenous transformation operator and the $\pi$ is the projection operator. For simplicity, these two operators are omitted: 

\begin{equation}
        \label{eq:warping_classic}
        \widehat{\pt}_{s} \sim \mbf{K} \pose{s}{t} \mbf{D}_{t}  \mbf{K}^{-1} \pt_{t}
\end{equation}
\begin{figure}
    \centering
    \includegraphics[width=\textwidth]{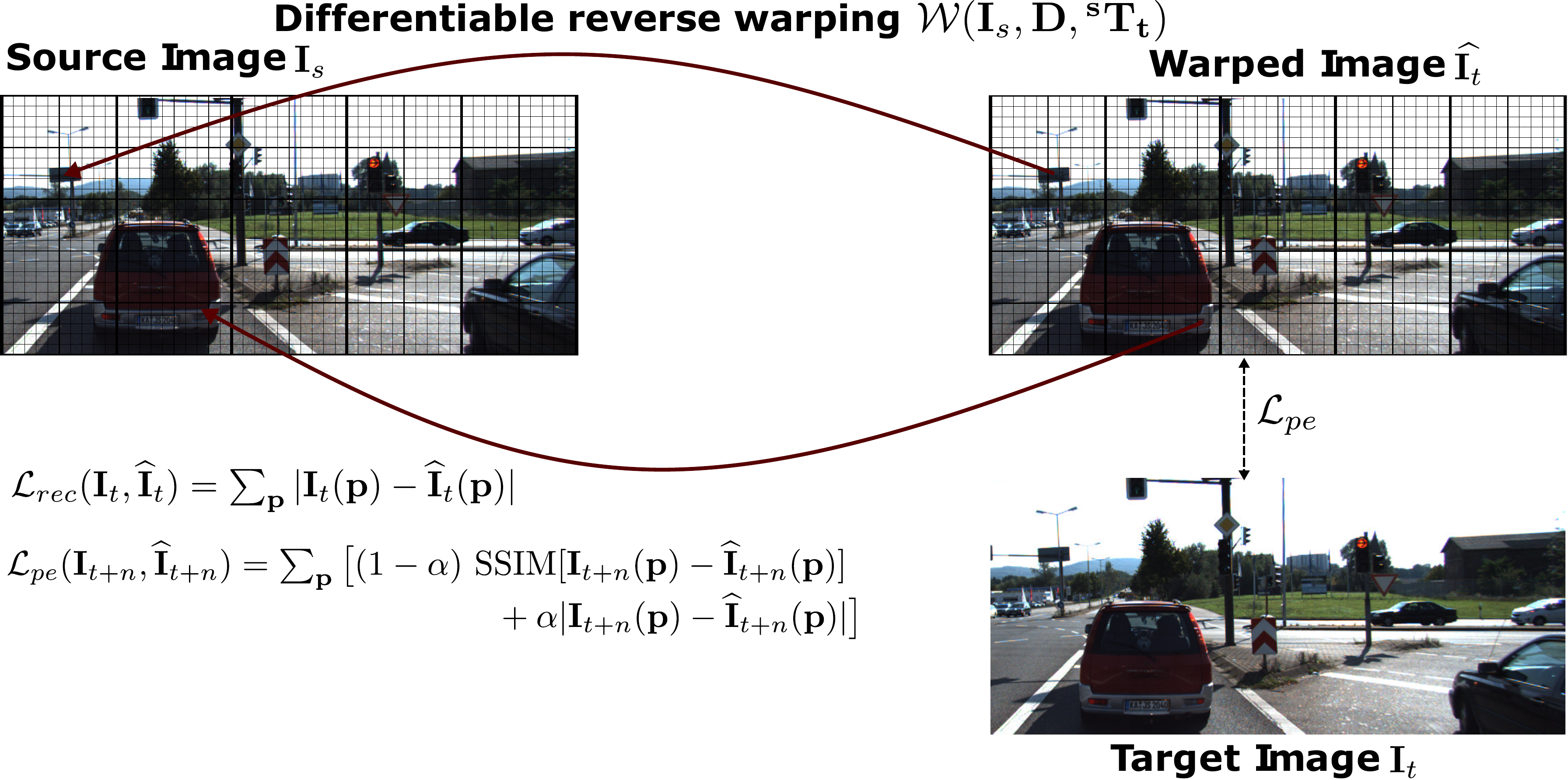}
    \caption{An illustration of the reverse (also called inverse) warping. Using the depth and the pose, the source image is warped into the target image. The self-supervised optimization is done using the photometric loss.}
    \label{fig:inver-warping}
\end{figure}
The image is reconstructed using the interpolation.~\reffig{inver-warping} shows the warping process. The points of the target image are back-projected using the camera parameters and the learned depth. The $\pose{s}{t}$ is applied to transform the point cloud. Finally, the point cloud is projected using the camera parameters. The target image is obtained using the interpolation. Therefore, by knowing the depth and pose, the mapping from the image $\mbf{I}_{s}$ is used to reconstruct $\Hat{\mbf{I}}_t$ through a bi-linear interpolation.

\subsection{Loss functions}
\label{sec:loss_function_ch3}
Let $\mathcal{L}$ be the objective function. The self-supervised setting casts the depth learning problem to an image reconstruction problem through the reverse warping. Thus, learning the parameters $\bm{\theta}$ involves learning $\widehat{\bm{\theta}} \in \bm{\Theta}$ that minimizes the following objective functions:
\begin{equation}
    \widehat{\bm{\theta}} = \argmin_{\bm{\theta} \in \bm{\Theta}} \frac{1}{n} \sum_{n} \mathcal{L} (f(\mbf{I}_t, \widehat{\mbf{I}}_t;\bm{\theta}))
    \label{eq:risk_minimzation}
\end{equation}
where $n$ is the number of training examples. There are several surrogate loss functions proposed to supervise the depth though image reconstruction, some of these losses are :
\begin{itemize}
    \item \textbf{Photometric loss: }
    
    Following~\cite{Zhou2017,Godard2019,Rares2020}, the photometric loss seeks to reconstruct the target image by warping the source images using the static/dynamic pose and depth. The $L_1$ loss is defined as follows:
    \begin{equation}
        \mathcal{L}_{rec}(\img{t}, \widehat{\mbf{I}}_{t}) = \sum_{\pt} | \img{t}(\pt) - \widehat{\mbf{I}}_{t}(\pt) |
    \end{equation}
    where $\widehat{\mbf{I}}_{t}(\pt)$ is the reverse warped target image obtained by~\refeq{warping_classic}. The L1 loss is a widely used loss function in computer vision tasks, and it is particularly useful for self-supervised depth prediction because it is robust to outliers. However, the L1 loss alone is not sufficient, as it does not take into account the structural similarity between the predicted image map and the ground-truth image.  

    \item \textbf{SSIM (Structural Similarity Index) }is used to improve the photometric loss for self-supervised depth prediction models. SSIM is a widely used metric for image quality assessment, and it measures the structural similarity between two images by comparing the luminance, contrast, and structure of the images. SSIM is particularly useful for self-supervised depth prediction because it is more sensitive to changes in the structure of the images than the L1 loss.
    
    Therefore, the photometric loss is defined as:
    \begin{equation}
    \begin{aligned}
      \mathcal{L}_\textup{pe}(\img{t}, \widehat{\mbf{I}}_{t}) = \sum_{\pt} \big[&(1 - \alpha) \textup{ SSIM}[ \img{t}(\pt) - \widehat{\mbf{I}}_{t}(\pt)]  \\
      & +  \alpha | \img{t}(\pt) - \widehat{\mbf{I}}_{t}(\pt) |\big]
    \end{aligned}
    \label{eq:ssim}
    \end{equation}
    \item \textbf{Depth smoothness: } An edge-aware gradient smoothness constraint is used to regularize the photometric loss. The depth map is constrained to be locally smooth through the use of an image-edge weighted $L_1$ penalty, as discontinuities often occur at image gradients. This regularization is defined as~\cite{heise2013pm}:
    \begin{equation}
    \begin{aligned}
      \mathcal{L}_{s}(D_{t}) = \sum_{p} \big[ &| \partial_x D_{t}(\pt) | e^{-|\partial_x \mbf{I}_{t}(\pt)|}    +\\
      &| \partial_y D_{t}(\pt) | e^{-|\partial_y \mbf{I}_{t}(\pt)|} \big] 
    \end{aligned}\label{eq:loss-disp-smoothness}
\end{equation}
\end{itemize}
In practice, to optimize the self-supervised depth prediction network, often, a L1 loss is used in combination with the SSIM and depth smoothness losses. This yields better results than using a single loss function alone.

\section{Dynamic object for self-supervised depth}
\label{sec:iros-intro}
Self-supervised monocular depth training methods presented in~\refsec{ssl} are based on the assumption of a rigid scene, meaning that the scene is static and the camera is moving. However, this assumption is often violated in real-world scenarios due to the presence of moving objects in the scene. This poses a challenge for depth inference models, as the motion of objects can significantly impact the accuracy of depth inference.
One potential solution to this issue is to mask the dynamic objects' points in the scene during training. This can be achieved through various methods such as: learned masking techniques~\cite{Zhou2017}, semantic guidance~\cite{Klingner2020} or auto-masking~\cite{Godard2019,Watson2021}. However, these methods only provide a workaround to the problem of non-rigid scenes, and they fail to utilize the information from moving objects that could be useful for further constraining depth inference.
To address the challenge of moving objects in the scene, various studies have proposed methods that explicitly incorporate information about moving objects into the depth inference models. For example, some studies have proposed methods that learn a per-object semantic segmentation mask and a motion field that account for the motion of the objects in the scene~\cite{Vijayanarasimhan,lee2021learning,xu2021moving}. Other studies have relied on optical flow to model the motion of objects in the scene~\cite{Ranjan2019,Yin2018}. While these methods are optimized for local rigidity, they do not take into account the different dynamics of different object classes. As a result, they may not provide as accurate depth inference as the methods that explicitly model the $6-$DOF motion of objects.

A proposition is made here to alleviate this assumption. Non-rigid scenes are learned by factorizing the motion into the dominant ego-pose and \textbf{a piece-wise rigid pose for each dynamic object} explicitly. Therefore, for static objects, only the ego-pose is used for the warping, while the dynamic objects are subject to two transformations using the motion of the camera and the motion of each moving object. The proposed method explicitly models the motion of each object, allowing accurate warping of the scene elements. 

In order to model the object motion in the scene, the proposed method makes use of the multi-head attention of the transformer network that matches moving objects across time and models their interaction and dynamics. This enables accurate and robust pose estimation for each object instance. The proposed method achieves SOTA results on the KITTI benchmark. In summary, the contributions of the method proposed in this chapter are:




\begin{itemize}
    \item A transformer-based network architecture that utilizes multi-head attention to match moving objects across time and accurately estimate their motion, enabling more robust and precise pose estimation for each object instance.
    \item An accurate and robust per-object pose is obtained by matching and modeling the interaction of the objects across time. 
    \item High quality depth inference, achieving competitive performance with respect to state-of-the-art results on the KITTI benchmark~\cite{Geiger2012CVPR}. 
    \item The demonstration that the KITTI benchmark has a bias favoring static scenes, and a method to test the quality of moving object depth inference.
\end{itemize}


\subsection{Related work}
\label{sec:iros-related}

Supervising the depth with a photometric loss is problematic when moving objects are present in the scene. This challenge has gained attention in the literature, a common solution is to disentangle the dominant ego-motion and the object motion.~\cite{chen2019self,Ranjan2019,Yin2018,hur2020self} leverage an optical flow network to detect moving objects by comparing the optical flow with depth-based mapping.
~\cite{lee2021attentive} learns a monocular depth in order to estimate the motion field as two stage learning.~\cite{Vijayanarasimhan} learns a per-object semantic segmentation mask and a motion field is obtained by factorization of the motion of each mask and the ego-motion.~\cite{safadoust2021self} addresses the object motion without additional labels by proposing a scene decomposition into a fixed
number of components where a pose is inferred for each component.~\cite{xu2021moving} relaxes the problem using local rigidity within a predefined window, and the motion of each window is predicted to account for moving objects.~\cite{luo2019every} leverages the geometric consistency of depth, ego-pose and optical flow and categorises each pixel as either rigid motion, non-rigid/object motion or occluded/non-visible regions.
A recent work that is closest to the proposed method is Insta-DM~\cite{lee2021learning}. In that method, the source and target images are masked with semantic masks and an object PoseNet is used to learn the pose from the masked RGB images. Alternatively, the method proposed in this chapter factorizes the motion into ego-motion and object-motion and exploits a transformer attention network to perform instance segmentation and learn a per-object motion.

\subsection{Problem formulation}



\begin{figure}
    \centering
    \includegraphics[width=\textwidth]{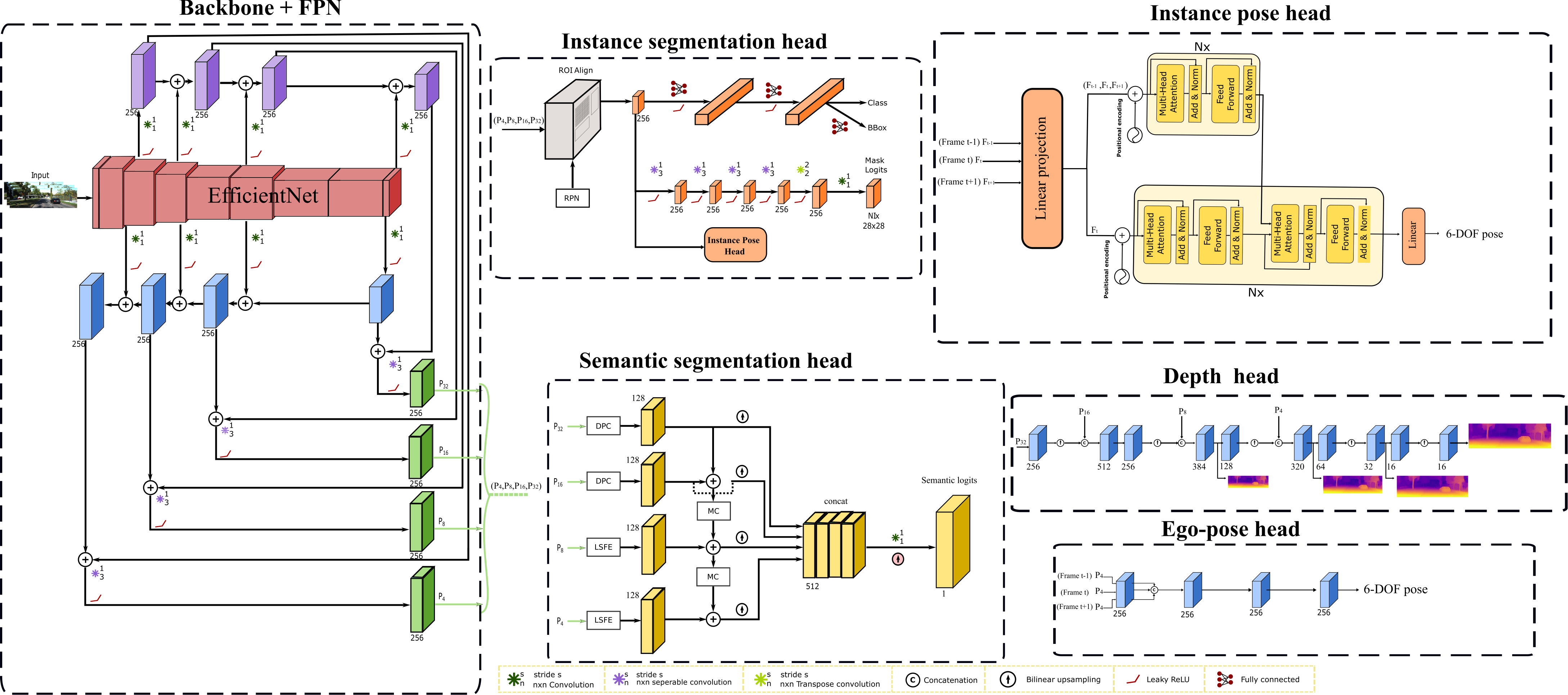}
    \caption{The proposed model architecture consisting of the EfficientNet backbone~\cite{tan2019efficientnet}, BiFPN~\cite{tan2020efficientdet}, the DPC~\cite{chen2018searching} semantic head, the MaskRCNN instance segmentation head~\cite{he2017mask}, the novel instance pose head, an ego-pose head and a depth head. During training, the FPN features ($P_4, P_8, P_{16}, P_{32}$) are extracted for the source $\img{t}$ and target frames $\img{t-1}$, $\img{t+1}$. These features are pooled using the proposals of the RPN and the ROI Align modules. The class, bounding box and instance mask heads use only the features of frame $\img{t}$. The Instance pose head uses both source and target frames as input. This head output a $6$ axis-angle parameters for each instance. Similarly, the ego-pose head uses the both source and target frames $P_4$ FPN' features as input. This head output a $6$ axis-angle parameters for the ego-pose. The depth head input the FPN features of the source frame $\img{t}$ and output a multiscale depth.}
    \label{fig:iros_architecture}
\end{figure}

In the method proposed in this chapter, rather than enforcing the rigid scene assumption, a proposition is made to alleviate this assumption. For each pixel, a \textbf{global rigid-scene pose} and a \textbf{piece-wise rigid pose} for each dynamic object is learned. This is more precise and consistent with the non-rigid real-world situations. An instance segmentation network~\cite{mohan2021efficientps} is extended to incorporate the pose information so that the network learns an additional $6$-DOF pose for each instance. Therefore, each instance $i$ is represented by the class $c^i$, bounding box $\mbf{B}^i$, mask $\bm{\mathcal{M}}^i$ and the additional pose $\pose{}{o}^i \in \mathbb{SE}[3]$ as illustrated in \reffig{iros_architecture}. The per-instance warping is defined as:
\begin{equation}\label{eq:iros_warping}
        \widehat{\pt}_{s} \sim \mbf{K}  \sum_{i=0}^{m} \big[ \mathcal{M}^i_{\pt_{t}} \mbf{T}_{o}^i  + (1-\bm{\mathcal{M}}^i_{\pt_{t}}) \textit{\textbf{I}}_4 \big] \pose{s}{t} \mbf{D}_{t}  \mbf{K}^{-1} \pt_{t}
\end{equation}
For simplicity, the homogenous and projection operator are omitted. $m$ is the number of dynamic object instances and $\textit{\textbf{I}}_4$ a $4\times 4$ identity matrix. For simplicity, the homogeneous pose and projection transformations are omitted in~\refeq{iros_warping}. The mask $\mathcal{M}^i$ is used to transform only the dynamic object $i$ with its pose $\mbf{T}_{o}^i$. Rigid scene points are transformed only with the pose $\pose{s}{t}$. Using the \refeq{iros_warping}, the image $\widehat{\mbf{I}}_t$ is obtained by inverse warping.

\section{Method}
\label{sec:iros-method}
\subsection{Architecture}
In order to explicitly model the motion of the moving objects, an instance pose head is introduced into an instance segmentation network. EfficientPS~\cite{mohan2021efficientps} has demonstrated SOTA results for panoptic and instance segmentation and is therefore adopted in this method for depth inference.
It consists of the EfficientNet backbone~\cite{tan2019efficientnet}, BiFPN~\cite{tan2020efficientdet}, MaskRCNN instance segmentation head~\cite{he2017mask} and the DPC~\cite{chen2018searching} semantic head.
The EfficientNet backbone has demonstrated its success as a task agnostic feature extractor for nearly all vision tasks. It is easily scalable allowing more complexity/FLOPS trade-off. The BiFPN allows low-level and high-level feature aggregation, thus, enabling a rich representation that accounts for the fine-details and more global abstraction at each feature map. During training, the FPN features ($P_4, P_8, P_{16}, P_{32}$) are extracted for the source and target frames. The two pose heads use both source and target features, while the instance, semantic, and depth heads use only the target features. The model architecture is shown in~\reffig{iros_architecture}. The additional heads are detailed in the following. 

\subsubsection{Instance pose head}  
The key idea of the proposed method is to factorize the motion by explicitly estimating the $6-$DOF pose of each object in addition to the dominant ego-pose. In order to accurately estimate this motion, the objects should be matched and tracked temporally and its interaction should be modeled. Inspired by the prior work on object tracking~\cite{meinhardt2021trackformer,xu2021transcenter}, a novel instance pose head that extends the instance segmentation is proposed using transformer module~\cite{Vaswani}. This head makes use of the multi-head attention to learn the association and interaction of the object across time.

The RPN network yields N proposals. The features of each proposal are pooled using a ROI Align module. These features are extracted for the three frames. Therefore, the input of the instance pose head is $b \times (s+1) \times N \times 256 \times 14\times 14$. Where $b$ is the batch size and $s$ is the number of sources images. 
The first operation is to project these features into the transformer embedding. The linear projection layer flattens the 3 last dimensions and a linear layer is used to learn an embedding of each proposal. This mapping is defined as $\textit{Linear projection}: \mathbb{R}^{B \times (s+1) \times N \times 256 \times 14\times 14}\rightarrow \mathbb{R}^{B \times (s+1)N \times 512}$.

The input of the encoder-decoder transformer is a $(s+1)N$ sequence with $512$ features. The transformer-encoder multi-head attention enables the matching of target frame proposals with respect to the source proposals across time, while the feed-forward learns the matched-motion features. For the transformer-decoder, only the target proposals are used for input. The multi-head attention aggregates the matched-motion features of the encoder to the target proposals and further learns the interactions of the objects by learning an attention between the proposals. 
Finally, a linear layer is used to infer the $6-$DOF pose per object, yielding $B\times N \times s \times6$ using a $6$ axis-angle convention parameters. The non-maximum-suppression used for the object detection head is employed to filter the $N=1000$ object proposals, keeping only the relevant objects. The object pose is inferred only for the filtered objects.
Non-maximum suppression (NMS) is a post-processing technique commonly used in object detection algorithms. It helps eliminate redundant and overlapping bounding box predictions to generate a more concise and accurate set of detections. NMS works by comparing the confidence scores of neighboring bounding boxes and suppressing those that have a significant overlap and lower confidence, keeping only the most confident and non-overlapping detections. This process ensures that only the most relevant and highest-scoring object instances are retained while removing redundant or duplicate predictions.

\subsubsection{Ego-pose branch} The ego-pose branch estimates the dominant pose of the camera. Since the low-level features that allow matching are usually extracted in the first layers, the $P_4$ features of the FPN for source and target features are used. The pose decoder is composed of 4 consecutive convolution with kernels of $k=3$ and the output channels of these 4 convolutions are $256,256,256, 6 \times \text{num\_frames\_to\_predict\_for}$. Since in this experiment the pose is predicted for $t-1$ and $t+1$, $\text{num\_frames\_to\_predict\_for}=2$. Therefore, this network outputs $6$ parameters for the pose transformation using the axis-angle convention.

\subsubsection{Depth branch} The depth branch consists of convolution layers with skip connections from the FPN module as in~\cite{Godard2019}. Similar to prior work~\cite{Zhou2017,Godard2019,Watson2021}, a multiscale depth is estimated in order to resolve the issue of gradient locality. The inference of depth at each scale consists of a convolution with a kernel of $1\times1$ and a Sigmoid activation. The output of this activation, $\sigma$, is re-scaled to obtain the depth $D = \frac{1}{a \sigma +b}$, where $a$ and $b$ are chosen to constrain $D$ between $0.5$ and $100$ units, similar to~\cite{Godard2019}.

To maintain a self-supervised learning setting, a frozen pretrained EffiecientPS that was trained on the Cityscapes benchmark~\cite{cordts2016cityscapes} is used. This pretrained model achieves $PQ=50.2$ and $SQ=76.8$ (see \refsec{eval_metric_seg} for metric definition) on Cityscapes test benchmark. As the representation that was trained for panoptic segmentation may ignore details that are crucial for depth inference. A duplicate of the Backbone and FPN is used for the depth and pose heads. This allows learning features optimized for depth inference without degrading the performances of the panoptic segmentation heads.

The objective function, denoted by $\mathcal{L}$, was previously defined in~\refsec{loss_function_ch3}. It involves minimizing a combination of two losses: the photometric loss ($\mathcal{L}{pe}$) and the depth smoothness loss ($\mathcal{L}{s}$). The final objective function is given by $\mathcal{L} = \mathcal{L}{pe} + \alpha_d \mathcal{L}{s}$, where $\alpha_d$ is a hyperparameter controlling the trade-off between the two losses.

\begin{table*}
\centering
\footnotesize
\resizebox{\textwidth}{!}{
\begin{tabular}{|l |cc||c c c c ||c c c|}
\hline
Method & Supervision &Resolution &Abs Rel& Sq Rel&  RMSE & RMSE log  &
$\delta<1.25$ &
$\delta<1.25^2$ &
$\delta<1.25^3$\\
\hline  
SfMlearner~\cite{Zhou2017} & M & 640$\times$192 & 0.183 & 1.595 & 6.709 & 0.270 & 0.734& 0.902 &0.959 \\
GeoNet~\cite{Yin2018}& M+F &416$\times$128  & 0.155 & 1.296 & 5.857  & 0.233 & 0.793 & 0.931 & 0.973\\
CC~\cite{Ranjan2019} & M+S+F& 832$\times$256& 0.140 & 1.070  & 5.326  & 0.217 &0.826 &0.941 &0.975 \\
Self-Mono-SF\cite{hur2020self} & M+F & 832$\times$256 & 0.125 & 0.978 & 4.877   &0.208 & 0.851 & 0.950 & 0.978 \\
Chen \ea~\cite{chen2019towards} & M+S & 512$\times$256   & 0.118 & 0.905 & 5.096 & 0.211 & 0.839 & 0.945 &0.977\\
Monodepth2~\cite{Godard2019} & M & 640$\times$192 & 0.115 & 0.903 & 4.863 & 0.193 & 0.877 & 0.959 & 0.981\\
Lee \ea~\cite{lee2021attentive} & M+F & 832$\times$256 &  0.113 & 0.835 & 4.693 & 0.191 & 0.879 & 0.961 & 0.981 \\
SGDepth~\cite{Klingner2020} & M+S & 1280$\times$384 & 0.113 & 0.835 & 4.693 & 0.191 & 0.879 & 0.961 & 0.981\\
SAFENet~\cite{lou2020safenet}& M+S & 640$\times$192 & 0.112 & 0.788 & 4.582 & 0.187 & 0.878 & 0.963 & 0.983 \\
Insta-DM~\cite{lee2021learning}& M+S & 640$\times$192 & 0.112& 0.777 &4.772 &0.191& 0.872& 0.959& 0.982 \\
PackNetSfm~\cite{Rares2020}& M &640$\times$192 & 0.111 & 0.785 & 4.601 & 0.189 & 0.878 & 0.960 & 0.982\\
MonoDepthSeg~\cite{safadoust2021self} & M & 640$\times$192 & 0.110 & 0.792 & 4.700 & 0.189 & 0.881 & 0.960 & 0.982 \\
Johnston \ea~\cite{johnston2020self} & M & 640$\times$192 & \underline{0.106} & 0.861 & 4.699 & 0.185 & \underline{0.889} & 0.962 & 0.982 \\
Manydepth~\cite{Watson2021}& M+TS& 640$\times$192  & \textbf{0.098} & \underline{0.770} & \textbf{4.459} & \textbf{0.176} & \textbf{0.900 }& \textbf{0.965} & \underline{0.983} \\
\hline
Ours& M+S & 640$\times$192 &    0.110  &   \textbf{0.719}  &   \underline{4.486}  &   \underline{0.184}  &   0.878  &   \underline{0.964}  &   \textbf{0.984 }  \\
\hline
\end{tabular}
}
\caption{Quantitative performance comparison of on the KITTI benchmark with Eigen split~\cite{Geiger2012CVPR}. For Abs Rel, Sq Rel, RMSE and RMSE log, lower is better, and for $\delta < 1.25$, $\delta < 1.25^2$ and $\delta < 1.25^3$ higher is better. The Supervision column illustrates the training modalities: (M) raw images (S)Semantic, (F) optical flow, (TS) Teacher-student. At test-time, all monocular methods (M) scale the estimated depths with median ground-truth LiDAR.The best scores are bold and the second are underlined}
\label{tab:results}
\end{table*}

\section{Experiments}
\label{sec:iros-results}
\subsection{Setup}
\begin{itemize}
\item{KITTI benchmark~\cite{Geiger2012CVPR}: } 
Following the prior work~\cite{Zhou2017,yang2018unsupervised,Watson2021,Godard2019,Wang2021}, the Eigen~\ea~\cite{Eigen} split is used with Zhou~\ea~\cite{Zhou2017} pre-processing to remove static frames. For evaluation, the common metrics used in the KITTI benchmark will be used ( see in CH\refsec{depth-eval} for more details).

\item{Implementation details: }
PyTorch~\cite{pytorch} is used for all the models. The networks are trained for 40 epochs and 20 for the ablation, with a batch size of 2. The Adam optimizer~\cite{kingma2014adam} is used with a learning rate of $lr=10^{-4}$ and $(\beta_1, \beta_2) =  (0.9,0.999)$. The exponential moving average of the model parameters is used with the $decay=0.995$. As the training proceeds, the learning rate is reduced at epoch 15 to $10^{-5}$. The SSIM weight is set to $\alpha = 0.15$ and the smoothing regularization weight to $\alpha_d = 0.001$. The depth head outputs 4 depth maps. At each scale, the depth is up-scaled to the target image size. The hyperparameters of EfficientPS are defined in ~\cite{mohan2021efficientps} with $N=1000$ before the Non-maximum-suppression. 
Two source images $\img{t-1}$ and $\img{t+1}$ are used. The input images are resized to $192 \times 640$. Two data augmentations were performed: horizontal flips with probability $p=0.5$ and color jitter with $p=1$.
\end{itemize}

\subsection{Results}\label{sec:results}
During the evaluation, the depth is capped to 80m. To resolve the scale ambiguity, the inferred depth map is multiplied by the median scaling. The results are reported in~\reftab{results}. The proposed method achieves competitive performances compared to the state-of-the-art (SOTA) and outperforms~\cite{Watson2021} with respect to the $\textit{Sq Rel}$ with an improvement of $6.62\%$. As expected, the proposed method is superior to the prior works that factorize the motion using the optical flow~\cite{Ranjan2019,Yin2018} as their estimated motion is only local, it does not take into account the class of the object. Besides, it outperforms other similar methods~\cite{lee2021learning} that factorize using the pose for each object.~\reffig{qualitative_2} illustrates the qualitative result comparison. As observed, the proposed method enables high quality depth inference. Compared to the SOTA methods, The method proposed during this thesis is able to represent well the dynamic objects. As the network did not mask the dynamic objects during training, the dynamic objects are better learned compared to the methods that masks the dynamic objects~\cite{Watson2021,Rares2020}.   

\subsubsection{Dynamic and static evaluation}
In contrast to training, where the points are categorized into moving and static-object points, testing is performed on all points that have Lidar ground truth. This does not take into account the relevance of the points and the static/dynamic category. Moving objects are crucial for autonomous driving applications. However, with this testing setup, it is not possible to convey how the model performs on moving objects, especially for methods that masks moving objects during training. This begs the question of whether or not a model trained with a rigid scene assumption learns to represent the depth of dynamic objects even when it is trained with only static objects? 

\begin{figure*}
    \centering
    \includegraphics[width=\textwidth]{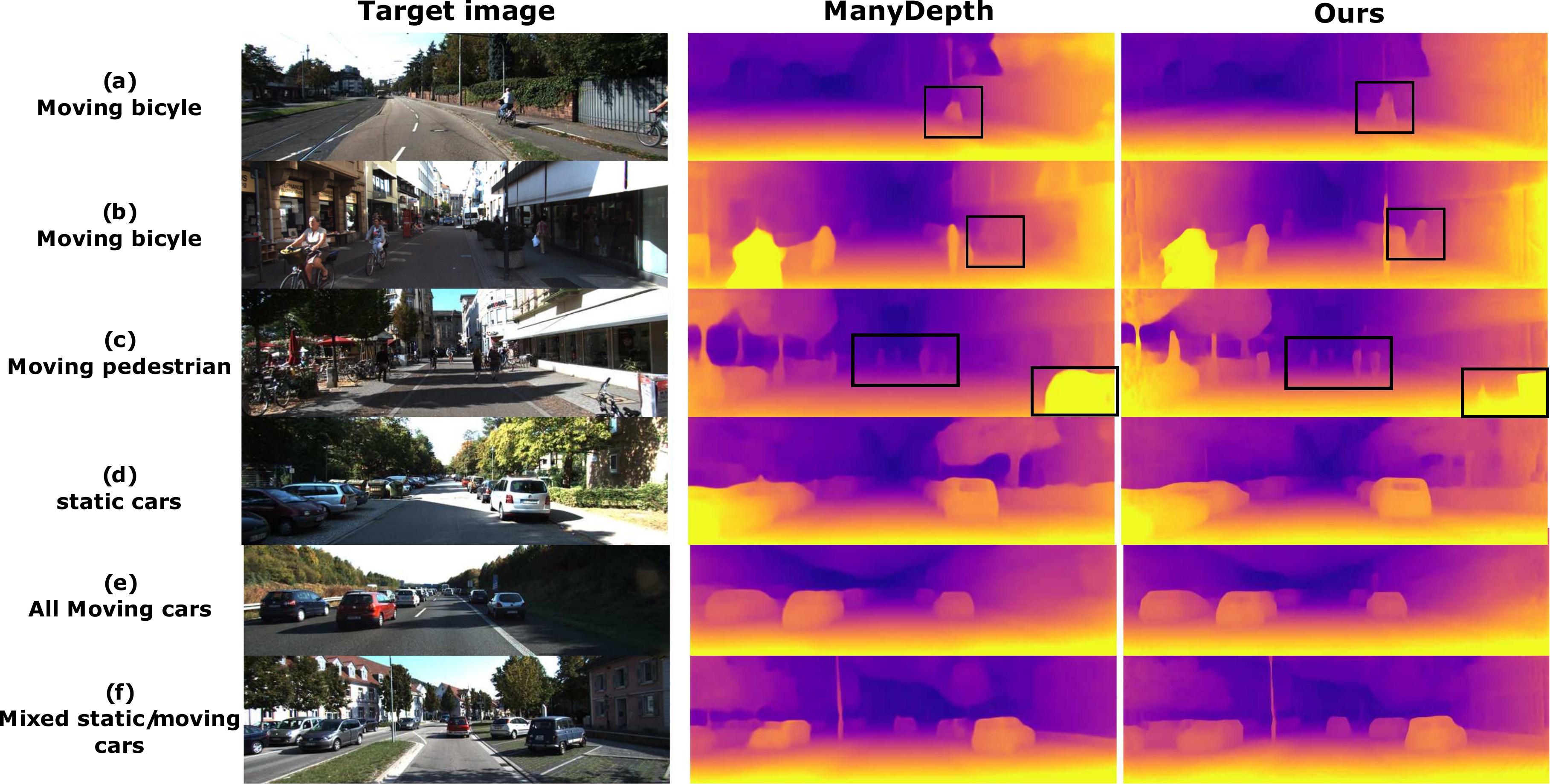}
    \caption{Qualitative results of the proposed method with SOTA methods~\cite{Watson2021}. 
    (a-b-c) show complex situations, as pedestrians and bicycles tend to always move in the KITTI dataset. The qualitative results show that the proposed method outperforms the baselines. 
    (d) The proposed method is on par with the baselines for static objects. (e) and (f) show cars as moving objects. Although the baseline~\cite{Watson2021} is trained with auto-masking, the dataset is rich with static cars that are not masked during training, this provides clues to learn the depth for moving cars. These results are validated further by the quantitative results reported in \reftab{staticVSdynamic}}
    \label{fig:qualitative_2}
\end{figure*}

In order to address this question, the performances of the different methods are evaluated separately with respect to static and dynamic motions. A mask of dynamic objects is used to segment moving objects, and the assessment can be carried out on each category separately. To avoid biasing the evaluation with the EffiecientPS mask, the evaluation mask is obtained using an independent MaskRCNN~\cite{he2017mask} trained with detectron2~\cite{wu2019detectron2}. The first observation that could be made is that the static objects represent $86.43\%$ of test points. This suggests that using the mean across all points will bias the evaluation towards the static objects. A better solution is to consider the per static/dynamic category mean. \reftab{staticVSdynamic} illustrates the evaluation of the method versus the current SOTA method video-to-depth inference~\cite{Watson2021}. The proposed method outperforms the SOTA~\cite{Watson2021} for the dynamic objects with a large difference $\Delta \textit{Sq Rel} = -0.698m$ while the gap for the static objects is only $\Delta \textit{Sq Rel} =+0.011$. The results show that degradation induced by considering the rigid scene assumption is significant. This exposes the limitation of the prior evaluation methods. The KITTI benchmark is biased towards static scenes. In order to unbias the evaluation, the mean per-category is used to balance the influence. The proposed method outperforms the video-to-depth inference method~\cite{Watson2021} with $\Delta \textit{Sq Rel} = -0.344m$. The analysis of \reftab{staticVSdynamic} and \reffig{qualitative_2} suggests that models with rigid scene assumption are still able to infer a depth for moving objects (probably due to the depth smoothness regularization and stationary cars), however, its quality is significantly degraded when compared to the static objects. 

Moreover, the results reported in~\reftab{staticVSdynamic} show that the proposed method outperforms Insta-DM~\cite{lee2021learning} with respect to both the static and dynamic objects. Insta-DM~\cite{lee2021learning} proposes an Obj-PoseNet $\mathcal{O}_\psi: \mathbb{R}^{2 \times H \times W \times 3} \rightarrow \mathbb{R}^{6}$ that takes per-object matched binary instance masks $(\mbf{M}_1, \mbf{M}_2)$ and outputs the object pose. It should be noted, however, that the Insta-DM has an unfair advantage since object matching (via binary masks) is provided as input in a supervised learning approach while the proposed method is self-supervised with matching being implicitly learned in the network. Even so, the proposed method still yields better results on average with respect to dynamic objects. 

\subsubsection{Ablation study}

\begin{table}
\begin{center}
\begin{tabular}{|c|c|c|c|c|c|}
    \hline
    Evaluation& Model & Abs Rel & Sq Rel & RMSE & RMSE log  \\ 
    \hline
    All points mean &
    \multirow{2}{*}{} 
    ManyDepth \cite{Watson2021} & \textbf{0.098} & \underline{0.770} & \textbf{4.459} & \textbf{0.176}   \\
    &
    Insta-DM~\cite{lee2021learning}& 0.112& 0.777 &4.772 &0.191  \\
    &Ours &  \underline{0.110}  &   \textbf{0.719}  &   \underline{4.486}  &   \underline{0.184} \\
    \hline
    \hline 
    Only dynamic &
    \multirow{2}{*}{} 
    ManyDepth \cite{Watson2021} &   0.192  &   2.609  &   7.461  &   0.288   \\
    &
    Insta-DM~\cite{lee2021learning}& \textbf{0.167}  &   \textbf{1.898}  &   \underline{6.975}  &   \underline{0.283} \\
    &
    Ours & \textbf{0.167}  &   \underline{1.911}  &   \textbf{6.724}  &   \textbf{0.271} \\
    \hline
    \hline 
    Only static &
    \multirow{2}{*}{}
    ManyDepth\cite{Watson2021} &    \textbf{0.085}  &  \textbf{ 0.613 } &   \textbf{4.128}  &   \textbf{0.150}    \\ &
    Insta-DM~\cite{lee2021learning} &  0.106  &   0.701  &   4.569  &   0.171   \\
    &
    Ours & \underline{0.101}  &   \underline{0.624}  &   \underline{4.269}  &   \underline{0.163} \\
    \hline
    Per category mean & \multirow{2}{*}{}
    ManyDepth\cite{Watson2021} &   0.139   &  1.611 & 5.794   &  \underline{0.219} \\ &
    Insta-DM~\cite{lee2021learning}  &   \underline{0.137}   &  \underline{1.299} &  \underline{5.772}  &  0.227\\ &
    Ours & \textbf{0.134}  &  \textbf{1,267}  &   \textbf{5,496}& \textbf{0,217}  \\
    \hline
    \end{tabular}

\caption{Quantitative performance comparison for dynamic and static objects. The proposed method outperforms the SOTA~\cite{Watson2021} that uses masking for the dynamic objects with a significant gap $\Delta \textit{Sq Rel} = -0.698m$. In addition, it outperforms Insta-DM~\cite{lee2021learning} which explicitly models dynamic objects.}
\label{tab:staticVSdynamic}

\end{center}
\end{table}

\begin{table*}
\begin{center}
\footnotesize
\resizebox{\textwidth}{!}{
\begin{tabular}{|c||c|c|c|c||c|c|c|c|}
\hline
Ablation & Backbone & Ego-pose input feature & Shared backbone & Piece-wise rigid pose  &Abs Rel & Sq Rel & RMSE & RMSE log  \\ 
\hline
\hline
A1& Resnet18~\cite{he2016deep} & Layer5 & - & - &	 0.121  &   0.914  &   4.890  &   0.196 \\
\hline
\hline
A2 &EfficientNet-b5 & $P_{16}$ & - & - &	  0.132  &   0.906  &   4.981  &   0.205 \\
A3&EfficientNet-b5 & $P_8$ & - & - &	  0.127  &   0.983  &   5.010  &   0.201 \\
A4& EfficientNet-b5 & $P_4$ & - & - &   0.121  &   0.894  &   4.886  &   0.197 \\
A5& EfficientNet-b5 &  $P_4$ &  \checkmark & - &   0.120  &   0.925  &   4.868  &   0.194 \\
\hline
\hline
A6& EfficientNet-b5 &  $P_4$ & \checkmark & \checkmark &   0.113  &   0.795  &   4.689  &   0.190  \\
A7& EfficientNet-b6 & $P_4$ & \checkmark & \checkmark &  0.110  &   0.719  &   4.486  &   0.184  \\
\hline
\end{tabular}}
\caption{An ablation study of the proposed method. The evaluation was done on KITTI benchmark using Eigen split~\cite{Eigen}. As observed, the effect of the backbone is minimal A1 vs A5, the choice of the input feature for ego-pose head is sensible A2 vs A3 vs A4, the performance of the proposed method is obtained mainly by the introduction of the piece-wise rigid pose A5 vs A6. Increasing the complexity of the model allows better performances and better training stability A6 vs A7}
\label{tab:ablations}
\end{center}
\end{table*}

\reftab{ablations} illustrates an ablation study performed to validate the contribution of the proposed method. The results strongly suggest that the performance of the proposed network is mainly obtained by the introduction of the motion factorization through the proposed instance pose head. 

\begin{itemize}
    \item \textbf{A1 versus A4:} Introducing a more complex architecture did not contribute to the improvement of the performances.
    \item \textbf{A4 versus A5:} Sharing the backbone for the depth network did not contribute to the improvement of the performances. However, it did reduce the complexity of the network.
    \item \textbf{A5 versus A6: } Introducing the piece-wise rigid pose warping induces an improvement of $\Delta \textit{Sq rel}=14.1\%$
    \item \textbf{A2 versus A3 versus A4: } The pose head is sensitive to the choice of the features level. $P_4$ is the optimal level for this application.
\end{itemize}

These results suggest that not only the models learn an accurate depth, but also accurate instance pose. This result demonstrates that the transformer network is able to match and learn the interaction of the objects across time. The model in $A5$ is on the same setting of the other SOTA methods~\cite{Watson2021,Rares2020}. Despite using this low performance baseline, the introduction of the dynamic warping enabled the proposed method to achieve the SOTA results. 

An interesting observation during training is that $A6$ under-fits the data (i.e., the validation loss is less than the learning loss). The test performances are not stable, the best model among the 20 epochs is reported for this backbone. In order to resolve this under-fitting, the complexity of the model is increased $A7$. This allows for a better stability of the training loss and test performance. The best results are obtained using this complexity.
The additional instance pose results in an additional run-time overhead during training. The training time for 1 epoch for A5 and A7 is $233mn$ and $58mn$ trained on RTX3090 respectively. However, the additional run-time is only for the training. At test-time, the depth network requires only a single pass of the image $\img{t}$ with roughly 34FPS for $A7$ model and 38FPS for $A6$ model using a single RTX3090.

\section{Disucssion}
\label{sec:iros-conclusion}
In this chapter, a novel instance poses head is introduced for self-supervising monocular depth inference. This head enables the factorization of the scene's motion. Thus, alleviating the rigid scene assumption. It is shown that it achieves the SOTA results on the KITTI benchmark~\cite{Geiger2012CVPR}. The ablation study further validates that the multi-head attention of the transformer network infer an accurate object pose. Moreover, the impact of the dynamic motion on this benchmark is exposed. Namely, the bias towards static objects, where $86.43\%$ of the test pixels correspond to static objects. A mean per static/dynamic category metric is proposed to unbias the assessment. 

One fundamental limitation of these single-image-to-depth methods is that these models rely on the prior knowledge such as object shape, textures, camera position with respect to the floor in order to recover the depth. Recovering the geometry with triangulation or matching is not possible, as the network uses single images only. However, even this capability of recovering 3D from 2D with good accuracy is already an impressive result.
Another limitation of the current method is the depends on the performance of the instance segmentation network. While the panoptic segmentation network works well on KITTI, the performance of this model is not guaranteed when scaling the training for other datasets. This might limit the possibility to apply this method on huge datasets where the self-supervision is more pertinent. 


\chapter*{}

\chapter{Video-to-depth forecasting}
\label{ch:ch5}
Now that self-supervised monocular depth inference has been presented, this next chapter will look at future depth forecasting. As discussed earlier, in this chapter, the term “forecasting” will be used to describe the methods that output the \textbf{future} depth of a sequence of images. Given a sequence of raw images, the aim is to forecast the 3D information using a self supervised photometric loss. The architecture is designed using both convolution and transformer modules. This leverages the benefits of both modules: the Inductive bias of CNN, and the multi-head attention of transformers, thus enabling a rich spatio-temporal representation that enables accurate depth forecasting. The approach performs significantly well on the KITTI dataset benchmark, with several performance criteria being even comparable to prior non-forecasting self-supervised monocular depth inference methods.

In the Section \refsec{icpr-intro}, we motivate our method. We discuss related work in \refsec{icpr-related}. We present
our approach in \refsec{icpr-method} and experimental results in \refsec{icpr-exp}. We conclude in \refsec{icpr-conclusion}.

This chapter is based on the following publication: 
\begin{itemize}
    \item \textbf{Conference paper:} Boulahbal Houssem Eddine, Adrian Voicila, and Andrew I. Comport. "Forecasting of depth and ego-motion with transformers and self-supervision." 2022 26th International Conference on Pattern Recognition (ICPR). IEEE, 2022. 
\end{itemize}

\section{Introduction}
\label{sec:icpr-intro}
Forecasting the future is crucial for intelligent decision-making. It is a remarkable ability of human beings to effortlessly forecast what will happen next, based on the current context and prior knowledge of the scene. Forecasting sequences in real-world settings, particularly from raw sensor measurements, is a complex problem due to the exponential time-space space dimensionality, the probabilistic nature of the future and the complex dynamics of the scene. Whilst much effort from the research community has been devoted to video forecasting~\cite{Mathieu2016,Finn,WilliamLotter2016,Rakhimov} and semantic forecasting~\cite{Terwilliger2019,Bhattacharyya2019,Graber2021,Saric2020}, depth and ego-motion forecasting have not received the same interest despite their importance. The geometry of the scene is essential for applications such as planning the trajectory of an agent. 
Anticipating is therefore important for autonomous driving autopilots or human/robot interaction, as it is critical for the agent to quickly respond to changes in the external environment. 

The first work that explored depth forecasting was carried out by Mahjourian \ea~\cite{Mahjourian2017}, the aim of that paper was to use the forecasted depth to render the next RGB image frame. They supervised the depth loss using ground-truth LiDAR scans and the warping was done using ground-truth poses.~\cite{Qi2019} used additional modalities for input, namely, a multi-modal RGB, depth, semantic and optical flow and forecasted the same future modalities. The supervision was carried out using the aforementioned ground-truth labels.~\cite{Hu2020} developed a probabilistic approach for forecasting using only input images and generated a diverse and plausible multi-modal future including depth, semantics and optical flow. However, it was supervised through ground-truth labels and the final loss was a weighted sum of future segmentation, depth and optical flow losses similar to~\cite{Qi2019}. While these methods enable forecasting the depth, they suffer from two shortcomings:~\cite{Hu2020,Qi2019,Mahjourian2017} require the ground-truth labels for supervision during training and testing and~\cite{Qi2019} uses a multi-modal input for inference that requires either ground-truth labels or a separate network. 

The work presented in this chapter addresses the problem of depth and ego-motion forecasting using only monocular images sequence with self-supervision. Monocular depth and ego-motion inference has been successful for self-supervised training~\cite{Wang2021,Almalioglu2019,Jaderberg2015,Chen2016,Godard2017,Godard2019,Chen2019,Ranjan2019,Rares2020,Yin2018}. The basic idea is to jointly learn depth and ego-motion supervised by a photometric reconstruction loss. In this chapter, it is demonstrated that it is possible to extend this self-supervised training to sequence forecasting. An accurate forecasting requires a knowledge of the ego-motion, semantics, and the motion of dynamic objects. Powered by the advances of transformers~\cite{Vaswani,Devlin2019,Brown2020,dosovitskiy2020,Liu2021}, and using only sensor input, the network learns a rich spatio-temporal representation that encodes the semantics, the ego-motion and the dynamic objects. Therefore, avoiding the need for extra labels for training and testing. The results on the KITTI benchmark~\cite{Geiger2012CVPR} show that the proposed method is able to forecast the depth accurately and outperform even non-forecasting methods~\cite{Eigen,Liu2016,Zhou2017,yang2018unsupervised}.

\section{Related work}
\label{sec:icpr-related}

\subsection{Sequence forecasting}
Anticipation of the future state of a sequence is a fundamental part of the intelligent decision-making process. The forecasted sequence could be an RGB video sequence~\cite{Finn,WilliamLotter2016,Babaeizadeh2018,Mathieu2016,Kumar2019,Rakhimov}, depth image sequence~\cite{weng2020inverting,Mahjourian2017}, semantic segmentation sequence~\cite{Terwilliger2019,Bhattacharyya2019,Luc2017,Graber2021,Chiu2020,Hu2020,Graber2021,Saric2020} or even a multi-modal sequence~\cite{Hu2020,Qi2019}. Early deep learning models for RGB video future forecasting leveraged several techniques including: Recurrent models~\cite{WilliamLotter2016}, variational autoencoder VAE~\cite{Babaeizadeh2018}, generative adversarial networks~\cite{Mathieu2016}, autoregressive model~\cite{Rakhimov} and normalizing flows\cite{Kumar2019}. These techniques have inspired subsequent sequence forecasting methods.
Despite the importance of using geometry for developing better decision-making, depth forecasting is still in early development.~\cite{Mahjourian2017} used supervised forecasted depth along with supervised future pose to warp the current image and generate the future image. Instead of using images as input,~\cite{weng2020inverting} used LiDAR scans and forecast a sparse depth up to 3.0s in the future on the KITTI benchmark~\cite{Geiger2012CVPR}.~\cite{Qi2019} used a multi-modal input/output and forecast the depth among other modalities.~\cite{Hu2020} handled the diverse future generation by utilizing a variational model to forecast a multi-modal output. The use of multi-modalities requires additional labels or pretrained networks. This makes the training more complicated. 
Instead, the work presented in this chapter leverages only raw images and forecasts in a self-supervised manner.

\subsection{Vision transformers}
The introduction of the Transformers in 2017~\cite{Vaswani} revolutionized natural language processing, resulting in remarkable results~\cite{Devlin2019,Brown2020,radford2021learning}. The year 2020~\cite{dosovitskiy2020,carion2020end} marked one of the earliest pure vision transformer networks. As opposed to recurrent networks that process sequence elements recursively and can only attend to short-term context, transformers can attend to complete sequences, thereby learning long and short relationships. The multi-head attention could be considered as a fully connected graph of the sequence's features. It demonstrated its success by outperforming convolution based networks on several benchmarks including classification~\cite{dosovitskiy2020,zhai2021scaling,li2021improved}, detection~\cite{li2021improved,carion2020end,li2021grounded} and segmentation~\cite{Liu2021,cheng2021masked}. This has led to a paradigm shift~\cite{liu2021we}, transformers are slowly winning "The Hardware Lottery"~\cite{hooker2021hardware}. However, training vision transformers is complicated as these modules are not memory efficient for images and need large dataset pretraining.~\cite{carion2020end} has demonstrated that it is possible to combine convolution and transformers to learn a good representation without requiring large pertaining.
The proposed method proposes to leverage a hybrid CNN and transformer network as in~\cite{carion2020end} that is designed to forecast the geometry of the scene. The proposed network is simple and yet efficient. It outperforms even prior monocular depth inference methods~\cite{Eigen,Liu2016,Zhou2017,yang2018unsupervised} that have access to the ground truth.

\section{The method}
\begin{figure*}
    \centering    
\includegraphics[width=\textwidth]{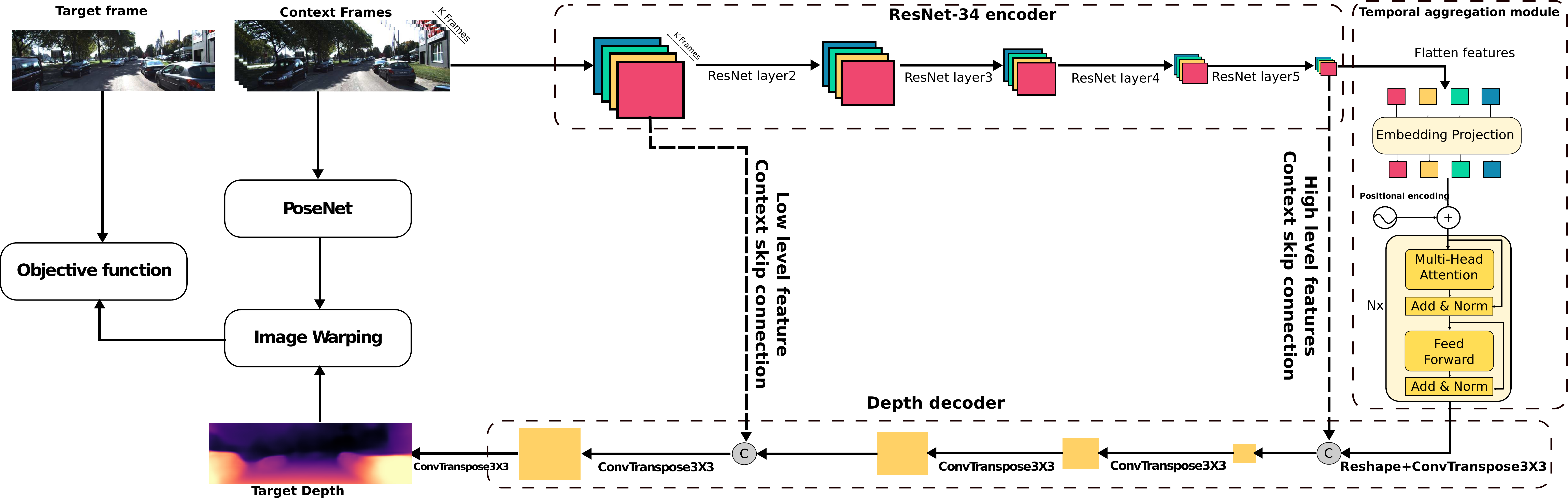}
    \caption{Illustration of the proposed architecture. Two sub-networks are used for training: The PoseNetwork as in network~\cite{Kendall2015,Godard2019} is used to forecast the ego-motion. The depth network combines both CNN and transformers. The Resnet34~\cite{he2016deep} encoder extracts the spatial features for each context frame. The embedding projection module projects these features into $R^{k \times d_{model}}$ where $k=4$ is the context frames. $N=3$ transformer encoders are used to fuse the spatial temporal to obtain a rich spatio-temporal features. The output of the transformer module encodes the motion of the scene. The decoder uses simple transposed convolution. In order to recover the context, skip connections are pooled from the encoder. Only the last frame features are pooled for the context. The decoder outputs a disparity map that will be used along with the pose network to warp the source images onto the target.}
    \label{fig:icpr_architecture}
\end{figure*}
\label{sec:icpr-method}
\subsection{The problem formulation}
\label{sec:formulation}
Let $\mbf{I}_t \in \mathbb{R}^{w \times h \times c}$ be the t-th frame in a video sequence $\mbf{I} = \{ \mbf{I}_{t-k:t+n}\}$. The frames $\mbf{I}_c = \{\mbf{I_{t-k:t}}\}$ are the context of $\mbf{I}_{t}$ and $\mbf{I}_f =\{\mbf{I}_{t+1:t+n} \}$ is the future of $\mbf{I}_{t}$. 
The goal of the future depth and ego-motion forecasting is to predict the future depth image of the scene $\mbf{D}_{t+n}$ and the ego-motion $\pose{t+n}{t}$ corresponding to $\mbf{I}_{t+n}$ given only the context frames  $\mbf{I}_{c}$:
\begin{equation}\small
    \centering
    (\mbf{\widehat{D}}_{t+n}, ^{t+n}\mbf{\widehat{T}}_{t}) = f(\mbf{I}_c;\bm{\theta})
    \label{eq:problem_formulation}
\end{equation}
where $f$ is a neural network with parameters $\bm{\theta}$.

In self-supervised learning depth inference, the problem is formulated as novel view synthesis by warping the source frame $\mbf{I}_{s}$ into the target frames $\mbf{I}_{tar}$ using the depth and the $\pose{s}{tar} \in \mathbb{SE}[3]$ pose target to source pose. The warping is defined as defined in~\refeq{warping_classic}: 
\begin{equation}
        \widehat{\pt}_{s} \sim \pi(\mbf{K} \pose{s}{t} H( \mbf{D}_{t}  \mbf{K}^{-1} \pt_{t}))
\end{equation}
  
Reconstructing the frame $\mbf{I}_{t+n}$ using the depth and the pose from only the context by a warping could be formulated as a maximum likelihood problem: 
 \begin{equation}
 \small
     \bm{\widehat{\theta}} = \argmax_{\bm\theta \in \bm{\Theta}} L(\mbf{I}_{t+n}| \mbf{I}_{c}; \bm{\theta} ) \equiv	 \argmax_{\bm\theta \in \bm{\Theta}} \sum_{m} P_{model}(\mbf{I}_{t+n}^m| \mbf{I}_{c}^m)
     \label{eq:likelihood}
 \end{equation} 
where $m$ is the number of samples. If $P_{model}$ is assumed to follow a Laplacian distribution $P_{model}(\mbf{I}_{t+n}| \mbf{I}_{c}) \sim Lap(\mbf{I}_{t+n};\bm{\mu}=\widehat{\mbf{I}}_{t+n};\bm{\beta}=\sigma^2 \mbf{I})$. $\widehat{\mbf{I}}_{t+n}$ is the warped image. Then, maximizing the~\refeq{likelihood} is equivalent to minimizing an $L_1$ error of $\mbf{\widehat{I}}_{t+n}$ and the known image frame $\mbf{I}_{t+n}$. Similarly, if the distribution is assumed to follow a Gaussian distribution, the maximization is equivalent to minimizing an $L_2$ error.

\subsection{The architecture}
The architecture of the network is depicted in \reffig{icpr_architecture}. The forecasting network is composed of two subnetworks: a pose net to forecast the future $\pose{s}{t}$, that transform the target to the source frame, and a depth network that forecast $\mbf{D}_{t+n}$ (see~\refeq{problem_formulation}).
Similar to \cite{Kendall2015}, the pose-net is composed of a classification network~\cite{he2016deep} as a feature extractor followed by a simple pose decoder as in~\cite{Godard2019}. The pose network forecasts 6 parameters using the axis-angle representation.
The depth network leverages a hybrid CNN and Transformer network as in~\cite{carion2020end} that is designed to forecast the geometry of the scene. This network benefits from both modules. The convolution module is used to extract the spatial features of the frames as it is memory efficient, easy to train and does not require large pretraining. The transformer module is used for better temporal feature aggregation. The multi-head attention could be considered as a fully-connected graph of the features of each frame. Therefore, the information is correlated across all the frames rather than incrementally, one step at a time, as in LSTM~\cite{hochreiter1997long}. The architecture consists of three modules: an encoder, temporal aggregation module and a decoder.  

\subsubsection{Encoder: } 
ResNet~\cite{he2016deep} is one of the most used foundation models~\cite{bommasani2021opportunities}. It has demonstrated its success as a task agnostic feature extractor for nearly all vision tasks. In this work, ResNet34 is used as feature extractor. It is pretrained on ImageNet~\cite{deng2009imagenet} for better convergence. Each context frame is fed-forward and a pyramid of features is extracted. These features encode the spatial relationship between each scene separately. Thus, at the output of this module, a pyramid of spatial features for each frame is constructed. These features will be correlated temporally using the Temporal aggregation module TAM.

\subsubsection{Temporal aggregation module}
Since its introduction, transformers have demonstrated their performance, outperforming their LSTM/RNN counterparts in various sequence learning benchmarks~\cite{Walker2021,raffel2019exploring,Brown2020,Devlin2019}. 
forecasting accurate depth requires knowledge of the static objects, accurate ego-motion and knowledge of the motion of the dynamic objects. The last layer of the encoder is assumed to encode higher abstraction features (\eg recognizing objects). Therefore, correlating temporally these features allows the extraction of the motion features of the scene. The TAM consists of two submodules: 
\begin{itemize}
    \item \textbf{Embedding projection: } The dimensions after flattening the feature output of the last layer of the encoder is not memory efficient for the transformers. The embedding projection maps these features as:\footnote{The batch is omitted} $\mathbb{R}^{K \times C \times H \times W} \longrightarrow \mathbb{R}^{K \times d_{enc}}$. 
    \item \textbf{Transformer encoder: } After projecting the features using the embedding layer, a Transformer encoder with $N$ layer, $m$ multi-head attention and $d_{enc}$ is used. It correlates the spatial features of the sequence, producing fused spatio-temporal features.
\end{itemize}
\subsubsection{Depth decoder} After the spatio-temporal fusion, the decoder takes these spatio-temporal features along with the context features as input and decodes them to produce a disparity map. As depicted in~\reffig{icpr_architecture}, the context of the scene is obtained by pooling the features of last frame in the encoder. Two levels are pooled and concatenated. $^{dec}f_{t+n} = [^{enc}f_{t}  \: , \:  TAM(^{enc}f_{t-4:t})]$. The high level features (skip connection before the TAM) enable learning the motion, while the low level features (skip connection at the start of ResNet) recover the finer details lost by the down-sampling. Therefore, the decoder maps (context + motion $\longrightarrow$ depth). 

Each level of the decoder consists of a simple sequential layer of: transposed convolution with a kernel of $3\times3$ with similar channels to the encoder, batch normalization and Relu activation in that order. The forecasting head consists of a convolution with a kernel of $1\times1$ and a Sigmoid activation. The output of this activation, $\sigma$, is re-scaled to obtain the depth $D = \frac{1}{a \sigma +b}$, where $a$ and $b$ are chosen to constrain $D$ between $0.1$ and $100$ units, similar to~\cite{Godard2019}. For training, each level has a forecasting head, but only the last head is used for inference. 

\subsection{Objective functions}
\label{sec:objetives}
As formulated in \refsec{formulation}, learning the parameters $\widehat{\bm{\theta}}$ involves maximizing the maximum likelihood of $P_{model}$. As presented in~\refsec{loss_function_ch3}, the loss functions that will be used to optimize the parameters of the network are:
\begin{itemize}
    \item \textbf{Photometric loss: } Following~\cite{Zhou2017,Godard2019,Rares2020} The photometric loss seeks to reconstruct the target image by warping the source images using the forecast pose and depth. An $L_1$ loss is defined as follows:
    \begin{equation}\small
        \mathcal{L}_{rec}(\img{t+n}, \widehat{\mbf{I}}_{t+n}) = \sum_{\pt} | \img{t+n}(\pt) - \widehat{\mbf{I}}_{t+n}(\pt) |
    \end{equation}
    where $\widehat{\mbf{I}}_{t+n}(\pt)$ is the reverse warped target image obtained by~\refeq{warping_classic}. This simple $L_1$ is regularized using SSIM~\cite{wang2004image} that has a similar objective to reconstruct the image. The final photometric loss is defined as:
    \begin{equation}\small
    \begin{aligned}
      \mathcal{L}_\textup{pe}(\img{t+n}, \widehat{\mbf{I}}_{t+n}) = \sum_{\pt} \big[&(1 - \alpha) \textup{ SSIM}[ \img{t+n}(\pt) - \widehat{\mbf{I}}_{t+n}(\pt)]  \\
      & +  \alpha | \img{t+n}(\pt) - \widehat{\mbf{I}}_{t+n}(\pt) |\big]
    \end{aligned}
    \end{equation}
    \item \textbf{Depth smoothness: } An edge-aware gradient smoothness constraint is used to regularize the photometric loss. The disparity map is constrained to be locally smooth through the use an image-edge weighted $L_1$ penalty, as discontinuities often occur at image gradients. This regularization is defined as~\cite{heise2013pm}:
    \begin{equation}\small
    \begin{aligned}
      \mathcal{L}_{s}(D_{t+n}) = \sum_{p} \big[ &| \partial_x D_{t+n}(\pt) | e^{-|\partial_x \mbf{I}_{t+n}(\pt)|}    +\\
      &| \partial_y D_{t+n}(\pt) | e^{-|\partial_y \mbf{I}_{t+n}(\pt)|} \big] 
    \end{aligned}
\end{equation}
\end{itemize}

Training with these loss functions is subject to major challenges: gradient locality, occlusion and out of view-objects. Gradient locality is a result of bilinear interpolation\cite{jaderberg2015spatial,Zhou2017}. The supervision is derived from the fours neighbors of $I(\pt_s)$ which could degrade training if that region is low-textured. Following~\cite{Godard2019,Godard2017,garg2016unsupervised}, an explicit multiscale approach is used to allow the gradient to be derived from larger spatial regions. A forecasting head is used at each level to obtain each level's disparity map during training. 
\refeq{warping_classic} assumes global ego-motion to calculate the disparity. Supervising directly using this objective is inaccurate when this assumption is violated (\eg the camera is static or a dynamic object moves with the same velocity as the camera). According to~\cite{Godard2019} this problem can manifest itself as ‘holes’ of infinite depth. This could be mitigated by masking the pixels that do not change the appearance from one frame to the next.
A commonly used solution~\cite{Zhou2017,Godard2019} is to learn a mask $\mu$ that weighs the contribution of each pixel, while~\cite{Zhou2017} uses an additional branch to learn this mask. This approach uses the auto-masking defined in~\cite{Godard2019} to learn a binary mask $\mu$ as follows:
\begin{align}
\mu(\mbf{I}_{t+n}, \widehat{\mbf{I}}_{t+n}, \mbf{I}_{t}) =\mathcal{L}_{pe}(\img{t+n}, \widehat{\mbf{I}}_{t+n}) < \mathcal{L}_{pe}(\img{t+n},{\mbf{I}}_{t})
\end{align}
$\mu$ is set to only include the loss when the photometric loss of the warped image $\widehat{\mbf{I}}_{t+n}$ is lower than the original unwarped image $\img{t}$.
The final objective function is defined as:
\begin{equation}\small
    \mathcal{L} = \sum_{l} \:[ \:\mu \: \mathcal{L}_p + \alpha_d{L}_s \:]
\end{equation}
where $l$ is the scale level of the forecast depth.

\section{Experiments}
\label{sec:icpr-exp}

\subsection{Setting}
\subsubsection{KITTI benchmark~\cite{Geiger2012CVPR}: } 
Following the prior work~\cite{Eigen,Liu2016,Zhou2017,yang2018unsupervised,Mahjourian2018,Godard2019,Wang2021}, the Eigen~\ea~\cite{Eigen} split is used with Zhou~\ea~\cite{Zhou2017}. Frames without sufficient context (starting images in video) are excluded from the training and testing. This split has become the defacto benchmark for training and evaluating depth that is used by nearly all depth methods.
\subsubsection{Baselines: }
\label{sec:baselines}  
As discussed above, previous work on depth forecasting has been supervised using LiDAR scans, and has used a multimodal network that provides depth. Their evaluation is neither performed on the Eigen split, nor does it use the defacto self-supervised metrics.
In order to fairly evaluate the proposed method, a self-supervised monocular formulation will be used to compare performance with the KITTI Eigen split benchmark. Comparisons will be made with three approaches: prior work on self-supervised depth inference~\cite{Eigen,Liu2016,Zhou2017,yang2018unsupervised,Godard2019,Wang2021}; copy of the last observed LiDAR frame as done in~\cite{Qi2019}; and ForecastMonodepth2, a modified version of~\cite{Godard2019} that is adapted for forecasting pose/depth.

\subsubsection{Evaluation metrics: }
For evaluation, the metrics of previous works~\cite{Eigen} are used for the depth (see \refsec{depth-eval}). To resolve the scale ambiguity, the forecast depth map is scaled by median scaling where $s = \frac{median(D_{gt})}{median(D_{pred})}$. During the evaluation, the depth is capped to 80m. For the pose evaluation, the Absolute Trajectory Error (ATE) defined in~\cite{sturm2012benchmark} is used to evaluate on the KITTI odometery benchmark~\cite{Geiger2012CVPR} for sequences 09 and 10. 
\subsubsection{Implementation details: }
PyTorch~\cite{pytorch} is used for all models. The networks are trained for 20 epochs, with a batch size of 8. The Adam optimizer~\cite{kingma2014adam} is used with a learning rate of $lr=10^{-4}$ and $(\beta_1, \beta_2) =  (0.9,0.999)$. As training proceeds, the learning rate is decayed at epoch 15 to $10^{-5}$. The SSIM weight is set to $\alpha = 0.15$ and the smoothing regularization weight to $\alpha_d = 0.001$. $l=4$ scales are used for each output of the decoder. At each scale, the depth is up-scaled to the target image size. $d_{model}=2048$, $m=16$ and $N=3$ for the TAM projection.
The input images are resized to $192 \times 640$. Two data augmentations were performed: horizontal flips with probability $p=0.5$ and color jitter with $p=1$.
$k=4$ frames are used for context sequence and $n=5$ is used for short term forecasting and $n=10$ for midterm forecasting as in~\cite{Qi2019} which corresponds to forecasting $0.5$s and $1.0$s into the future. 
The ForecastedMonodepth2 is the same as~\cite{Godard2019} with a modified input. The context images are concatenated and used as input for both depth and pose networks.

\subsection{Depth forecasting results}
\begin{table*}[t!]
\renewcommand{\arraystretch}{0.90}
\centering
{
\small
\setlength{\tabcolsep}{0.3em}
\resizebox{\textwidth}{!}{

\begin{tabular}{c|ccc|cccc|ccc}
\hline
\textbf{Method} &
Forecasting &
Resolution & 
Supervision &
Abs Rel &
Sq Rel &
RMSE log &
RMSE &
$\delta<1.25$ &
$\delta<1.25^2$ &
$\delta<1.25^3$\\
\hline    
Eigen \ea~\cite{Eigen} &-& 576 x 271 & D  & 0.203 & 1.548 & 0.282 & 6.307 & 0.702 & 0.898 & 0.967 \\
Liu \ea~\cite{Liu2016}&-& 640 x 192 & D  & 0.201 & 1.584  & 0.273 & 6.471 & 0.680 & 0.898  & 0.967 \\
SfMLearner~\cite{Zhou2017} & -  & 416 x 128 & SS & 0.198 & 1.836& 0.275  & 6.565 & 0.718 & 0.901 & 0.960\\
Yang \ea~\cite{yang2018unsupervised} & - & 416 x 128 & SS & 0.182 & 1.481  & 0.267& 6.501 & 0.725 & 0.906 & 0.963\\
Vid2Depth~\cite{Mahjourian2018} & - & 416 x 128 & SS & 0.159 & 1.231  & 0.243 & 5.912& 0.784 & 0.923 & 0.970 \\
Monodepth2~\cite{Godard2019} & - & 640 x 192  & SS& 0.115 & 0.882 & 0.190 & 4.701 & 0.879 & 0.961 & 0.982\\
Wang \ea~\cite{Wang2021}  & - & 640 x 192  & SS & 0.109 & 0.779 & 0.186& 4.641 & 0.883 & 0.962 & 0.982\\
\hline
LiDAR Train set mean & -  & 1240 x 374& - & 0.361 & 4.826 & 0.377& 8.102  & 0.638 & 0.804 & 0.894\\
ForecastMonodepth2 & 0.5sec & 640 x 192 & SS & \underline{0.201}  &   \textbf{1.588}   &   \underline{0.275}  &   \textbf{6.166 }&   \underline{0.702}  &   \underline{0.897}  &   \underline{0.960}\\
\textbf{Ours}& 0.5sec & 640 x 192 & SS & \textbf{0.178}  &   \underline{1.645}   &   \textbf{0.257} &   \underline{6.196}  &   \textbf{0.761}  &   \textbf{0.914}  &   \textbf{0.964} \\
\hline
Copy last LiDAR scan & 1sec & 1240 x 374 & - &0.698  &  10.502  &  15.901  &   7.626  &   0.294  &   0.323  &   0.335\\
ForecastMonodepth2 & 1sec & 640 x 192&  SS& \underline{0.231}  &   \textbf{1.696}   &   \underline{0.303} &   \underline{6.685}  &   \underline{0.617}  &   \underline{0.869}  &   \underline{0.954} \\
\textbf{Ours}& 1sec &640 x 192 & SS & \textbf{0.208}  &   \underline{1.894}   &   \textbf{0.291 }&   \textbf{6.617}  &   \textbf{0.701}  &   \textbf{0.882}  &   \textbf{0.949}\\
\hline
\end{tabular}
}}

\caption{Quantitative performance comparison of  on the KITTI benchmark with Eigen split~\cite{Geiger2012CVPR} for distances up to 80m.
In the \emph{Supervision} column,  D refers to depth supervision using LiDAR groundtruth and (SS) self-supervision. At test-time, all monocular methods (M) scale the depths with median ground-truth LiDAR.}
\label{tab:depth-accuracy}
\end{table*}

\begin{table}
    \centering 
    \begin{tabular}{c|c|c|c}
         Method & forecasting & Seq.09 & Seq.10 \\
         \hline
         Mean Odom & - &0.032 $\pm$ 0.026 & 0.028 $\pm$ 0.023\\
         ORB-SLAM~\cite{mur2015orb}  & - & 0.014 $\pm$ 0.008 & 0.012 $\pm$ 0.011\\ 
         SfMLearner~\cite{Zhou2017} & - & 0.021 $\pm$ 0.017 & 0.020 $\pm$ 0.015 \\
         Monodepth2~\cite{Godard2019} & - & 0.017 $\pm$ 0.008 & 0.015 $\pm$ 0.010 \\
         Wang \ea~\cite{Wang2021} &  - & 0.014 $\pm$ 0.008 & 0.014 $\pm$ 0.010 \\
         \hline
         Ours & 0.5s & 0.020 $\pm$ 0.011 & 0.018 $\pm$ 0.011 \\
         \hline
    \end{tabular}
    \caption{ATE error of the proposed method and the prior non-forecasting methods on KITTI~\cite{Geiger2012CVPR}. The proposed method is comparable to these methods even if it only accesses past frames.}
    \label{tab:pose}
\end{table}

\reftab{depth-accuracy} shows the results of the proposed method on the KITTI benchmark~\cite{Geiger2012CVPR}. As specified in \refsec{baselines}, the method is compared to three approaches: prior work on depth inference; copying last frame; and adapting monodepth2~\cite{Godard2019} for future forecasting. The proposed method outperforms the forecasting baselines for both short and midterm forecasting, especially for short range forecating. The results are even comparable to non-forecasting methods~\cite{Eigen,Liu2016,Zhou2017,yang2018unsupervised} that have access to $\mbf{I}_{t+n}$. The gap between state-of-the-art depth inference and the proposed forecasting method is reasonable due to the uncertainty of the future, the unobservability of certain events such as a new object entering the scene and the complexity of natural videos that requires modeling correlations across space-time with much higher input dimensions.

\begin{figure*}
    \centering
    \includegraphics[width=\textwidth]{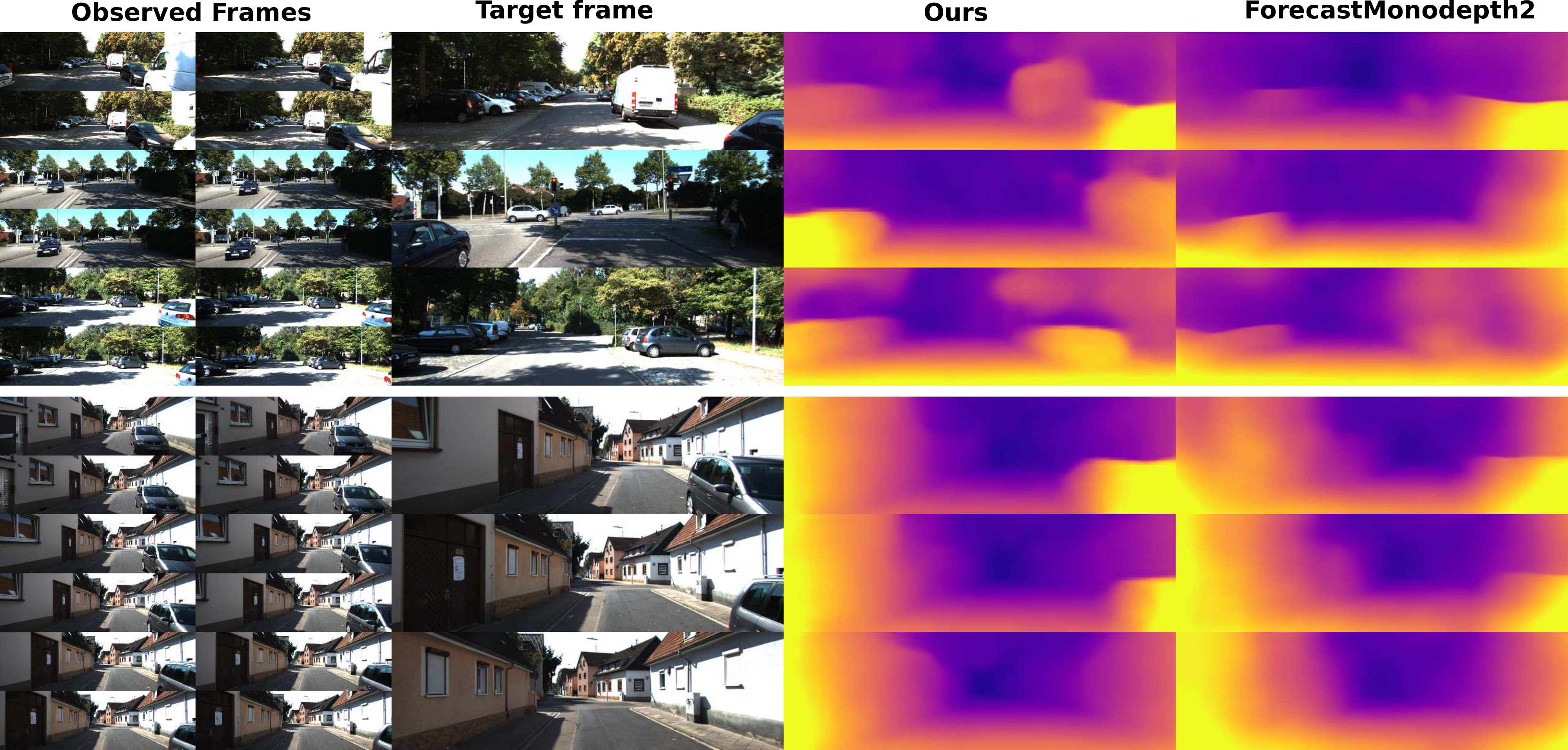}
    \caption{Qualitative results of the comparison of the proposed method with the ForecastMonodepth2 baseline. This comparison shows that the proposed method performs better than the baseline, especially for nearby dynamic objects. This observation is further validated in Table IV. In addition, the baseline method is showing a lack of detection of moving objects, which leads to a degradation of the forecasted depth. The proposed method is able to detect moving objects, thus accurately forecasting the depth of the scene.}
    \label{fig:icpr_qualitative}
\end{figure*}

\reffig{icpr_qualitative} shows an example of depth forecasting on the Eigen test split. Several observations can be made: 
\begin{itemize}
    \item The network handles correctly the out-of-view object.
    \item The network learned the correct ego-motion: The position of the static objects is accurate.
\end{itemize}
These results suggest that the network is able to learn a rich spatio-temporal representation that enables learning the motion, geometry, and the semantics of the scene. Thus, the proposed method extends the self-supervision depth inference to perform future forecasting with comparable results. A further analysis is done to evaluate and validate the choices of the network in~\refsec{ablation}.

\subsubsection{Results with respect to distance}

\begin{table*}
\centering
\resizebox{\textwidth}{!}{
\begin{tabular}{|c|cc|ccccccc|}
    \hline
    \textbf{Range} &
    Method &
    Forecasting &
    Abs Rel &
    Sq Rel &
    RMSE log &
    RMSE &
    $\delta<1.25$ &
    $\delta<1.25^2$ &
    $\delta<1.25^3$\\
    \hline
    \multirow{3}{*}{\textbf{[00m 10m]}}
    & Monodepth2~\cite{Godard2019} & - &  0.066  &   0.264  &  0.106& 1.076    &   0.959  &   0.987  &   0.994  \\
    & ForecastMonodepth2 & 0.5s &  0.138  &   \textbf{0.586} &   0.178 &   1.697 &   0.847  &   0.957  &   0.985  \\
    & \textbf{Ours} & 0.5s &   \textbf{0.112}  &   0.595  &  \textbf{0.155} & \textbf{1.573}    &   \textbf{0.893}  &   \textbf{0.964}  &   \textbf{0.986}  \\
    \hline
    \multirow{3}{*}{\textbf{[10m 30m]}}
    &Monodepth2~\cite{Godard2019} & - &  0.119  &   0.858   &   0.192  &   3.706 &   0.876  &   0.956  &   0.978   \\
    & ForecastMonodepth2 &0.5s&0.192  &   1.673  &   0.258  &   5.169   &   0.725  &   0.906  &   0.963  \\
    & \textbf{Ours} &0.5s&\textbf{ 0.167}  &  \textbf{ 1.453} &   \textbf{0.241 }&   \textbf{4.803}    & \textbf{  0.782}  &   \textbf{0.921}  &  \textbf{ 0.965}  \\
    \hline
    \multirow{3}{*}{\textbf{[30m 80m]}}
    &Monodepth2~\cite{Godard2019} & - &  0.188  &   3.094  &  11.115  &   0.288  &   0.709  &   0.897  &   0.950   \\
    & ForecastMonodepth2 &0.5s& \textbf{0.213}  &   \textbf{3.526}  &  \textbf{11.940}  &   \textbf{0.292 } &   \textbf{0.631}  &   \textbf{0.874}  &   \textbf{0.953}  \\
    & \textbf{Ours} &0.5s& 0.224  &   4.052  &  12.638  &   0.312  &   0.622  &   0.862  &   0.941  \\
    \hline
\end{tabular} }
\caption{Quantitative performance comparison on the KITTI benchmark with Eigen split~\cite{Geiger2012CVPR} for multiple distances range. For Abs Rel, Sq Rel, RMSE and RMSE log lower is better, and for $\delta < 1.25$, $\delta < 1.25^2$ and $\delta < 1.25^3$ higher is better. Three ranges are considered: short range [0 10m] which represents 37.95\%, medium-range [10 30]which represents 50.74\% and long-range[30 80] which represents 11.30\%. The results shows that the proposed method is able to forecast good depth and outperform the baseline at short and medium forecasting range.}
\label{tab:distance_filtred}

\end{table*}

In order to further analyze the depth forecasting results, an assessment based on the ground-truth LiDAR distance is done.~Table IV shows the comparison of the non-forecasting method Monodepth2~\cite{Godard2019}, ForecastMonodepth2 and the proposed methods. 

The results suggest that the proposed method outperform the adaptation of Monodepth2 for short-range with a improvement of the Abs Rel of $-16.7\%$ and medium-range with an improvement of the Abs Rel of $-8.8\%$. These regions are the most significant regions of the forecasting as they have enough parallax for the ego-motion and dynamic object motion. Besides, this region assesses several challenges, including out-of-view objects and occlusion. For the long-range forecasting, the results show that the two methods perform badly due to the lack of parallax in this region and down-sampling that ignores small objects. Moreover, this region has a high likelihood of new-objects entering the scene, which the forecasting is unable to handle by definition. The reported performances and the qualitative results suggest that the two forecasting networks only fit the road and completely ignore any other object. These results are shown qualitatively in \reffig{qualitative_2}.

\subsection{Ego-Motion forecasting results}
\reftab{pose} shows the results of the proposed network on the KITTI odometry benchmark~\cite{Geiger2012CVPR}. Similar to depth, the assessment is made by comparing with non-forecasting prior works. To avoid data leakage, the network is trained from scratch on the sequences 00-08 of the KITTI odometry benchmark. 
The network takes only the context images $\mbf{I}_c$ and forecasts $\pose{t}{t+n}$. The ATE results in \reftab{pose} show the proposed network achieved a competitive result relative to other non-forecasting approaches. All the methods are trained in the monocular setting, and therefore scaled at test time using ground-truth. These results suggest that using the proposed architecture along with the self-supervised loss function successfully learns the future joint depth and ego-motion.

\begin{table}
\centering
{
\small
\setlength{\tabcolsep}{0.3em}
\begin{tabular}{c|cccc}
\hline
\textbf{Method} &
Abs Rel &
Sq Rel &
RMSE log & 
RMSE 
\\
\hline
\textbf{Ours} & 0.178  &   1.645    &   0.257&   6.196 \\
(a) Without TAM & 0.205  &   1.745    &   0.296 &   6.565 \\
(b) Shared pose/depth features & 0.208  &   1.745    &   0.282  &   6.529 \\
(c) Single scale & 0.208  &   1.950  &   0.283   &   6.595  \\
(d) Disable auto-masking &  0.193  &   1.774   &   0.273  &   6.374 \\
\hline
\end{tabular}
}
\caption{Ablation study results showcasing the effects of different modules in the proposed method. (a) Effect of the Temporal Aggregation Module (TAM) on performance metrics. The TAM module significantly improves performance across all metrics by better encoding the spatio-temporal relationship between images. (b) Effect of sharing the encoder of depth and ego-motion networks. Sharing the encoder leads to degradation in performance as it restricts the network from finding the best local optima for both tasks. (c) The benefit of using multiple scales in the proposed method. The network benefits from the multiscale approach, as demonstrated by improved results compared to using a single scale. (d) Effect of auto-masking on forecasted depth. Auto-masking improves all evaluation criteria by rejecting outliers that hinder optimization and consequently enhancing accuracy.}
\label{tab:ablation}
\end{table}

\subsection{Ablation study}
\label{sec:ablation}
To further analyse the network, several ablations are made.~\reftab{ablation} depicts a comparison of the proposed model with several variants. The evaluation is done for short-term forecasting $n=5$ using $k=4$ context frames.

\subsubsection{Effect of the Temporal Aggregation Module} 
In order to evaluate the contribution of the multi-head attention, a variant of the proposed method is designed by replacing the TAM module by a simple concatenation of the last layer features. From \reftab{ablation}, the improvement induced by the TAM module is significant across all metrics. These results suggest that the performance obtained by the proposed method is achieved through the TAM module. Since the TAM aggregates the temporal information across all frames using a learned attention, the temporal features are better correlated and the final representation successfully encodes the spatio-temporal relationship between the images.

\subsubsection{Effect of sharing the encoder of depth and ego-motion}
Since both pose and depth networks encode the future motion and geometry of the scene, it is expected that sharing the encoders of these networks yield better results. However, as reported in~\reftab{ablation}, the degradation is significant. Even though these tasks are collaborative, sharing the encoder will result in a set of parameters $\widehat{\bm{\theta}}$ that are neither the best local optima for the depth nor for the pose. By alleviating this restriction and separating the encoders, the network learns better local optima for both pose and depth.

\subsubsection{The benefit of using multiple scales} 
In order to evaluate the multiscale extension, a variant of the proposed method that uses only one scale is trained. As illustrated in the \reftab{ablation}, the network benefits from the multiscale. The reverse warping 
uses bi-linear interpolation. As mentioned earlier, each depth point depends only on the four neighboring warped points. By using a multiscale depth at training-time the gradient is derived from a larger spatial region directly at each scale. 

\subsubsection{Effect of auto-masking}
\reftab{ablation} compares the proposed method with a variant without using the auto-masking defined is~\refsec{objetives}. The results show that using auto-masking improves all four evaluation criteria. This demonstrates that, using auto-masking, rejects these outliers that inhibit the optimization. This leads to better accuracy of the forecasted depth.

\section{Discussion}
\label{sec:icpr-conclusion}
The work presented in this chapter proposed an approach for forecasting future depth and ego motion using only raw images as input. This problem is addressed as end-to-end self-supervised forecasting of the future depth and ego motion. Results showed significant performances on several KITTI dataset benchmarks~\cite{Geiger2012CVPR}. The performance criteria are even comparable with non-forecasting self-supervised monocular depth inference methods~\cite{Eigen,Liu2016,Zhou2017,yang2018unsupervised}. The proposed architecture demonstrates the effectiveness of combining the inductive bias of the CNN as a spatial feature extractor and the multi-head attention of transformers for temporal aggregation. The proposed method learns a spatio-temporal representation that captures the context and the motion of the scene.

\subsection{Limitations and perspectives}
Even though the proposed forecasting method yields good results, there exists a gap with respect to non-forecasting methods. Several limitations contribute to this:
\begin{itemize}
    \item A common assumption across the presented methods is that the environment is deterministic and that there is only one possible future. However, this is not accurate since there are multiple plausible futures. Given the stochastic nature of the forecasting proposed here, the network will tend to forecast a blurry depth map that represents the mean of all the possible outcomes~\cite{Babaeizadeh2018}.  
    
    \item  The network does not forecast the correct boundaries of the objects. This is due to the formulation as a maximum likelihood problem with a Laplacian distribution assumption and the deterministic nature of the architecture. As a result, the boundaries of the dynamic objects are smoothed. 
    
    \item Due to the problem formulation, the scale of the forecast depth is ambiguous. this is a fundamental problem to the monocular methods. As the distance of the camera to the floor is constant, this could be used to disambiguate the scale.
    
    \item The model fails to account for the motion of distant dynamic objects due to lack of parallax.
    
\end{itemize}

\chapter*{}

\chapter{Video-to-video future depth with spatio-temporal consistency}
\label{ch:ch6}
In the previous chapters, depth inference and forecasting were explored. However, our research was accompanied by several limitations that were discovered along the way. This chapter delves deeper into these limitations.

One of the primary limitations in the previous depth inference work was the fact that most methods do not take advantage of multiple frames as input. They rely on the scene clues such as the object shape prior. A model can better understand the geometry of the scene and better understand the motion of the ego and the dynamic objects by utilizing multiple frames as input. As for depth forecasting, one of the biggest limitations we encountered was that the model tends to output a blurry output that is a mean of all possible future situations and a single future depth. Moreover, these models do not take into account the motion of objects in the scene, this can have a significant impact on future depth estimates. To address the limitations, development of more sophisticated models is required to accurately predict the depth of a scene.

In this chapter, a self-supervised model that simultaneously predicts a sequence of future frames from video input with a novel spatial-temporal attention (ST) network is proposed. The ST transformer network allows constraining both temporal consistency across future frames whilst constraining consistency across spatial objects in the image at different scales. This was not the case in prior works for depth prediction, which focused on predicting a single frame as output. The proposed model leverages prior scene knowledge such as object shape and texture similar to single-image depth inference methods, whilst also constraining the motion and geometry from a sequence of input images. Apart from the transformer architecture, one of the main contributions with respect to prior works lies in the objective function that enforces spatio-temporal consistency across a sequence of output frames rather than a single output frame. As will be shown, this results in more accurate and robust depth sequence forecasting. The model achieves highly accurate depth forecasting results that outperform existing baselines on the KITTI benchmark. Extensive ablation studies were performed to assess the effectiveness of the proposed techniques. One remarkable result of the proposed model is that it is implicitly capable of forecasting the motion of objects in the scene, rather than requiring complex models involving multi-object detection, segmentation, and tracking. In the \refsec{new-intro}, we motivate our method. We present our approach in \refsec{new-method} and experimental results in \refsec{new-results}. We conclude in \refsec{new-conclusion}.
 
This chapter was based on the following paper: 

\begin{itemize}
    \item \textbf{To be submitted:} Boulahbal Houssem Eddine, Adrian Voicila, and Andrew Comport. "STDepthFormer: Predicting Spatio-temporal Depth from Video with a Self-supervised Transformer Model." arXiv preprint arXiv:2303.01196 (2023). 

\end{itemize}

\section{Introduction}
\label{sec:new-intro}


State-of-the-art approaches, such as~\cite{Watson2021,guizilini2022multi,boulahbal2022forecasting} (see Chapter~\ref{ch:ch4}), have developed models that output a single depth image. The underlying model is then used to perform inference or forecasting tasks separately. These approaches are, however, limited because they cannot enforce spatio-temporal consistency in the output, as they do not predict a sequence. By introducing a model that predicts a sequence of depth images, the model proposed here can apply motion and geometric constraints to the output which improves the accuracy and sharpness of the forecasting and forces the predicted images to be more deterministic (ie. it does not average across possible future outcomes as in prior works). 


On one hand, the majority of self-supervised monocular depth inference methods~\cite{Eigen,boulahbal2022instance,safadoust2021self,lee2021attentive,Wang2021,Ranjan2019,Gordon2019,Godard2017,Zhou2017,Godard2019,Johnston2020,Rares2020} rely on a single frame as input. While this approach is effective at leveraging prior knowledge such as object shape and textures, it is limited in its ability to learn the geometry and the motion of the scene. By contrast, using multiple frames~\cite{Watson2021,guizilini2022multi,9864127} as input has the potential to provide a more comprehensive view of the scene and to help the model better understand the relationships between objects and their motions.

Depth forecasting self-supervised methods~\cite{Mahjourian2017,Qi2019,Hu2020,boulahbal2022forecasting}, on the other hand, often produce a blurry depth map that represents the mean of all possible future scenarios~\cite{boulahbal2022forecasting}. This approach fails to produce an accurate depth, which limits its usefulness in decision-making contexts.

To address these limitations, a self-supervised model is proposed that can simultaneously output a depth sequence encompassing inference and forecasting. By using multiple image frames as input and output, the model can learn about the geometric consistency of the scene, which enables it to predict more accurate depth sequences as output. The proposed model enforces a spatio-temporal consistency in the output depth sequence by warping neighboring images onto the target image using a geometric and photometric warping operator that depends on the output depths. As will be detailed further in \refsec{objectivefn}. The results show that this effectively constraints the output depth forecasting to choose the most probable outcome of the future depth, instead of using the mean of all outcomes, avoiding the issue of blurry depth maps and leading to more precise depth. 

As a result, the proposed approach produces an accurate depth sequence. In summary, the contributions of the proposed method are:

\begin{itemize}
    \item A self-supervised model that predicts a spatially and temporally consistent depth sequence that captures both present and future depth information, allowing for more comprehensive and accurate depth.
    \item A transformer-based multi-frame architecture that implicitly learns the geometry of the scene in an image-based end-to-end manner. Interestingly, the proposed model is capable of forecasting the motion of objects in the scene, even in the absence of explicit motion supervision.
    \item The method achieves highly accurate depth forecasting results that outperform existing baselines in the KITTI~\cite{Geiger2012CVPR} benchmark.
    \item Improved generalization for depth inference tasks over SOTA.     
    \item A comprehensive analysis of the proposed method is conducted through several ablation studies.

\end{itemize}

\begin{figure*}
    \centering
    \includegraphics[width=\textwidth]{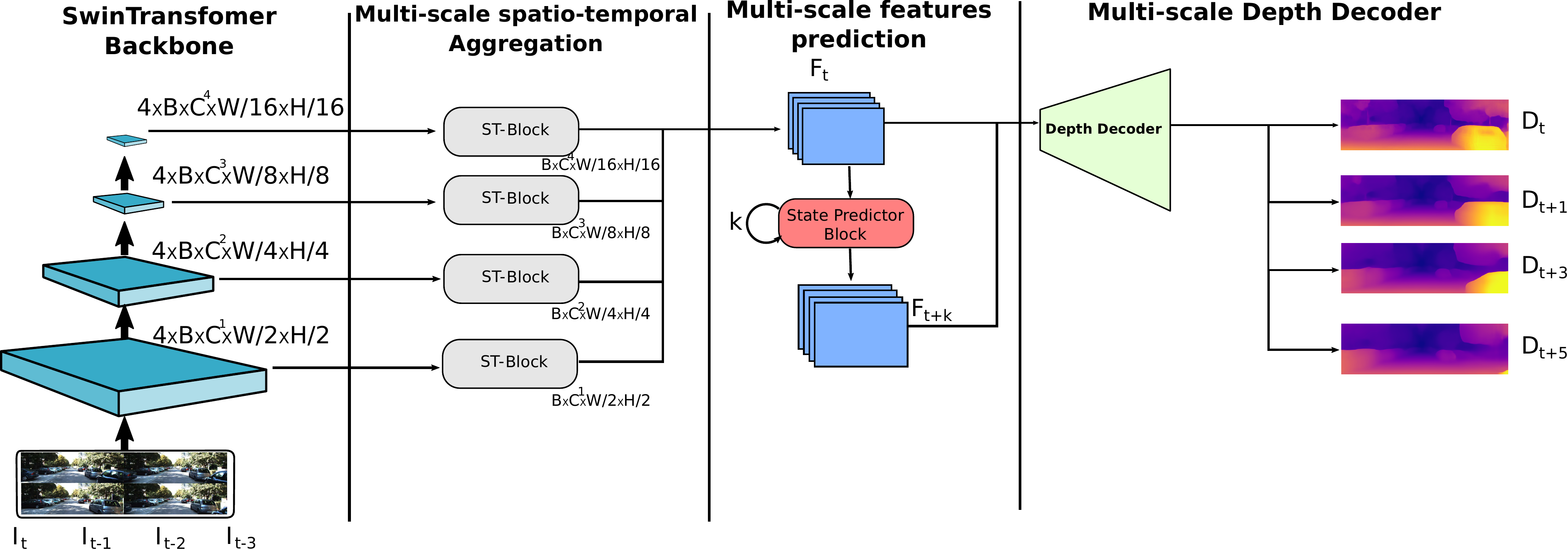}
    \caption{Architecture of the proposed method. The network comprises four stages. Firstly, the spatial feature of each frame is extracted using a SwinTransformer backbone shared across the context frames. Secondly, the features are correlated spatio-temporally using the ST-block shown in \reffig{spblock}. Thirdly, a learned function $f$ is used to transition from $F_{t+k-1}$ to $F_{t+k}$, and this module consists of SwinTransformer blocks as well. Finally, the depth decoder employs skip connections to utilize multi-scale features and outputs 4 depth states: $(\mbf{D}_{t}, \mbf{D}_{t+1} ,\mbf{D}_{t+3} \mbf{D}_{t+5})$}
    \label{fig:architecture-chapter3}
\end{figure*}

\section{Method}
\label{sec:new-method}
\subsection{Problem formulation}
The aim of monocular depth inference and forecasting is to predict an accurate depth sequence through the mapping, $\mbf{D}_{t:t+n} =~f(\img{t-k:t}; \bm{\theta})$ where $\img{t-k:t}$ are the $k$ context images and $\mbf{D}_{t:t+s}$ are the $s$ depth target states. In self-supervised learning, this model is trained via novel view synthesis by warping a set of source frames $\mbf{I}_{src}$ to the target frame $\mbf{I}_{tgt}$ using the learned depth $\mbf{D}_{tgt}$ and the target to source pose $\pose{src}{tgt} \in \mathbb{SE}[3]$~\cite{jaderberg2015spatial}. The differentiable warping is defined in~\ref{eq:warping_classic}
\begin{equation}
        \widehat{\pt}_{src} \sim \pi (\mbf{K} \pose{src}{tgt} H(\mbf{D}_{tgt}  \mbf{K}^{-1} \pt_{tgt}))
\end{equation}
The depth network takes $k=4$ context images as input. With $k=4$, it is possible to learn the velocity and the acceleration without exploding the memory. The network produces $tgt=\{0,1,3,5\}$ depth outputs. As the pose network is only used for supervision during training, providing the future images will help the pose network to learn better. Therefore, for each depth state $tgt$, the pose network input is the triplet of images $(tgt-1, tgt, tgt+1)$. It outputs two poses, $\pose{tgt-1}{tgt}$ and $\pose{tgt}{tgt+1}$.

\subsection{Architecture}

The proposed model is related to classic \textit{structure-from-motion}. During training, self-supervision is achieved by using an image warping function, and two networks are used: a pose network and a depth network. At test time, only the depth network is used to output the depth.
\subsubsection{The depth network}
\reffig{architecture-chapter3} shows the architecture of the proposed method. The depth network uses $k=4$ context inputs. The architecture comprises four stages: 

\textbf{1. Spatial feature extraction:} SwinTransformer backbone~\cite{liu2021swin} is used to extract the features of each frame. The swin-tiny variant is used with a number of layers : [$2, 2, 6, 2$], with depths of [$3, 6, 12, 24$], a patch embedding channel of $7$, and an embedding dimension of $96$. It is pretrained on the ImageNet dataset~\cite{deng2009imagenet}. See~\refsec{swin_transfomer} for more details. This feature extractor is shared across the context frames. The feature map at each scale is extracted as input for the next module. As the purpose of this module is to extract spatial information only, calculating the gradient for only one context frame is sufficient. Experimentally, no differences were observed between calculating the gradient for all four frames and only one frame. Therefore, backpropagation is carried out only on the first frame to minimize the memory footprint.
    
\begin{figure}
    \centering
    \includegraphics[width=\textwidth]{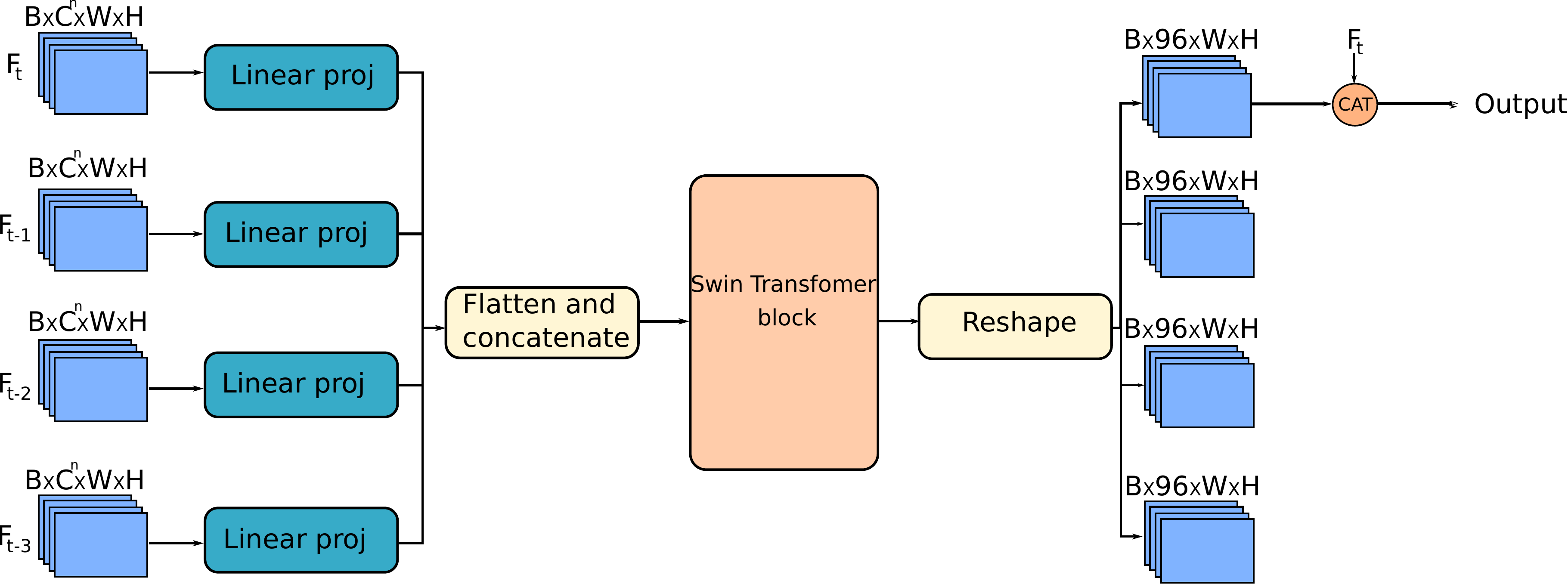}
    \caption{Architecture of a multiscale spatio-temporal aggregation network using linear projection and SwinTransformer layers for feature spatio-temporal correlation.}
    \label{fig:spblock}
\end{figure}

\textbf{2. Multi-scale spatio-temporal aggregation:} Next, the features are correlated spatio-temporally using the proposed novel ST-block. \reffig{spblock} shows the architecture of this fusion block. At each feature scale, each feature map of each frame is projected using a \textit{Conv2D} with $kernal=1$ outputting an embedding of dimension $96$. These features are concatenated as a sequence of patches to construct embeddings that will be used as input to the transformers. This sequence is then provided to the transformer~\cite{liu2021swin}. This block has a depth of $2$ and embeddings of $96$ and the number of heads at each scale is [$3, 6, 12, 24$] from high to low resolution. The attention map performs the spatio-temporal correlation of these features. The sequence is reshaped to its original shape and the first feature map is contacted with the context feature $F_t$. Finally, another projection layer outputs the spatio-temporal features to recover the channel to $C^n$.

\begin{figure*}
    \centering
    \includegraphics[width=\textwidth]{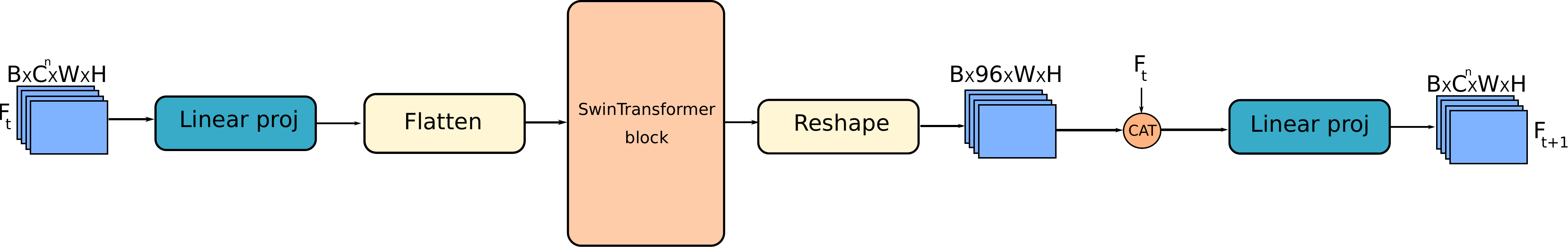}
    \caption{SwinTransformer-based state predictor block. The input feature map $F_t$ is projected onto an embedding dimension of size $96$ and flattened into patches for the SwinTransformer block. The output is reshaped and concatenated with the input using a skip connection. A linear projection generates the features of $Ft+1$ with size $B \times C_n \times W \times H$ where $C_n$ is the original channel}
    \label{fig:temporal-block}
\end{figure*}

\textbf{3. Multiscale feature prediction:} A transition function $f$ is used to relate each feature to a state in the output sequence. At each scale, this learned function $f$ is used to transition from $F_{t+k-1}$ to $F_{t+k}$. This function is recursive and defined as:
        \begin{align}
            F_{t+k} &= f(F_{t+k-1}) 
        \end{align}
This module is composed of SwinTransformer blocks~\cite{liu2021swin}. It is shared and used recursively across all $n$ frames to be forecast. \reffig{temporal-block} shows the architecture of the state predictor block. The input feature map $F_t$ is projected to have an embedding dimension of $96$. This map is flattened to patches of size $1$ to be used as input to the SwinTransfomer block. Similarly, this block has a depth of $2$, embeddings of $96$ and the number of heads of each scale is [$3, 6, 12, 24$] from high to low resolution. The output is reshaped to its original dimensions and concatenated with the input with a skip connection. A linear projection is used to obtain the features of $F_{t+1}$ with size: $B\times C^n \times W \times H$. 

\textbf{4. Depth decoder: } This module is shared across all state features. It consists of Transposed2DConvolution with ReLU as activation and a kernel size of $k=3$, which is similar to~\cite{Godard2019}. Skip connections are employed since the previous stage outputs multi-scale features. In this method, four depth states are output: $(\mbf{D}_{t}, \mbf{D}_{t+1} ,\mbf{D}_{t+3}, \mbf{D}_{t+5})$.

\subsubsection{The pose network}
This network is an off-the-shelf model taken from~\cite{Godard2019} that takes the triplet $(tgt-1, tgt, tgt+1)$ and outputs two poses: $\pose{tgt-1}{tgt}$ and $\pose{tgt}{tgt+1}$. This model is used only for self-supervised training and is discarded at evaluation. 

It is worth noting that current state-of-the-art methods utilize a plane sweep approach, such as the one proposed by \cite{Watson2021,guizilini2022multi}, that involve explicitly providing the pose and camera parameters to the depth network and constructing a matching volume during training and evaluation. Alternatively, the proposed method adopts a different approach that learns this information implicitly. This presents several benefits, most notably the ability for the two networks, the depth and the pose, to operate independently. This independence from the pose network and camera parameters is particularly significant, as it allows the proposed network to generalize better and perform more robustly. Empirical evidence supporting this claim is presented in~\refsec{generalization}, where the experimental results demonstrate the superiority of the proposed approach.

\subsection{Objective functions}
\label{sec:objectivefn}
As the self-supervision is done by reconstructing the frames $\mbf{I}_{tgt}$ such as $tgt \in \{0,1,3,5\}$ using the depth and the pose with the warping, this can be formulated as a maximum likelihood problem: 
 \begin{align}
 \footnotesize
     \widehat{\bm{\theta}} &= \argmax_{\bm\theta \in \bm{\Theta}} L(\mbf{I}_{t+5}, \mbf{I}_{t+3} ,\mbf{I}_{t+1} ,\mbf{I}_{t}| \mbf{I}_{t-4:t+6}; \bm{\theta} ) \\ \notag 
     \widehat{\bm{\theta}} &\equiv	 \argmax_{\bm\theta \in \bm{\Theta}} \sum_{m} P_{model}(\mbf{I}_{t+n}^m| \mbf{I}_{c}^m,\mbf{I}_{f}^m)
 \end{align} 
 $\bm\theta$ is the model parameters, $\mbf{I}_{c}^m$ are the context frames that will be provided to the depth network and $\mbf{I}_{f}^m$ are the future frames that will be provided to the pose network. As presented in the~\refsec{loss_function_ch3}, the photometric loss and structural similarity index measure SSIM~\cite{wang2004image}, along with the depth smoothness, are used to optimize the parameters. 
\begin{align}
\footnotesize
  pe(\img{tgt}, \widehat{\mbf{I}}_{(tgt\pm1 \rightarrow tgt)} )= \sum_{\pt} \big[&(1 - \alpha) \textup{ SSIM}[ \img{tgt}(\pt) - \widehat{\mbf{I}}_{(tgt\pm1 \rightarrow tgt)}(\pt)]  \\ \notag
      & +  \alpha | \img{tgt}(\pt) - \widehat{\mbf{I}}_{(tgt\pm1 \rightarrow tgt)}(\pt) |\big]
\end{align}
Such that $\widehat{\mbf{I}}_{(tgt\pm1 \rightarrow tgt)}$ is reconstructed from two views : $\img{tgt-1}$ and $\img{tgt+1}$. Similar to~\cite{Godard2019}, the minimum projection loss of the two frames is used to handle occlusions leading to: 
\begin{align}
\mathcal{L}_\textup{ph}(\img{tgt})= min\big[pe(\img{tgt}, \widehat{\mbf{I}}_{(tgt-1 \rightarrow tgt)} ) , pe(\img{tgt}, \widehat{\mbf{I}}_{(tgt+1 \rightarrow tgt)} ) \big]
\end{align}
To further improve the training, outlier rejection is performed. Similar to~\cite{Godard2019} this is done using auto-masking which is defined as: 

\begin{align}
\footnotesize
    \mu  = \big[ min_{tgt}\big(pe(\img{tgt}, \widehat{\mbf{I}}_{(tgt\pm 1\rightarrow tgt)}), pe(\img{tgt}, \mbf{I}_{(tgt\pm1)}) \big) \big]
\end{align}
where [ ] is the Iverson bracket. $\mu$ is set to only include the loss of pixels where the re-projection error
of the warped image $\widehat{\mbf{I}}_{(tgt\pm 1\rightarrow tgt)}$ is lower than that of the original, unwarped image $\mbf{I}_{(tgt\pm1)}$. An edge-aware gradient smoothness constraint is used to regularize the photometric loss. The disparity map is constrained to be locally smooth.  
\begin{align}
\footnotesize
    \begin{aligned}
      \mathcal{L}_{s}(D_{tgt}) = \sum_{p} \big[ &| \partial_x D_{tgt}(\pt) | e^{-|\partial_x \mbf{I}_{tgt}(\pt)|}    +
      &| \partial_y D_{tgt}(\pt) | e^{-|\partial_y \mbf{I}_{tgt}(\pt)|} \big] 
    \end{aligned}
\end{align}
Temporal consistency is enforced during training through a loss function that enforces geometric constraints simultaneously across multiple output frames, namely $\mbf{D}_{t+5}$, $\mbf{D}_{t+3}$, $\mbf{D}_{t+1}$, and $\mbf{D}_{t}$ via a warping function. The image-based loss function minimizes the pair-wise photometric consistency by warping neighboring images for each central target ($\img{t+5}$, $\img{t+3}$, $\img{t+1}$, and $\img{t}$). The warping function depends on the output depth and pose outputs from the network. This constrains the model to respect image consistency between these frames. For example, $\img{t+1}$ is minimized with respect to the warped $\widehat{\mbf{I}}_{t}$ and the warped $\widehat{\mbf{I}}_{t+2}$. The gradient locality problem~\cite{Zhou2017} is handled using a pyramid of depth outputs, and the optimization is done on all these levels. The final loss function is defined as: 
\begin{align}
      \mathcal{L} = \frac{1}{m}\sum_{m}\sum_{tgt}\sum_{l=1}^{l=4} \mu \mathcal{L}_\textup{ph}(\img{tgt}) + \alpha_s \mathcal{L}_{s}(D_{tgt})
\end{align}
where $m$ is the batch size, $tgt\in{0,1,3,5}$ and $l$ represents the multiscale output depth.
 \section{Results}
\label{new-results}

\begin{figure*}
    \centering
    \includegraphics[width=\textwidth]{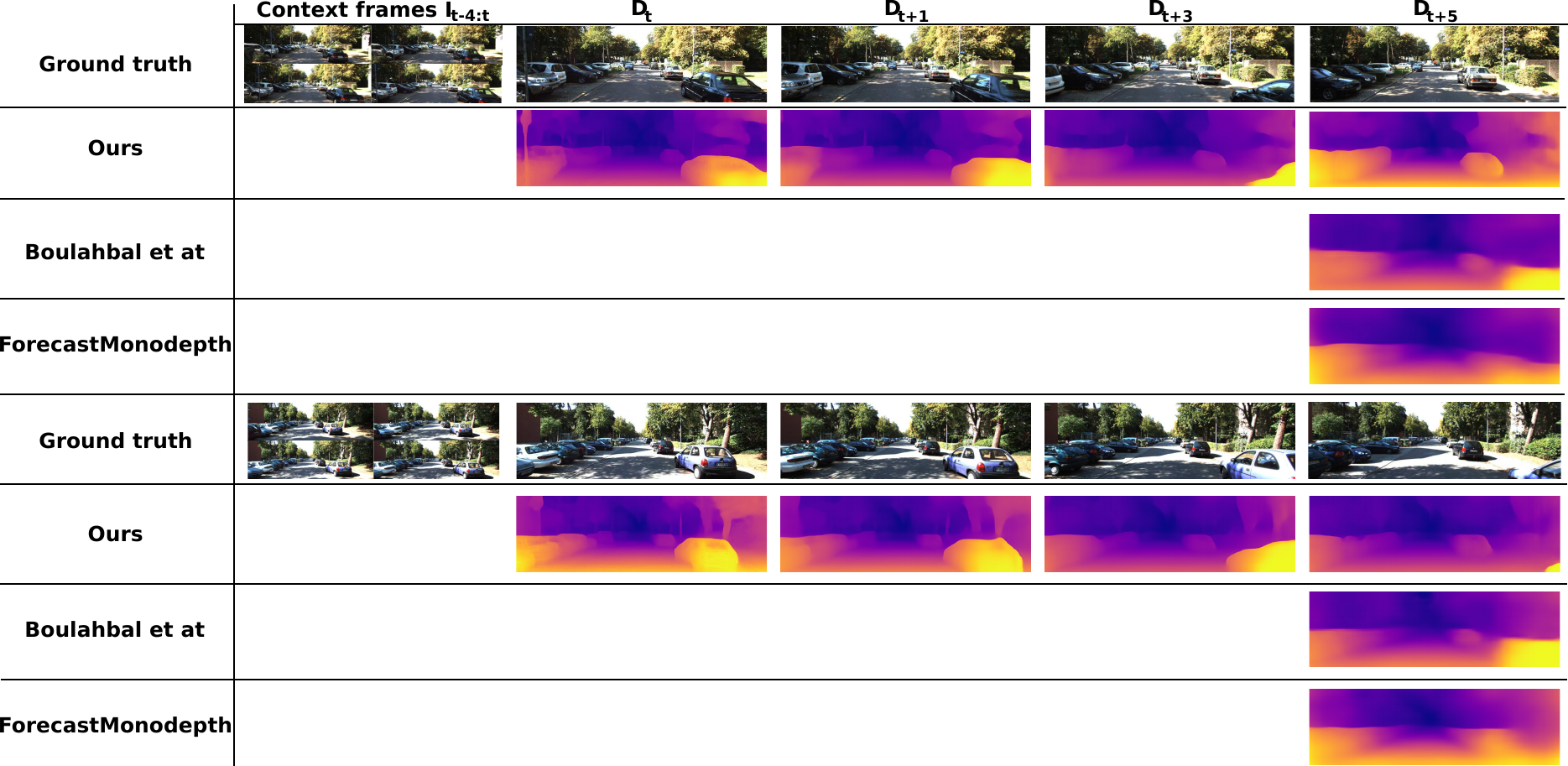}
    \caption{ Qualitative comparison of the proposed method and the prior work on KITTI Eigen test benchmark. The proposed method is able to generate an accurate future depth sequence that exhibits significantly more details compared to the prior work. The depth map generated by the proposed method is remarkably sharp and not blurry. This superior performance can be attributed to the fact that the proposed method was specifically trained for depth inference with spatio-temporal consistence across the forecast range, resulting in an enforced deterministic output. As a result, the proposed approach predicts the most probable future instead of averaging all possible futures, as done in the prior work.}
    \label{fig:qualitative_perf}
\end{figure*}

\label{sec:new-results}
\subsection{Experimental setup}

\textbf{Datasets:} The KITTI benchmark~\cite{Geiger2012CVPR} is the defacto benchmark for evaluating depth methods. The Eigen \ea~\cite{Eigen} is used with Zhou \ea\cite{Zhou2017} preprocessing to remove static frames. In order to test the generalization of the method, the Cityscapes~\cite{cordts2016cityscapes} and the Robotcar~\cite{RobotCarDatasetIJRR} datasets are used. The Cityscapes dataset does not provide ground truth LiDAR depth and uses the classical SGM method~\cite{4359315} to obtain the depth. The LiDAR depth is projected onto the image to obtain the ground truth for the Robotcar dataset.

\textbf{Baselines:} Several depth inference method were used for the comparison~\cite{Watson2021,guizilini2022multi}. For forecasting, a comparison is made only with methods that perform self-supervision with respect to frame $5$ as done in~\cite{boulahbal2022forecasting}. To test the performance of the method with respect to dynamic objects, the analysis provided in~\cite{boulahbal2022instance} is used.

\textbf{Hyperparameters:} The networks are trained for $6$ epochs, with a batch size of $4$. The Adam optimizer~\cite{kingma2014adam} is used with a learning rate of $lr = 10^{-4}$ and $(\beta_1, \beta_2) = (0.9, 0.999)$. The SSIM weight is set to $\alpha = 0.15$ and the smoothing regularization weight to $\alpha_s = 0.001$. $l = 4$ scales are used for each output of the decoder. At each scale, the depth is upscaled to the target image size. The input images are resized to $192 \times 640$. Two data augmentations were performed: horizontal flips with probability $p = 0.5$ and color jitter with p = 1. The activation of depth decoder, $\sigma$, is re-scaled to obtain the depth $D = \frac{1}{a \sigma +b}$, where $a$ and $b$ are chosen to constrain $D$ between $0.1$ and $100$ units for training and $(0.5,100)$ for evaluation, similar to~\cite{boulahbal2022instance}. The scale ambiguity is resolved using median scaling, similar to the prior work~\cite{Godard2019,Watson2021}.

\begin{table*}
\centering
\footnotesize
\begin{tabular}{|l|c||c c c c| c c c|}
\hline 
Predicted frame & Method &Abs Rel& Sq Rel&  RMSE & RMSE log  &
$\delta<1.25$ &
$\delta<1.25^2$ &
$\delta<1.25^3$\\
\hline \hline
$\mbf{t=0}$ 
& SfMLearner~\cite{Zhou2017} &0.198 & 1.836& 0.275  & 6.565 & 0.718 & 0.901 & 0.960\\
& Yang \ea~\cite{yang2018unsupervised} & 0.182 & 1.481  & 0.267& 6.501 & 0.725 & 0.906 & 0.963\\
& GeoNet~\cite{Yin2018} & 0.155 & 1.296 & 5.857  & 0.233 & 0.793 & 0.931 & 0.973\\
& CC~\cite{Ranjan2019} & 0.140 & 1.070  & 5.326  & 0.217 &0.826 &0.941 &0.975 \\

& Monodepth2~\cite{Godard2019}& 0.115 & 0.903 & 4.863 & 0.193 & 0.877 & 0.959 & 0.981\\
& Lee \ea~\cite{lee2021attentive}  &  0.113 & 0.835 & 4.693 & 0.191 & 0.879 & 0.961 & 0.981 \\
& PackNetSfm~\cite{Rares2020} & 0.111 & 0.785 & 4.601 & 0.189 & 0.878 & 0.960 & 0.982\\
& Manydepth~\cite{Watson2021}& \textbf{0.098} & \textbf{0.770} & \textbf{4.459} & \textbf{0.176} & \textbf{0.900 }& \textbf{0.965} & \textbf{0.983} \\

& \textbf{Ours} & 0.110  &   0.805  &   4.678  &   0.187  &   0.879  &   0.961  &   0.983 \\
\hline
$\mbf{t=1}/ 0.1sec$ & \textbf{Ours} & 0.121  &   0.989  &   5.026  &   0.203  &   0.863  &   0.951  &   0.978 \\
\hline
$\mbf{t=3}/ 0.3sec$ & \textbf{Ours} & 0.146  &   1.295  &   5.493  &   0.227  &   0.824  &   0.935  &   0.971 \\
\hline
$\mbf{t=5}/ 0.5sec$ 
&
ForecastMonodepth2~\cite{boulahbal2022forecasting} & 0.201  &   1.588   &   6.166   &   0.275 &   0.702 &  0.897  &   0.960\\

& Boulahbal~\ea~\cite{boulahbal2022forecasting}&  0.178  &  1.645   &   6.196 &   0.257  &   0.761  &   0.914  &   0.964  \\

& Ours & \textbf{0.165}  &   \textbf{1.489}  &   \textbf{5.805  }&  \textbf{ 0.245}  &   \textbf{0.792}  &   \textbf{0.921 } &  \textbf{ 0.964 }\\
\hline 
\end{tabular}
\caption{Quantitative performance of the proposed method on the KITTI benchmark~\cite{Geiger2012CVPR} with eigen~\cite{Eigen} benchmark for the frames $D_t, D_{t+1}, D_{t+3}, D_{t+5}$. for Abs Rel, Sq Rel, RMSE and RMSE log lower is better. For $\delta<1.25$, $\delta<1.25^2$, $\delta<1.25^3$ higher is better. The proposed method is able to output an accurate depth at different time steps. The performance of the future depth is even comparable to depth inference method that have access to the target frame.  }
\label{tab:results-1}
\end{table*}




\begin{table}
\begin{center}
\footnotesize
\resizebox{\columnwidth}{!}
{
\begin{tabular}{|c|c|c|c|c|c|}
    \hline
    Evaluation& Model & Abs Rel & Sq Rel & RMSE & RMSE log  \\ 
    \hline
    All points mean &
    ManyDepth \cite{Watson2021} & 0.098 & 0.770 & 4.459 & 0.176   \\
    &
    Boulahbal~\ea~\cite{boulahbal2022instance} &  0.110  &   0.719  &   4.486  &   0.184 \\
 &
    Ours & 0.110  &   0.805  &   4.677  &   0.187  \\
 &

    Ours + stereo &  0.107  &   0.751  &   4.805  &   0.189 \\

    \hline
    \hline 
    Only dynamic &
    ManyDepth \cite{Watson2021} &   0.192  &   2.609  &   7.461  &   0.288   \\
    &
    Boulahbal~\ea~\cite{boulahbal2022instance} & 0.167  &   1.911  &   6.724 &   0.271 \\
&
    Ours &  0.178  &   2.089  &   6.963  &   0.278  \\
     &

    Ours + stereo &  0.155  &   1.668  &   6.401  &   0.260  \\
    
    \hline
    \hline 
    Only static &
    ManyDepth\cite{Watson2021} &  0.085  & 0.613  &   4.128 &   0.150    \\ &
    
    Boulahbal~\ea~\cite{boulahbal2022instance} & 0.101  & 0.624  &   4.269  &   0.163 \\
    &
    Ours  & 0.099  &   0.684  &   4.462  &   0.165\\
     &

    Ours + stereo & 0.099  &   0.684  &   4.679  &   0.173 \\
    \hline
    \hline
    Per category mean &
    ManyDepth\cite{Watson2021} &   0.139   &  1.611 & 5.794   &  0.219\\ &
    Boulahbal~\ea~\cite{boulahbal2022instance} & 0.134  &  1,267  &   5,496 & 0,217  \\
     &
    Ours & 0.138  &   1.386  &  5.712  &   0.222 \\
     &

    Ours + stereo & 0.127 &  1.176  &  5.540 &   0.217 \\
    \hline
    \end{tabular}
}
\caption{Quantitative performance comparison for dynamic and static objects at $t=0$0. The proposed method outperforms the SOTA~\cite{Watson2021} on the dynamic objects. The stereo variant is the best model for the dynamic and the per category mean.}
\label{tab:stdepth_staticVSdynamic}

\end{center}
\end{table}
\subsection{Multi-step depth forecasting results}
The findings of the study reveal that the proposed method exhibits a faster convergence rate, with a reduced number of epochs compared to prior works. Specifically, the proposed method achieves convergence in just $6$ epochs, whereas previous approaches required $20$ epochs.

It is of particular interest to examine \reftab{results-1}, which displays the performance of the proposed method across different predicted future steps. Notably, it can be observed that as time progresses, the uncertainty of the future increases, leading to larger errors in the predictions. 

Furthermore, \reftab{results-1} presents the results of comparing the depth forecasting of the proposed method with prior works. As expected, the proposed method outperforms the prior work, with a significant gap of $\Delta AbsRel = 7.3\%$. This finding is further substantiated by the qualitative observations portrayed in \reffig{qualitative_perf}, where the generated output depth maps produced by our proposed method demonstrate superior precision and intricate details when compared to prior approaches. The forecasted depth at the different time steps is even comparable to other methods that does depth inference and have access to the target frame image. $t=1$ it is better than~\cite{Ranjan2019,Yin2018,yang2018unsupervised,Zhou2017}. and $t=5$ is better than~\cite{yang2018unsupervised,Zhou2017}. This demonstrates that the enforcing the spatio-temporal consistency results in an accurate depth sequence.

\subsection{Handling dynamic objects}
\label{sec:dynamic-obj}

In the interest of conducting a thorough analysis of the proposed method, it is important to consider its ability to handle dynamic objects in the scene. One limitation of a previous approach (\cite{Watson2021}) was its inability to handle such objects, and thus we conducted an experiment to address this issue.

The proposed model will be evaluated against~\cite{Watson2021,boulahbal2022instance} using the methodology introduced in~\cite{boulahbal2022instance}. Furthermore, a variant of the proposed method is employed which leverages stereo images during training. The pose network is completely discarded, and the extrinsic parameters of the stereo pair are used to warp the one view into the other. This variant model operates under the assumption of a rigid scene, thereby avoiding any issues related to warping dynamic objects.


The results, presented in~\reftab{stdepth_staticVSdynamic}, show that the proposed method outperforms the ManyDepth baseline on dynamic objects with a significant improvement of $\Delta AbsRel=13.0\%$, and it has a comparable result when the unbiased per category mean is used. Although the proposed variant with stereo images during evaluation performs almost equally well for the static scenes, most of the improvement is observed on the dynamic objects. These findings suggest that although the proposed model is not explicitly trained on dynamic objects, it is able to learn their dynamics implicitly. One possible explanation for this is that the as model utilizes multi-scale attention, which allows it to capture the motion of dynamic objects even without being specifically supervised to do so. This highlights the effectiveness of attention mechanisms in capturing spatio-temporal dependencies and modeling the dynamics of the scene. Overall, these experiments highlight the importance of handling the dynamic objects in depth inference and forecasting.

\subsection{Depth inference generalization}
\label{sec:generalization}

In order to compare the proposed architecture with other methods that perform single depth-image inference, it is proposed to train the model only for this (output only depth at $D_t$). The comparison is only made with respect to methods that leverage multi-fame input for the depth network. Prior methods~\cite{Watson2021,guizilini2022multi} perform a plane sweep operation to compute a cost volume. The plane sweep algorithm explicitly uses the pose of the scene and requires the camera intrinsic parameters. The proposed depth network model, on the other hand, performs the matching implicitly using the transformers and does not depend on any other network.

\begin{figure}
    \centering
    \includegraphics[width=\textwidth]{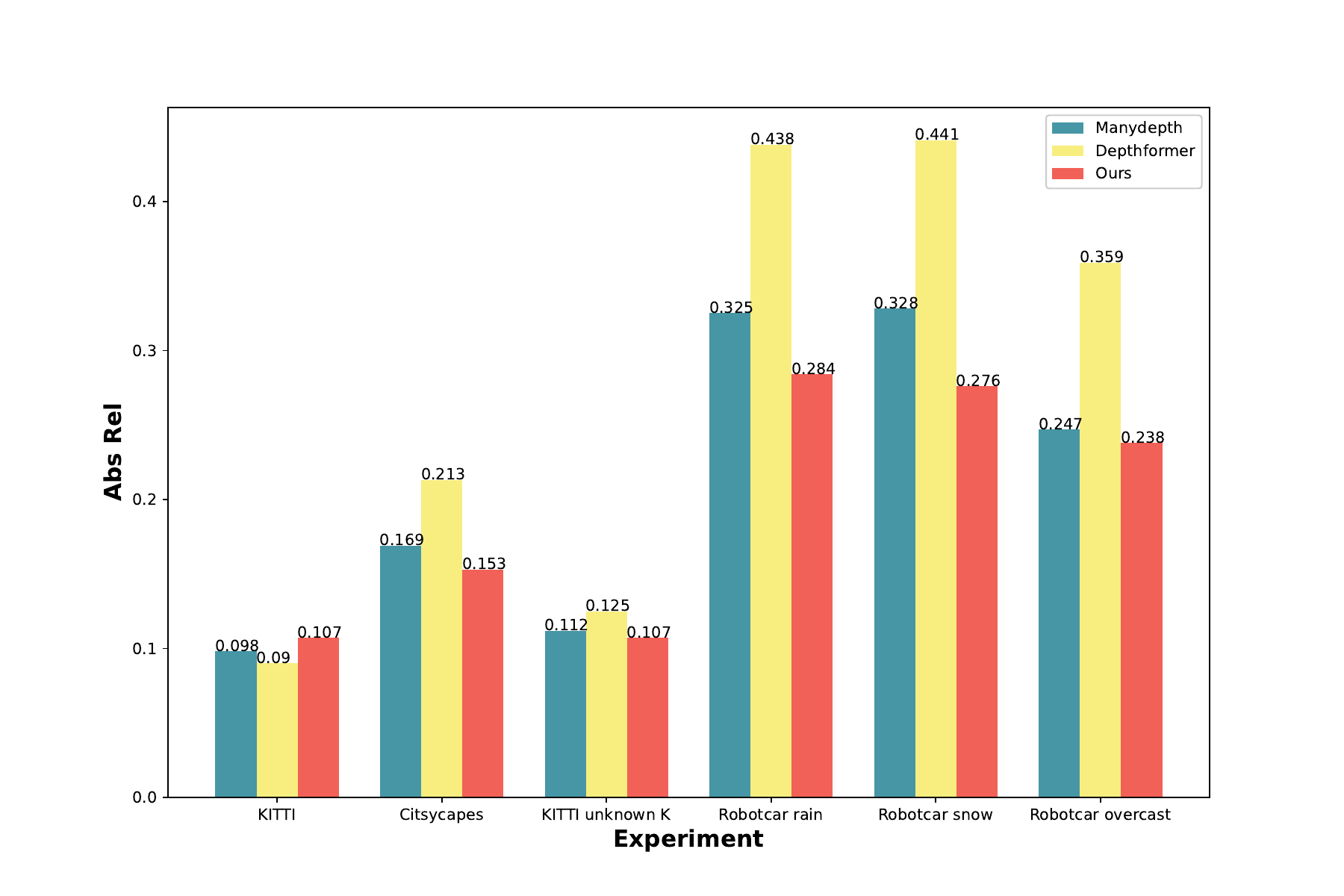}
    \caption{Depth inference generalization study. The proposed architecture is compared to ManyDepth and DeptFormer on different generalization scenarios: Domain gap evaluation on Cityscapes, sensitivity to camera parameters, and weather perturbations on the Robotcar dataset. As shown, the proposed method outperforms the baselines in all three generalization settings, suggesting its ability to generalize well to different scenarios.}
    \label{fig:generalization_study}
\end{figure}

\reftab{results-1} shows the comparison results of the proposed method with ManyDepth~\cite{Watson2021} on KITTI benchmark~\cite{Geiger2012CVPR}. As expected, the models that explicitly use the pose information have better performance on the KITTI benchmark for depth inference. These assessments, however, do not evaluate the ability of the networks to generalize to new scenes. Therefore, a generalization study was performed to better assess the models in this respect:
\begin{itemize}
    \item \textbf{Testing the domain gap:} The models pre-trained on the KITTI dataset are directly evaluated on the Cityscapes~\cite{cordts2016cityscapes} dataset without retraining.  
    \item \textbf{Testing the sensibility to the camera parameters:} The focal length of the camera is replaced with a focal length $f=1$ and the optical center is chosen as $(\frac{W}{2},\frac{H}{2})$. This evaluation is performed on the KITTI dataset. 
    \item \textbf{Testing weather perturbation:} The evaluation is done on $3$ sequences of the Robotcar dataset: overcast, snow and rain sequences.
\end{itemize}
As observed in~\reffig{generalization_study} the proposed method outperforms the baselines for these generalization settings. This suggests that while the baselines are able to perform better on the KITTI dataset, they do generalize better in other settings. This could be explained by the fact that the generalization of these methods depends on both the generalization of the pose and depth network. The proposed model, on the other hand, performs matching implicitly using the attention of transformers and does not depend on any other network, which makes it less sensitive to variations in pose and camera parameters. 

\subsection{Ablation study}
\begin{figure}
    \centering
    \includegraphics[width=0.7\textwidth]{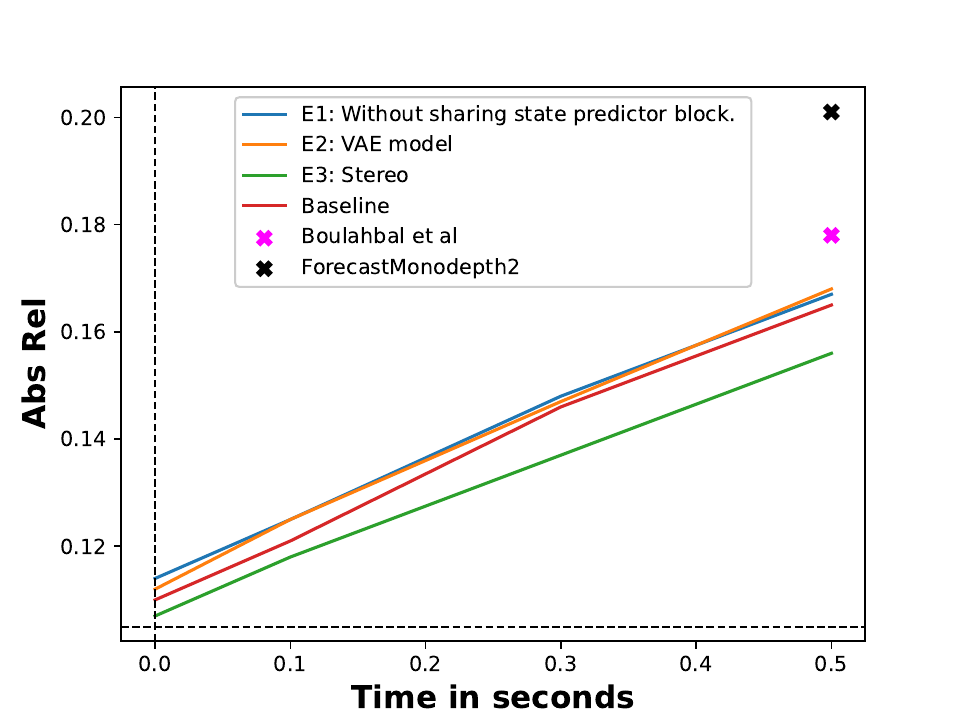}
    \caption{Ablation studies for improving depth forecasting performance. The \textit{Abs~Rel} performance of various evaluated models is shown in the figure. (i) E1, tests the model without sharing the state predictor block. (ii) E2, involves the use of a VAE model to output multi-hypothesis future depth (iii) E3, assess the model with the stereo pose.}
    \label{fig:ablations-chap6}
\end{figure}

Several ablations were performed in an effort to improve the performance of the proposed model. Figure~\ref{fig:ablations-chap6} displays the \textit{Abs~Rel} performance of the various evaluated models. Specifically, the following ablation studies were carried out:
\begin{itemize}
    \item E1: Tests the model without sharing the state predictor block. As observed, sharing the state predictor helps the model to output a better depth as multiple passes helps the network to generalize better.
    
    \item E2: This experiment involved the use of a VAE model, where the latent variables of the state predictor block were assumed to follow a Gaussian distribution. The aim of this experiment was to output a multi-hypothesis future depth. However, the first observation we made was that the model collapsed to a single modality and failed to output multiple hypotheses. As the decoder was perturbed with the Gaussian distribution, the output is less accurate with respect to the baseline.

    \item E3: Aim to assess the model with dynamic objects. More details are provided in~\refsec{dynamic-obj}

\end{itemize}

These experiments demonstrate that the proposed method holds several advantages: providing a spatial-temporal consistent depth sequence that represents present and future depth, superior depth forecasting compared to the prior work, and better generalization for depth inference.   

\label{sec:res-gen}

\section{Discussion}
\label{sec:new-conclusion}
 In conclusion, this chapter presents a novel self-supervised model that predicts a sequence of future frames from video-input using a spatial-temporal attention (ST) network. The proposed model outperforms existing baselines on the KITTI benchmark for depth forecasting and achieves highly accurate and robust depth inference results. The novelty of the proposed model lies in its use of a transformer-based multi-frame architecture that implicitly learns the geometry and motion of the scene, while also leveraging prior scene knowledge such as object shape and texture. Furthermore, the proposed model enforces spatio-temporal consistency across a sequence of output frames rather than a single output frame, resulting in more accurate and robust depth sequence forecasting. Several ablation studies were conducted to assess the effectiveness of the proposed techniques. The proposed model provides a significant contribution to the field of depth prediction, and holds great promise for a wide range of applications in computer vision. Future research could explore the generation of accurate multi-hypotheses future depth, building upon the promising results presented in this chapter. 
\chapter*{}

\chapter{Conclusion}
\label{ch:ch7}
\section{Summary of the thesis}
In this thesis, I have thoroughly examined the potential of self-supervised approaches for depth prediction and demonstrated their ability to provide a rich representation of a scene, enabling a more comprehensive understanding of motion and geometry. The task of predicting the future depth of a scene is undoubtedly challenging, yet it has significant implications for intelligent systems, particularly in the field of autonomous driving (AD) and advanced driver assistance systems (ADAS). The research presented in this thesis addressed the challenges of depth prediction using self-supervised learning techniques. Various scenarios were explored:

\begin{itemize}

    \item In Chapter~\ref{ch:ch3}, the generalization of deep learning models was explored. This was done through domain adaptation methods, by utilizing generative adversarial networks, specifically conditional GANs for style transfer. During this exploration, a fundamental limitation of cGANs was revealed, superficially their lack of complete conditionality. To address this issue, the chapter presents an innovative solution referred to as the “\ac method”. The main objective of the \ac method is to enhance conditional GANs and empower them with full conditionality.

    \item Chapter~\ref{ch:ch4} explored image-to-depth map inference and extended the classical methods with dynamic objects. We have presented a solution for the static-scene assumption of the classical SFM model, using a novel transformer-based method that outputs a pose for each dynamic object. 

    \item Chapter~\ref{ch:ch5} explored video-to-depth mapping. This was the first attempt to forecast the future depth using self-supervision. The proposed model used a sequence of the past and present frames, and the model outputs a depth map that represents the future depth at step $k$. A novel transformer-based architecture was proposed to aggregate the temporal information, this enabled the network to learn a rich spatio-temporal representation.

    \item Chapter~\ref{ch:ch6} presented video-to-video depth. This model takes a sequence of images of past and present images and outputs a sequence of the present and future depth maps. This method addressed the limitations of the previous methods and extended the forecasting into a sequence of future depth. We have presented our self-supervised model that simultaneously predicts a sequence of future frames from video input with a novel spatial-temporal attention (ST) network.
\end{itemize}

Accurately predicting future depth can help these systems better anticipate and react to changes in their environment, which is crucial for their safe and effective operation. The applications of self-supervised depth prediction extend beyond AD and ADAS, as this method offers an efficient way to enable good understanding of videos. Given the promising results of self-supervised depth prediction, it is worth considering the possibility of applying this technique at scale to create vision models akin to the popular GPT-4 language model. Such models would have broad applications, ranging from autonomous systems to robotics, and beyond.

\section{Perspective and future work}
To build upon the findings of this thesis, future research could explore the following areas:

\begin{itemize}
    \item \textbf{Predicting multiple plausible future depth:\\} %
    Depth prediction self-supervision uses a differentiable warping and an image reconstruction as pretext task. This warping assume is that the environment is deterministic and that there is only one possible future. However, the future is stochastic by nature. The uncertainty of motion grows with time and there exists a multiple possible outcomes. In \reffig{crossing} the pedestrians could decide to cross the road or not. While the past context could provide a hint of the future actions of the pedestrian, his actions are not deterministic. Therefore, generating multiple hypotheses for the future is crucial for path planning and for safety applications. In Chapter~\ref{ch:ch6}, We tried to use a VAE to model to generate multiple future scenarios, but we observed that the model had collapsed into a single mode. We suspect that the warping function and the image reconstruction with a deterministic video (there is only one scenario observed) makes the training collapse into the single mode (\ie the most likely mode based on the context frames). One solution is to train on a dataset with multiple hypothesis. We believe that this avenue of research will have a high impact on safety applications. 
    \begin{figure}
        \centering
        \includegraphics[width=0.75\textwidth]{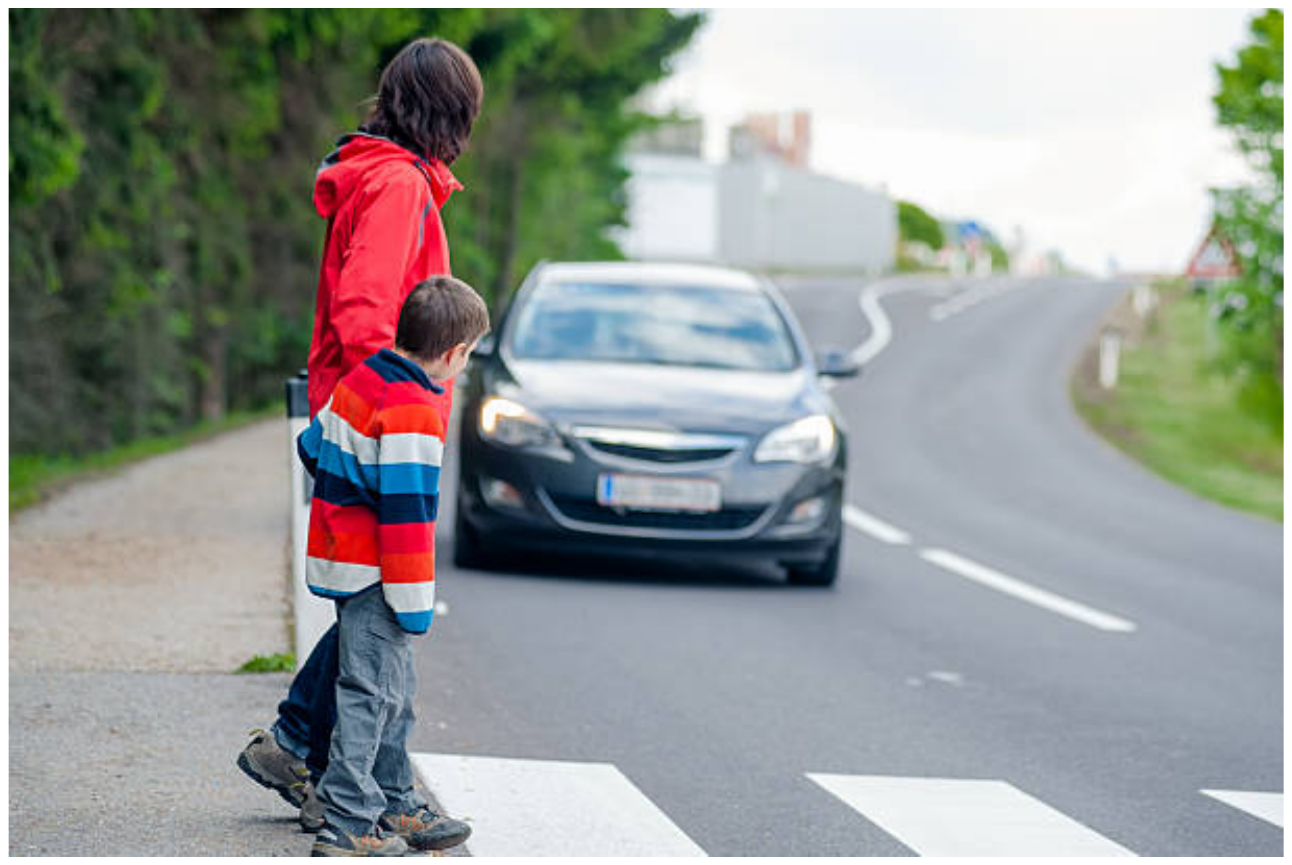}
        \caption{An example of the future multi-hypothesis. The pedestrians may or may not cross the street, the situation is uncertain.}
        \label{fig:crossing}
    \end{figure}

    \item \textbf{Scaling the training for large-scale datasets:\\} 
    As we mentioned in Chapter~\ref{ch:ch3} the advantage of the self-supervision is the ability to train on large-scale dataset as only videos are required. The current results show that even for a small and limited dataset such as KITTI the model is able to obtain accurate depth with a closing gap with respect to the supervised methods. Training with large-scale datasets is challenging for data collection, storage, and especially training. However, this approach could enable the model to not only achieve outstanding performance and generalization, but also to attain a genuine scene understanding akin to GPT models. As such, this research direction has the potential to make substantial contributions to the field of computer vision. 

    \item \textbf{Domain generalization:\\} 
    While training on large scale-datasets could be a way to go forward with improving the generalization, it is possible to improve the generalization with other techniques like domain adaptation and style transfer. If we have prior knowledge on what domains the model will encounter, it is possible to adapt the model directly to these domains. However, generative techniques are susceptible to hallucination. One possible solution is to use the acontrario cGAN~\cite{Boulahbal2021BMVC} to help the generative model to be consistent with the conditioning input and the depth model to generalize better. The pursuit of improved generalization in depth models is an ongoing endeavor with potential to improve downstream safety applications. 

    \item \textbf{Improving depth with semi-supervised learning:\\} 
    Although this thesis has primarily focused on self-supervised methods, we believe that semi-supervised methods hold significant potential for enhancing performance and generalization in computer vision applications. The utilization of supervised labels provides a well-constrained signal for supervising the network and achieving improved accuracy in the predictions. Furthermore, self-supervision aids the network in achieving better generalization by enabling training on large-scale datasets that may feature significant domain shifts. The integration of semi-supervised methods into the training pipeline presents a promising research direction with the potential to advance the field of computer vision. However, to fully realize the potential of semi-supervised methods, additional research is required to investigate the optimal strategies for combining supervised and self-supervised methods.


\end{itemize}

In conclusion, this thesis has demonstrated the potential of self-supervised depth prediction as a powerful tool for enabling a more comprehensive understanding of scenes and their dynamics. The implications of this technique for intelligent systems, particularly in the field of autonomous driving and advanced driver assistance systems, are significant and promising.

\chapter*{}

\appendix
\chapter{Computer vision basics}
\label{ch:apdx1}
Computer vision deals with how computers can understand and interpret images and videos. It involves the development of algorithms that can analyze and understand the environment represented as a set of \textbf{image}, Understanding this environment involves recognizing the entities and their \textbf{motion} and reasoning about their interactions. this enables to perform useful tasks such as object recognition, image classification, scene understanding, object tracking, 3D reconstruction and more. In the following section, we begin to define the rigid-body transformation and the process of acquiring images and the geometry of multiple-views.

\section{Rigid-body transformation}
\begin{figure}
    \centering
    \includegraphics[width=0.5\textwidth]{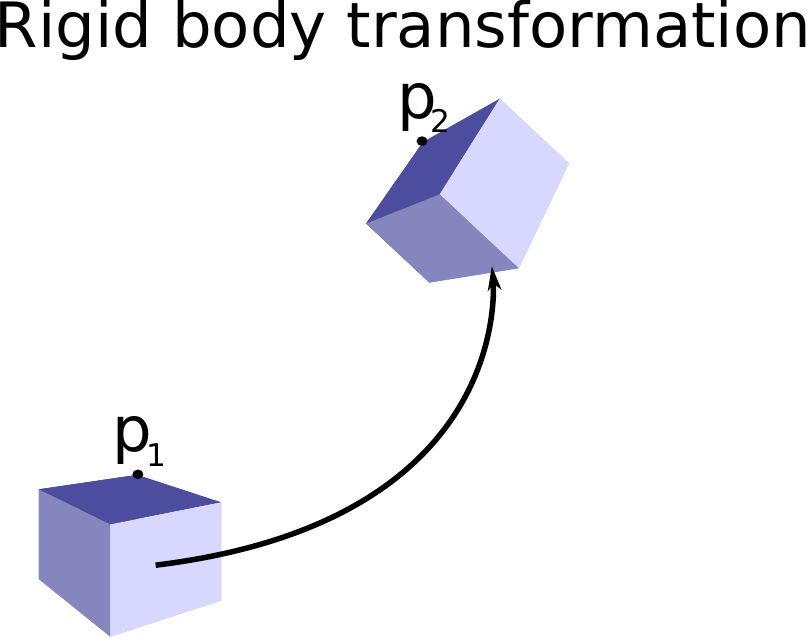}
    \caption{An example of a rigid-body transformation}
    \label{fig:rigid_body}
\end{figure}

In order to describe the motion of an object, in principle, the trajectory of all points of that objects should be specified. However, as this object do not have any deformation or change in its shape, specifying the motion of one point is sufficient. This known as rigid-body transformation. This type of transformation can include rotations, translations, and combinations of both. In the context of computer vision, rigid-body transformation is often used to describe the movement of objects within an image or video, and can be used to track the motion of those objects over time. It could be defined formally as : \\ 
\textbf{Rigid-body transformation: }A map $g : \mathbb{R}^3\longrightarrow\mathbb{R}^3$ is a rigid-body transformation if it preserves the norm and the cross product of  any two vectors : 
\begin{enumerate}
    \item norm : $||g(\mbf{v})|| = ||\mbf{v}|| , \mbf{v} \in \mathbb{R}$. 
    \item cross product: $g(\mbf{v}) \times g(\mbf{u}) = g(\mbf{v} \times \mbf{u}) \mbf{v},\mbf{u} \in \mathbb{R}^3$.
\end{enumerate}
Rigid-body transformation can include rotations, translations, and combinations of both. For example, A point in the world frame at instance  $p_{1}^w$ can be transformed with a rotation and translation as shown in the \reffig{rigid_body}. The equation that relates the two points can be expressed as: 
\begin{align}
    \label{eq:rigid-body}
    p_{2}^w =^{2}R_{1} p_{1}^w + ^{2}T_{1}
\end{align}
where $^{2}R_{1}$ is the rotation matrix and  $^{2}T_{1}$ is the translation matrix.

\subsection{Rotation matrix representation}
The rotation matrix is $3\times3$. However, a rotation have only $3$-DOF. Therefore, this 9 parameters matrix could be expressed using only 3 parameters. There exists several minimal parametrization of the rotation matrix such as Euler angles, quaternions and axis-angle parametrization. 

\begin{figure}
        \centering
        \includegraphics[width=0.25\textwidth]{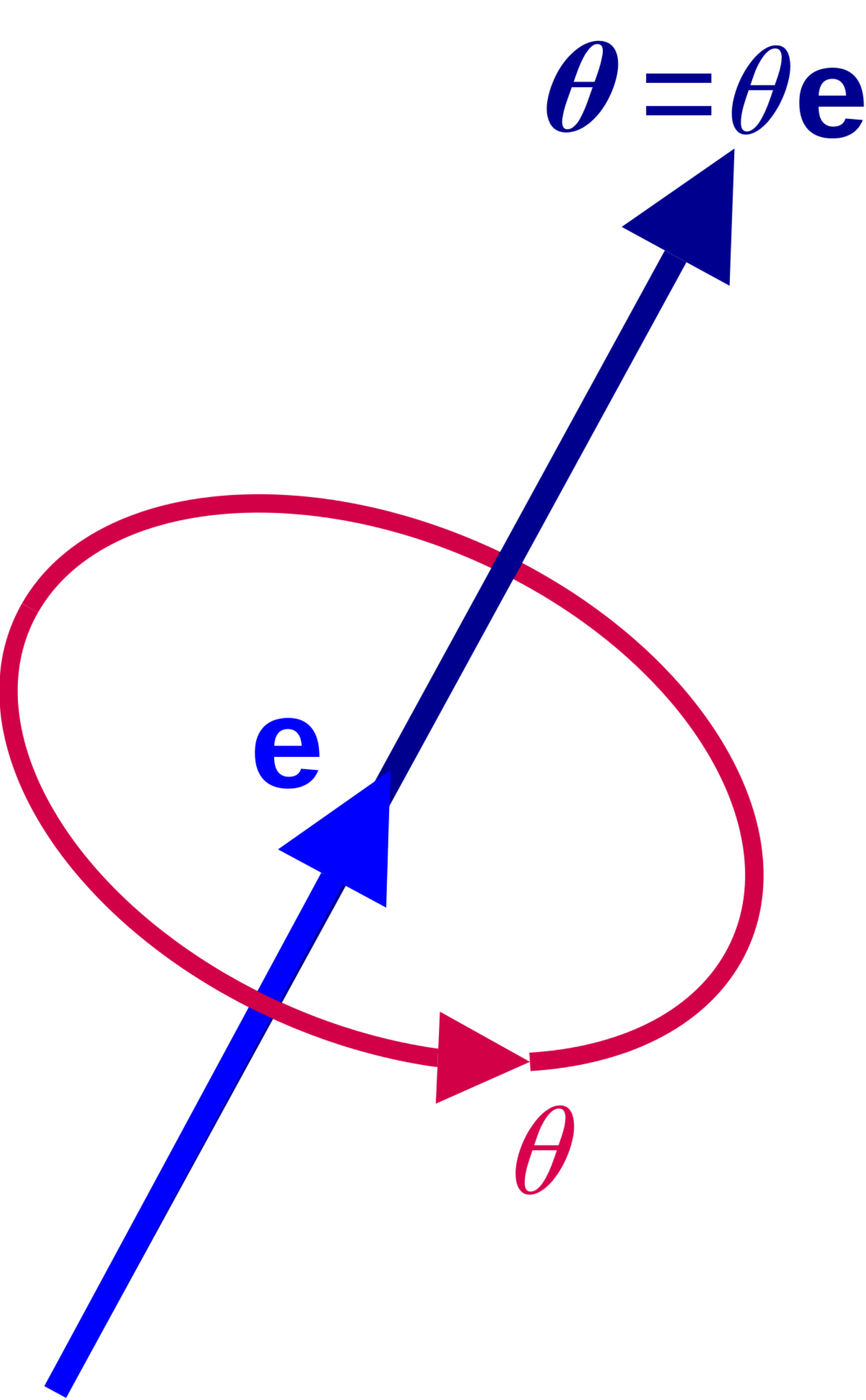}
        \caption{An example of the axis-angle rotation convention. The rotation is parameterized by the vector $\bm{\theta}= \theta \mbf{e}$ where the vector $\mbf{e}$ gives the direction and $\theta$ is a scalar that gives the angle.}
        \label{fig:axis-angle}
    \end{figure}

\begin{itemize}
    \item \textbf{Euler angles :} a commonly used method for rotation representation, where a rotation is decomposed into three consecutive rotations around different axes, as shown in \reffig{euler}. The simplicity and ease of interpretation of Euler angles make them a popular choice in many applications. However, it is important to note that Euler angles are not unique and can produce the same rotation with different parametrization depending on the order of the rotations.
\begin{figure}
    \centering
    \includegraphics[width=\textwidth]{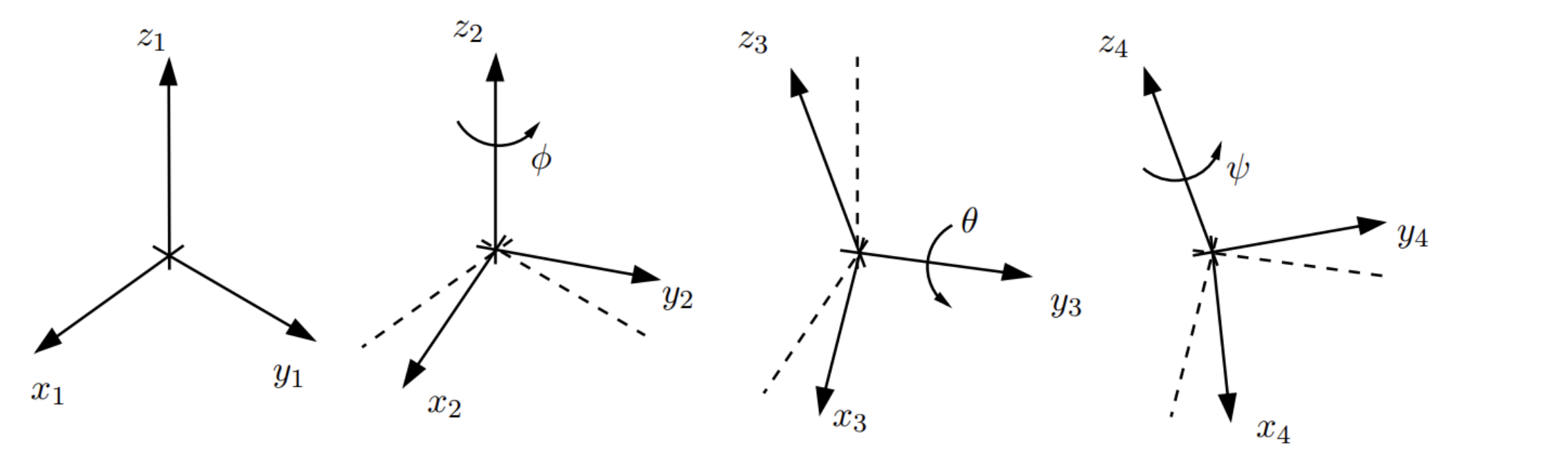}
    \caption{An example of an Euler rotation representation $Z_1Y_2Z_3$. Consider a Cartesian coordinate.  In order to define Euler angles, three canonical rotations are applied. First rotate around the z-axis by $\phi$, then around the new y-axis by $\bm{\theta}$, and finally around the new z-axis by $\psi$.}
    \label{fig:euler}
\end{figure}

    \item \textbf{Axis-angle representation} It represents the rotations with a single angle and axis of rotation. The axis of rotation is defined as a unit vector in 3D space, and the angle represents the magnitude of rotation about this axis. \reffig{axis-angle} represents an example of axis-angle rotation convention. The rotation is parameterized by the vector $\bm{\theta}= \theta \mbf{e}$ where the vector $\mbf{e}$ gives the direction and $\theta$ is a scalar that gives the angle.

\end{itemize}

Axis-angle representation of rotation is generally considered to be better than Euler angles in several ways:
\begin{itemize}
    \item \textbf{Unique representation:} Axis-angle provides a unique representation for a rotation, while Euler angles can lead to singularities and result in multiple solutions for a single rotation.
    \item \textbf{Avoiding Gimbals lock:} Euler angles can suffer from Gimbals lock, a phenomenon where two of the rotational degrees of freedom become locked to each other, causing the rotation to become ambiguous. Axis-angle does not suffer from Gimbals lock.

\end{itemize}

\subsection{Homogeneous representation}
Since \refeq{rigid-body} combine an addition and multiplication, it is possible to represent that equation with a single matrix multiplication. This is achieved by introducing a homogeneous vector, which is a 4-dimensional vector that includes an additional element of 1 at the end $\mbf{p} = (x,y,z,1)$. 

The homogeneous transformation is represented as a $4\times4$ matrix, where the first $3\times3$ elements represents the rotation and the last column represents the translation. This matrix can be used to transform a $3D$ point in one coordinate system to another coordinate system by matrix multiplication. The advantage of using a homogeneous transformation is that it allows for a compact and efficient way to represent both rotation and translation, as well as combining multiple transformations into a single matrix multiplication.

\begin{align}
    \mbf{p}_1 = \mbf{T} \mbf{p}_2 = \begin{bmatrix}
    \mbf{R} & \mbf{t} \\
    \mbf{0}_{3}& 1
    \end{bmatrix}\mbf{p}_2
\label{eq:homogenous}
\end{align}
$\mbf{T}$ is the homogeneous matrix that transforms the homogeneous point $p_1$ into the point $\mbf{p}_2$ 

It is important to note that the pose of an object refers to its position and orientation in space. It is often represented by a combination of translation $(x, y, z)$ and rotation (roll, pitch, yaw) parameters, or as a $4\times4$ transformation matrix that describes the same information.
The rigid-body transformation, on the other hand, is a mathematical operation that describes how points in one coordinate system can be transformed to another coordinate system, while preserving the distances and angles between points. It is often used to describe the relationship between two different coordinate systems. Therefore, the pose of an object or a camera describes its location and orientation in a particular coordinate system, while the rigid-body transformation describes how to transform points between two different coordinate systems.

\section{Pinhole camera model}
A camera is a device that captures images by detecting and measuring the intensity of electromagnetic radiation, such as light. It consists of a lens and a light sensor. The lens is used to control the direction and intensity of the incoming light, while the light sensor measures the amount of light that falls on it and converts it into an electrical signal. This measurement, known as irradiance, is a measure of the power per unit area of the light incident on the sensor, and it is typically expressed in watts per square meter $(W/m^2)$. There are several types of cameras that use different approaches to capture and process images, including pinhole cameras, fish-eye cameras, and event-based cameras. Pinhole cameras are the most common and widely used type of camera, and they are found in a wide range of applications, including smartphones, webcams and even cars. These cameras are cheap and well-documented.

An image is a representation of the visual perception of the world. This representation encodes the world in a $2D$ array of pixels. Each of these pixels stores the color intensity for that location of the image. More Formally, an image is mapping $I$ that assign to a location $(x,y)$ a positive value : 
\begin{align}
    \mbf{I}(x,y) : \mathbb{R}^2 \Longrightarrow \mathbb{R}^{3+}
\end{align}
 An RGB image stores the color intensity of red, green and blue. For a digital image, the value of the intensity $\mbf{I}(x,y)$ is discretized in 8 or 16 bit representation. The domain of $(x,y)$ is also discretized $(x,y) \in ((0,W) \in \mathcal{N}, (0,H) \in \mathcal{N})$. Where $W$ is the width and $H$ is the height of the image.

\subsubsection{Geometric model for pinhole camera image formation}
\begin{figure}
    \centering
    \includegraphics[width=0.7\textwidth]{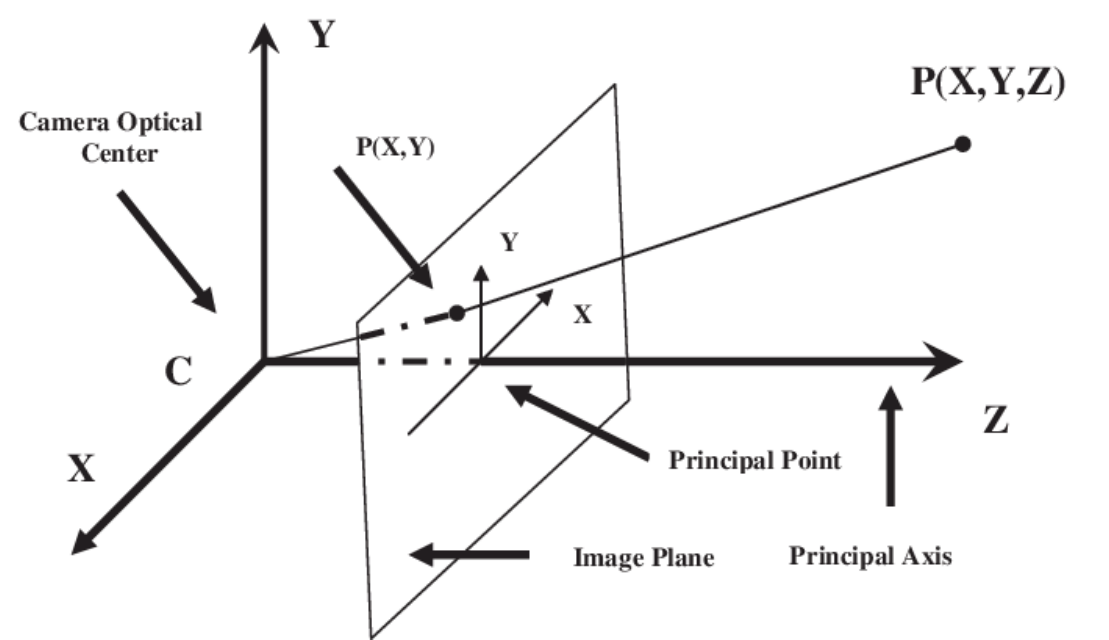}
    \caption{Fronto-prallel pinhole camera model. The point P in the world frame is projected to the image coordinates. This process is defined using the extrinsic matrix that relates the world frame and camera frame, The projection to the image frame using the focal length and The image frame transformation using the optical center.}
    \label{fig:pinhole-model}
\end{figure}

In order to accurately model and predict the behavior of a pinhole camera, it is necessary to define a mathematical model that describes the relationship between the input light and the output image. This model takes into account the properties of the lens and sensor. By understanding and applying this mathematical model, it is possible to establish a correspondence between the points in the 3D space and their project 2D image. The pinhole camera model is shown in \reffig{pinhole-model}

One mathematical model that can describe the image formation includes 3 transformations : 

\begin{enumerate}
    \item \textbf{Projection into camera frame:} it transforms the point from the world frame into the camera frame : if $P_c$ have the coordinates $P_w = [X_w,Y_w,Z_w]$ we could obtain the coordinates of this point relative to the camera frame given by the rigid body transformation :
    \begin{align}
        \mbf{P_c} = ^c\mbf{T}_w \mbf{P}_w =   \begin{bmatrix}
    \mbf{R} & \mbf{t} \\
    \mbf{0}_{3}& 1
    \end{bmatrix} \mbf{P}_w 
    \end{align}

     \item \textbf{Projection into the image coordinate:} Using the fronto-parallel pinhole camera model, the 3D point $P_c$ is projected to the image frame coordinates :  
    \begin{align}
        \mbf{p} &= \begin{bmatrix}
            x \\y
        \end{bmatrix} = \frac{f}{Z} \begin{bmatrix}
            X \\Y
        \end{bmatrix}  \\ 
        Z \mbf{p} &= \begin{bmatrix}
            f  & 0  & 0 &0 \\
            0  & f  & 0 &0  \\
            0  & 0  & 1 &0 
        \end{bmatrix} \begin{bmatrix}
            X \\ Y \\ Z \\ 1
        \end{bmatrix}  = \mathbf{K}_f \begin{bmatrix}
            X \\ Y \\ Z \\ 1
        \end{bmatrix} 
    \end{align}
    $f$ represents the focal length of the camera.
    \item \textbf{Coordinates transformation from normalized coordinates to pixel coordinates:} first converting from metric to pixels and converting the origin to be the top-left of the image. This transformation can be expressed as :
    \begin{align}
       Z \begin{bmatrix}
       x' \\ y ' \\ 1
       \end{bmatrix} =  \begin{bmatrix}
            s_x  & s_{\bm{\theta}}  & o_x \\
            0  & s_{y}   & o_x \\
            0  & 0  & 1 
        \end{bmatrix} \begin{bmatrix}
            x \\ y \\ 1
        \end{bmatrix} 
    \end{align}
    $s_x$ and $s_y$ converts the metric coordinates into the pixel coordinates. $s_{\bm\theta}$ is the \textit{skew} factor usually close to zero for digital cameras. $o_x$ and $o_y$ are the coordinates of the optical center.
\end{enumerate}

In summary, the pinhole camera model can be defined as follows: 

\begin{align}
    Z \begin{bmatrix}
       x' \\ y ' \\ 1 \end{bmatrix} &= \begin{bmatrix}
            f s_x  & f s_{\bm{\theta}}  & o_x \\
            0  & f s_{y}   & o_x \\
            0  & 0  & 1 
        \end{bmatrix} \begin{bmatrix}
            1  & 0  & 0 &0 \\
            0  & 1  & 0 &0  \\
            0  & 0  & 1 &0 
        \end{bmatrix} \begin{bmatrix}
    \mbf{R} & \mbf{t} \\
    \mbf{0}_{3}& 1
    \end{bmatrix} \begin{bmatrix}
            X_w \\ Y_w \\ Z_w \\ 1
        \end{bmatrix} \\
    Z \mbf{p}' &= \mbf{K} \: \bm{\Pi} \: \mbf{T} \: \mbf{P}_w
\end{align}
Where $\mbf{K}$ is the intrinsic camera matrix. $\bm{\Pi} $ is the projection matrix, and $\mbf{T}$ is the extrinsic parameters. 

\subsection{Epipolar geometry}
Epipolar geometry is a mathematical concept that describes the relationship between two views of the same scene. Consider two images $(\mbf{I}_1$, $\mbf{I}_2)$ of the same scene from a different view. If a point $\mbf{X}$ have coordinates $\mbf{x}_1$ and $\mbf{x}_2$ relative to the frames of each camera and $^1T_2$ is the pose of the second camera with respect to the first then: 
\begin{align}
    \label{eq:pose}
    \mbf{x}_1 = ^1T_2 \mbf{x}_2
\end{align}
\textbf{Epipolar constraint:} The epipolar constraint that relates the two images $x_1$ and $x_2$ is defined as follows: 
    $$\mbf{x}_2^T \hat{\mbf{T}}\mbf{R} \mbf{x}_1 = 0$$
The matrix $$\mbf{E} = \hat{\mbf{T}}\mbf{R} $$ is called the essential matrix. It encodes the relative pose of the two cameras. 
\begin{figure}
    \centering
    \includegraphics[width=\textwidth]{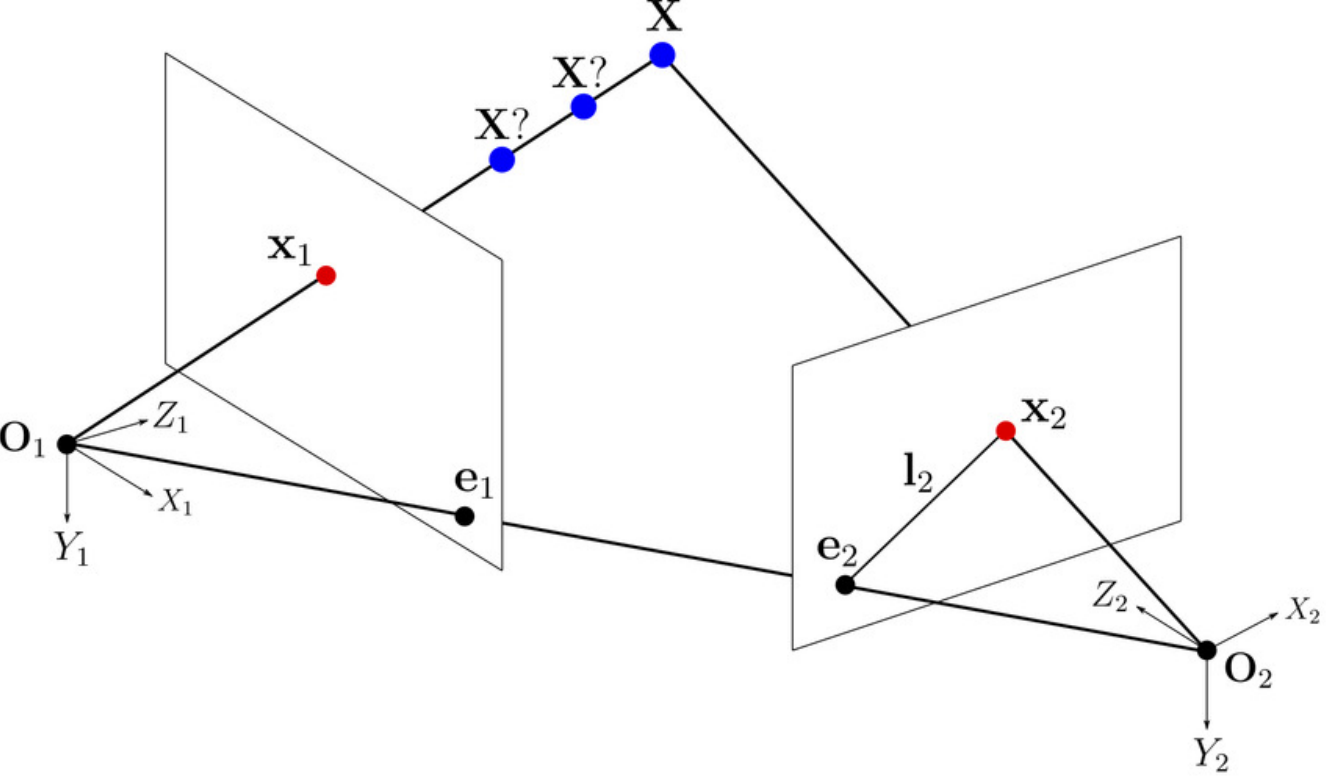}
    \caption{Illustration of the epipolar geometry model of two cameras with optical centers $O_1$ and $O_2$. The point $X$ is projected as $x_1$ for the first camera and $x_2$ for the second camera. The epipoles are defined at the intersection of the image planes and the plane $(O_1,O_2,X)$. The projection of the line $(O,x_1)$ on the other camera is called the epipolar line. The corresponding point $x_2$ is situated at that line.}
    \label{fig:epipolar}
\end{figure}
\reffig{epipolar} show the projection, the point $\mbf{X}$ in the view. The intersection of the line $(o_1, o_2)$ is called \textbf{epipoles} denoted by $e_1, e_2$. The lines $l_1$ and $l_2$ are called the \textbf{epipolar lines} which are the intersection of the plane $(O_1,O_2,X)$ with the two image plane.

The epipolar geometry is a powerful tool for establishing correspondences between stereo pairs and resolving the scale ambiguity in depth prediction with deep learning. It is possible to project a IR pattern to replace one of the images and to establish the correspondence based on the captured pattern and the ground truth pattern, this is the principle of an active depth sensor.

\subsection{Classical methods for depth prediction}
In computer vision, depth perception has been studied extensively as a means of understanding and interpreting visual scenes. We could distinguish two approaches for depth prediction: monocular methods and multi-view methods 
\subsubsection{Classical methods for monocular depth prediction}
The depth is recovered from the motion of the camera. The scene is captured from different view and by knowing the relative position of camera it possible to recover the depth up to a certain scale. This algorithm is known as Structure from motion (SFM). Here is a typical structure from motion algorithm : 

\begin{enumerate}
    \item Load a set of images and detect keypoints and extract the descriptors (such as SIFT or ORB) for each image.
    \item Find correspondences between the keypoints of different images using descriptor matching.
    \item Estimate the fundamental matrix for each pair of images with correspondences.
    \item Compute the essential matrix for each pair of images.
    \item Use the essential matrix to compute the camera pose for each image.
    \item Triangulate the 3D positions of the corresponding points using the camera pose for each image.
\end{enumerate}

\subsubsection{Classical methods for multi-view depth prediction}
One example to perform multi-view depth prediction is stereo matching. It is a computer vision technique used to estimate the 3D structure of a scene from two or more images taken from different viewpoints. It involves finding corresponding points between the images and using these correspondences to compute the depth of each point in the scene.

There are many algorithms for stereo matching, but one of the most popular is the block matching algorithm. This algorithm works by dividing each image into small blocks, and then comparing the blocks from one image to the blocks in the other image to find the best match. The difference in position between the matching blocks is used to estimate the depth of the points in the scene. Here is a typical stereo matching algorithm : 
\begin{enumerate}
    \item Load two images of the same scene taken from different viewpoints.
    \item Rectify the images.
    \item Set the search window size and block size.
    \item For each block in the left image:
            \begin{enumerate}
            \item Search for the best matching block in the right image within the search window.
            \item Calculate the difference in position between the matching blocks.
            \item Use the difference in position to estimate the depth of the points in the block.
            \end{enumerate}
\item Repeat the process for each block in the right image.
\end{enumerate}

These classical methods are limited in their ability to handle complex and varied scenes, and are prone to errors and ambiguities. With the advent of deep learning, it has become possible to learn more robust and effective features for depth perception from large amounts of data. Instead of relying on hand-crafted features that may not encode relevant information for depth prediction. Deep learning approaches learn from the data the optimal features for depth prediction. These methods have achieved significant progress and outperform classical methods.

\chapter{Acontrario conditional GAN}
\label{ch:apdx2}
Supplementary material is presented here as follows, Section~\ref{sec:mode-collapse} provides an additional evaluation of mode collapse for the depth prediction model. Section~\ref{sec:loss_extra} looks into the choice of weighting the different parts of the proposed loss function. Details are provided for reproducibility in Section~\ref{sec:implementation}. Finally, an analysis of the training procedure is provided in Section~\ref{sec:dynamics} to show that the training procedures did not encounter any degenerate situations.

\section{Mode collapse analysis}
\label{sec:mode-collapse}
Mode collapse is the setting in which the generator learns to map several inputs to the same output. A collapsing model is by construction unconditional. Only a few measures have been designed to explicitly evaluate this issue~\cite{richardson2018gans,wang2003multiscale,arora2018gans}. MS-SSIM~\cite{wang2004image,wang2003multiscale} measures a multi-scale structural similarity index and birthday paradox~\cite{arora2018gans} concerns the probability that, in a set of $n$ randomly chosen outputs, some pair of them will be duplicates. Another approach, NDB~\cite{richardson2018gans}, presents a simple method to evaluate generative models based on relative proportions of samples that fall into predetermined bins. 

\begin{figure}
    \centering
    \includegraphics[width=0.7\textwidth]{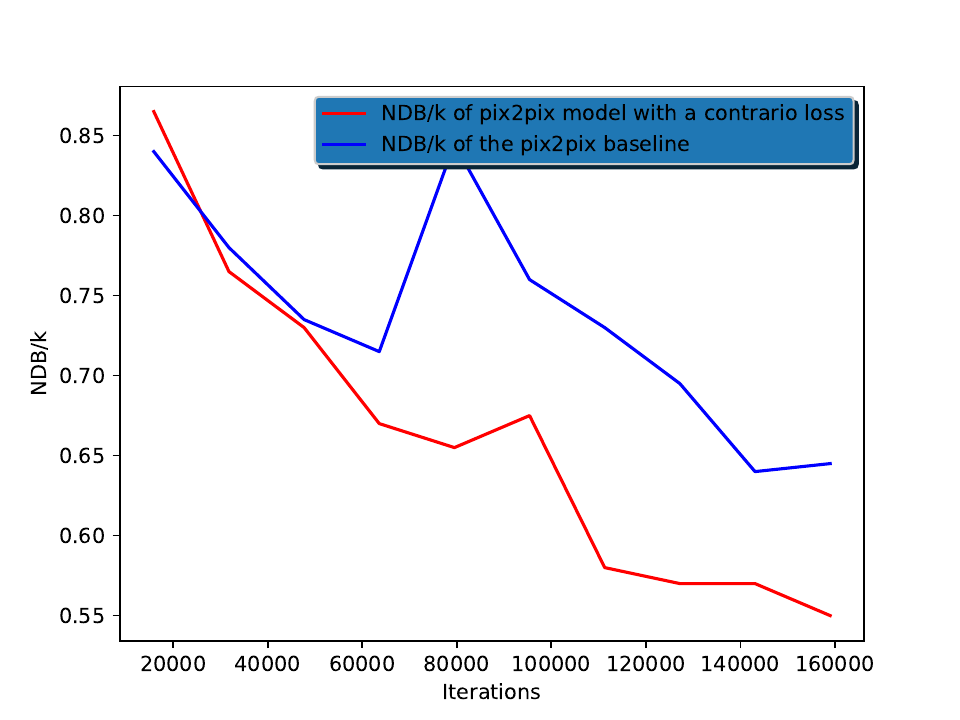}
    \caption{An analysis of mode collapse using the NDB criteria (lower values are better) throughout training on the NYU depthV2 dataset. It can be concluded from this evaluation that the proposed approach is much better at avoiding mode collapse due to the restricted search space of the generator.}
    \label{fig:collapse}
\end{figure}

The analysis provided in this section is an extension of the experiments done on depth prediction. Figure~\ref{fig:collapse} shows the evolution of the NDB measure over training iterations using the NDB score (the less, the better) for both pix2pix baseline and \ac cGAN models trained on the NYU Depth V2 training set~\cite{Silberman:ECCV12}. Out of the 12 trained models, the best model (in terms of RMSE $\log$) is chosen for the evaluation. For clustering and evaluating NDB, non overlapping patches of $64\times64$ are considered. At the end of the training the NDB/k ($k=100$) of the \ac cGAN is $0.550$ while the baseline achieves only $0.645$. This indicates that \ac model \textbf{generalizes better}. This is also observed qualitatively in Figure~\ref{fig:depth_qualitative}. Training with the counter examples helps the discriminator to model conditionality. Thus, the generator search space is restricted to only conditional space. The generator is penalized for non-conditionality even if the generation is realistic.

\section{Loss function analysis}
\begin{figure}
\centering    
    \includegraphics[width=0.7\textwidth]{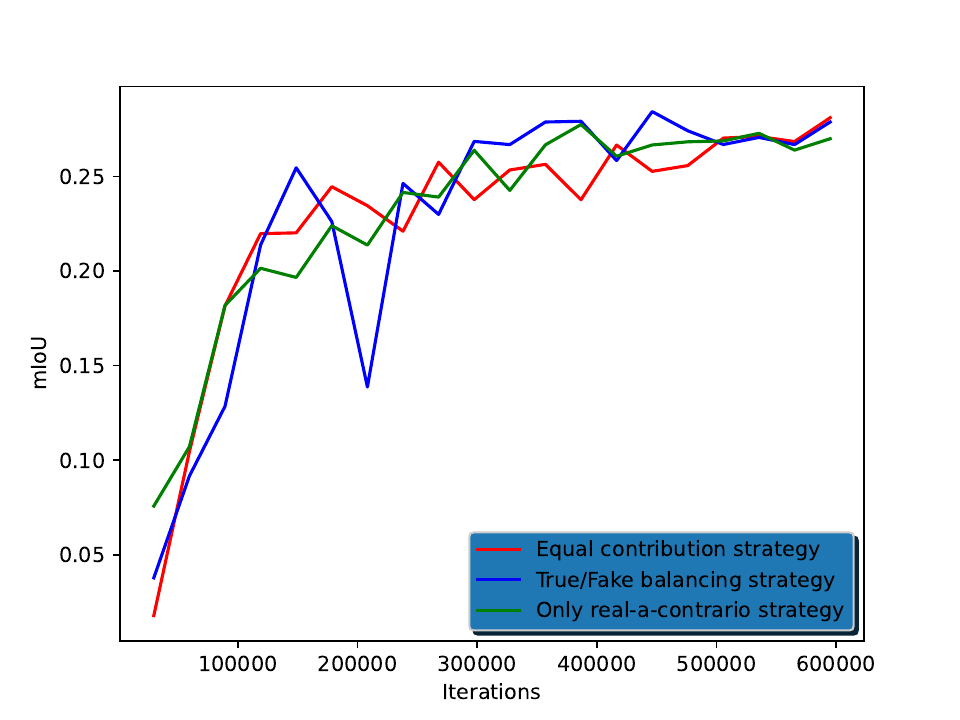}
    \caption{The mIoU evaluation for different choice of $\lambda_i$. The strategy 1 of giving equal contribution yield the best results. However, there is no major difference on the convergence or the performances at epoch 200 between the different strategies}
    \label{fig:lambda_choice}  
\end{figure}

An ablation study on Eq~\ref{eq:overall_loss} was performed. Each term that contributes to the adversarial loss is weighted by $\lambda_i$. Eq~\ref{eq:overall_loss} becomes: 
\begin{align}
    \mathcal{L}_{adv}= \min_{G} &\max_{D} \:  \Big[\lambda_1\mathbb{E}_{\mbf{x} \sim p(\mbf{x}) ,\mbf{y} \sim p(\mbf{y|x})}\big[log(D(\mbf{x},\mbf{y})]\big] + \lambda_2\mathbb{E}_{\mbf{x}\sim p(\mbf{x})}\big[log[1 - D(\mbf{x},G(\mbf{x}))]\big] \Big] +  \notag \\
    &\max_{D} \:\Big[ \lambda_3\mathbb{E}_{\mbf{\tilde x} \sim p(\mbf{\tilde x}) ,\mbf{y}\sim p(\mbf{y})}\big[log(1 - D(\mbf{\tilde x},\mbf{y}))\big]  + 
        \lambda_4\mathbb{E}_{\mbf{\tilde x} \sim p(\mbf{\tilde x}) ,\mbf{x} \sim p(\mbf{x})}\big[log(1 - D(\mbf{\tilde x},G(\mbf{x})))\big] \Big]
\end{align}
\label{sec:loss_extra}
Three strategies were considered for the weighting. The models were trained on the Cityscapes label-to-image dataset with the same settings described earlier (Section~\ref{sec:lb-to-im}). Figure~\ref{fig:lambda_choice} shows the mIoU for different \ac cGAN models trained with different choices  for $\lambda_i$. 
\begin{itemize}
\item \textbf{Strategy 1:} Equal contribution for each term : $\lambda_1 = \lambda_2 = \lambda_3 = \lambda_4$.

\item \textbf{Strategy 2:} Balancing the "fake" and "true" contributions. Since there are 3 data pairings classified as fake and only 1 real pair as true, equal balancing of true/fake gives: $\lambda_1 = 1 , \lambda_2 = \lambda_3 = \lambda_4 = 0.33$

\item \textbf{Strategy 3:} Testing the significance of both \ac error terms for fake and real images. In this case only 3 terms with real-a-contrario is tested : $\lambda_1 = \lambda_2 = \lambda_3 = 0.5, \lambda_4 = 0$.

\end{itemize}
In this simple test, Strategy 1 gives the best results. Strategy 2 seems less stable. Strategy 3 succeeds to learn conditionality, however, it may not capture conditionality for generated images during training. Each of these strategies succeed to model conditionality, however, Strategy 1 converges faster and yields a better final result in terms of mIOU. 

\section{Reproducibility}
\label{sec:implementation}

Various experiments were performed using different datasets and input-output modalities. Some extra detail is provided here for reproducibility purposes. In all the experiments using the pix2pix baseline, random jitter was applied by resizing the $256\times 256$ input images to $286\times 286$ and then randomly cropping back to size $256\times 256$. All networks were trained from scratch. Weights were initialized from a Gaussian distribution with mean $0$ and standard deviation $0.02$. The Adam optimizer was used with a learning rate of $0.0002$, and momentum parameters $\beta_1 = 0.5$, $\beta_2 = 0.999$. A linear decay is applied starting from epoch $100$, reaching $0$ at epoch $200$. Dropout is used during training. As in the original implementation~\cite{Isola2017ImagetoImageTW}, the discriminator is a PatchGan with a receptive field of $70\times 70$. Similarly pix2pixHD~\cite{Wang2018HighResolutionIS}, SPADE~\cite{Park2019SemanticIS} and CC-FPSE~\cite{Liu2019LearningTP} were trained with the same hyper-parameters as mentioned is their respective papers.  
For label-to-image, a U-Net256 with skip connections was used for the generator. A U-Net with $9$ ResNet blocks was used for depth prediction, the last channel is $1$ instead of $3$ and the activation of the last convolution layer generator is \textit{Relu} instead of \textit{Tanh}. 

For the image-to-label task, a U-Net256 with skip connections was used for the generator but the output channel size was chosen to be $19$ instead of $3$ for segmentation of $19$ classes. The activation of the last convolution layer of the generator was changed to a softmax to predict class probability for segmentation purposes.

\section{Training details}
\label{sec:dynamics}
\begin{figure}
    \centering
    \includegraphics[width=0.48\textwidth]{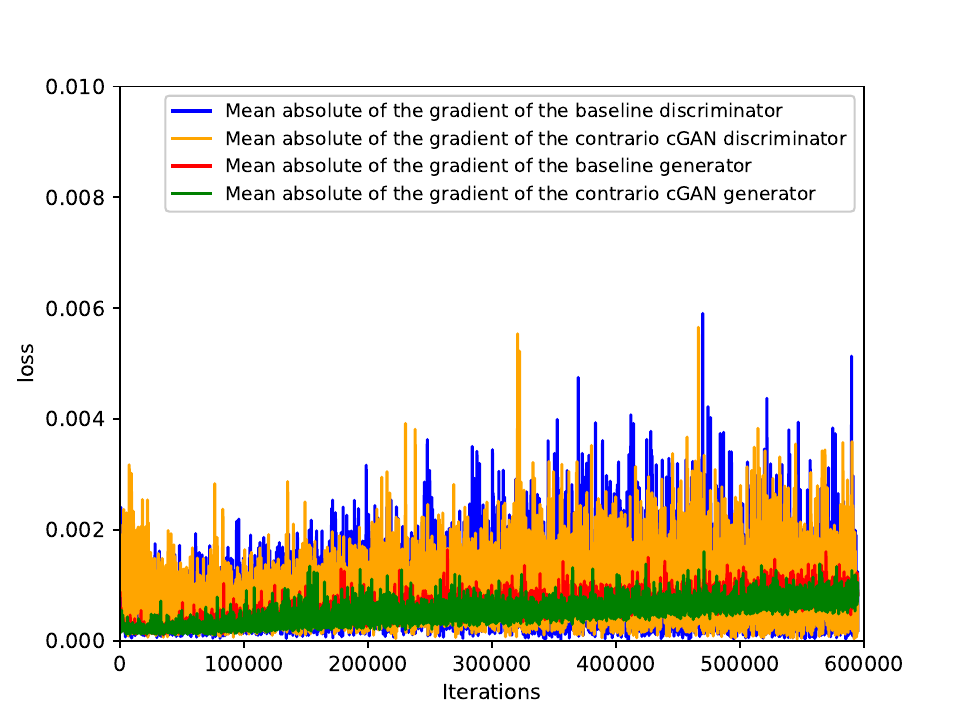}
    \includegraphics[width=0.48\textwidth]{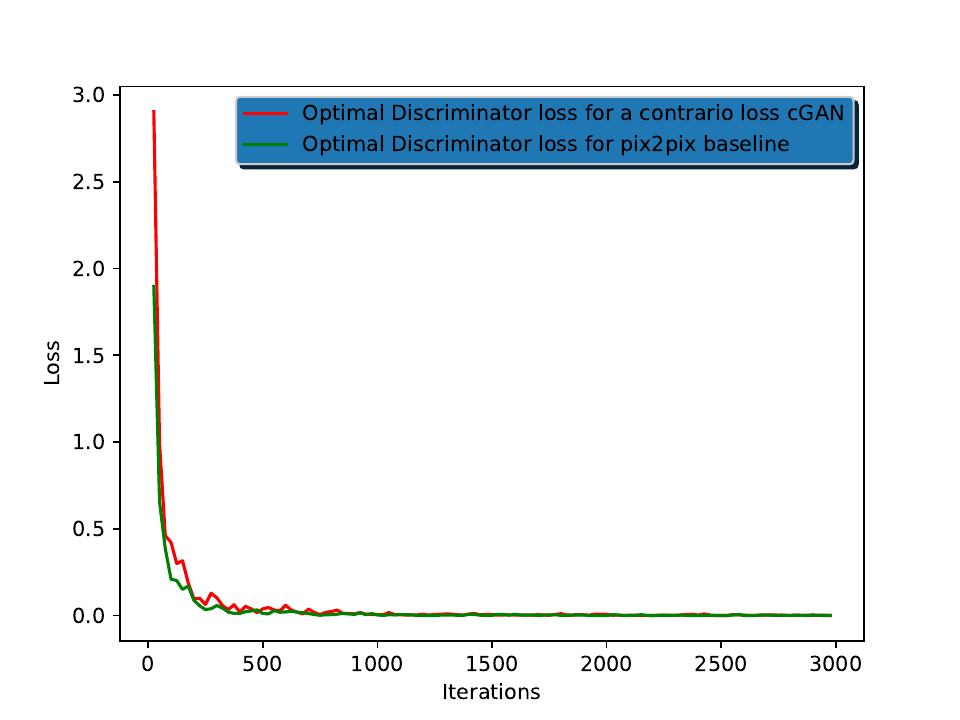}
    \caption{(a) The mean absolute value of the gradients of the generator and discriminator for both baseline and \ac cGAN models trained on Cityscapes\cite{cordts2016cityscapes}. The gradient is stable and it is neither vanishing nor exploding. (b) The loss function of the optimal discriminators when the generator is fixed. Both losses converge rapidly to $0$.}
    \label{fig:dynamics}
\end{figure}

Figure~\ref{fig:dynamics}(a) shows the gradient of the classic and proposed \ac cGANs trained on Cityscapes~\cite{cordts2016cityscapes} label-to-image with and without \ac(see Section \ref{sec:lb-to-im}). The mean absolute value of the gradient is reported in order to demonstrate the stability of the training. Neither vanishing nor exploding gradient is observed for both models. 
Figure~\ref{fig:dynamics}(b) shows the training loss of the optimal discriminator trained as described in Section~\ref{sec:eval} for both models with the generator fixed at epoch $200$. Both models converge rapidly to $0$. Allowing the discriminator to converge for one epoch is enough to obtain the optimal discriminator with a fixed generator.

 \begin{figure}
    \centering
    \includegraphics[width=0.6\textwidth]{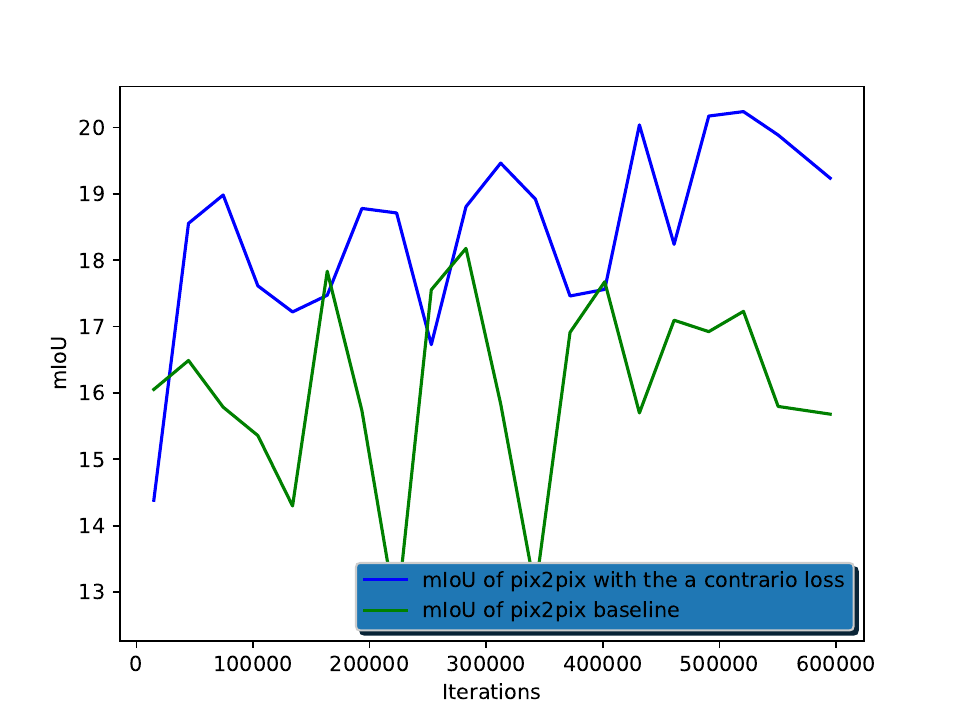}
    \caption{mIoU for the Cityscape image-to-label dataset throughout training. The proposed method consistently obtains more accurate results and finishes with a largely different score at the end of training $19.23$ versus for the baseline $15.97$.}
    \label{fig:city}
\end{figure}

\bibliographystyle{plain}
\bibliography{bib}

\end{document}